\documentclass[10pt,twocolumn,letterpaper]{article}

\usepackage[pagenumbers]{cvpr} % To force page numbers, e.g. for an arXiv version

\usepackage{colortbl}%
\usepackage{graphicx}
\usepackage{amsmath}
\usepackage{amssymb}
\usepackage{booktabs,caption}
\usepackage{multirow}   
\usepackage{bm}
\usepackage{dsfont}
\usepackage[linesnumbered,ruled,vlined]{algorithm2e}
\usepackage{makecell}
\usepackage[flushleft]{threeparttable}
\usepackage{enumitem}
\usepackage{siunitx}
\usepackage[accsupp]{axessibility} % Improves PDF readability for those with disabilities.

\definecolor{MyBlack}{HTML}{323A45}

\def\eg{\emph{e.g.,\ }}
\def\ie{\emph{i.e.,\ }}

\definecolor{cvprblue}{rgb}{0.21,0.49,0.74}
\usepackage[pagebackref,breaklinks,colorlinks,citecolor=cvprblue]{hyperref}
\usepackage[capitalize]{cleveref}

\def\paperID{52} % *** Enter the Paper ID here
\def\confName{CVPR}
\def\confYear{2025}

\begin{document}
\title{The Tenth NTIRE 2025 Efficient Super-Resolution Challenge Report}

%%%%%%%%% Author Information %%%%%%%%%
%-----------------------------------%
% Author Information & Maketitle
%-----------------------------------%
\author{
Bin Ren$^*$\and
Hang Guo$^*$\and
Lei Sun$^*$\and
Zongwei Wu$^*$\and
Radu Timofte$^*$\and
Yawei Li$^*$\and
% Team EMSR
Yao Zhang\and
Xinning Chai\and
Zhengxue Cheng\and 
Yingsheng Qin\and
Yucai Yang\and
Li Song\and 
% Team XiaomiMM
Hongyuan Yu\and 
Pufan Xu\and 
Cheng Wan\and 
Zhijuan Huang\and 
Peng Guo\and 
Shuyuan Cui\and 
Chenjun Li\and 
Xuehai Hu\and 
Pan Pan\and 
Xin Zhang\and 
Heng Zhang\and 
% Team ShannonLab
Qing Luo\and 
Linyan Jiang\and 
Haibo Lei\and 
Qifang Gao\and 
Yaqing Li\and 
% Team TSSR
Weihua Luo\and
Tsing Li\and
% Team mbga
Qing Wang\and
Yi Liu\and
Yang Wang\and
Hongyu An\and
Liou Zhang\and
Shijie Zhao\and
% Team VPEG_C
Lianhong Song\and
Long Sun\and
Jinshan Pan\and
Jiangxin Dong\and
Jinhui Tang\and
% Team XUPTBoys
Jing Wei\and
Mengyang Wang\and
Ruilong Guo\and
Qian Wang\and
% Team HannahSR
Qingliang Liu\and
Yang Cheng\and
% Team Davinci
Davinci\and
Enxuan Gu\and
% Team SRCB % !!! NO valid Report from this team!!!
% Dafeng Zhang\and
% Yang Yong\and
% Team Rochester
Pinxin Liu\and
Yongsheng Yu\and
Hang Hua\and
Yunlong Tang\and
% Team IESR
Shihao Wang\and
Yukun Yang\and
Zhiyu Zhang\and
% Team ASR
Yukun Yang\and
% Team VPEG_O
% Long Sun\and
% Lianhong Song\and
% Jinshan Pan\and
% Jiangxin Dong\and
% Jinhui Tang\and
% Team mmSR
Jiyu Wu\and
Jiancheng Huang\and
Yifan Liu\and
Yi Huang\and
Shifeng Chen\and
% Team ChanSR
Rui Chen\and
% Team Pixel Alchemists
Yi Feng\and
Mingxi Li\and
Cailu Wan\and
Xiangji Wu\and
% Team LZ
Zibin Liu\and
Jinyang Zhong\and
% Team Z6
Kihwan Yoon\and
Ganzorig Gankhuyag\and 
% Team TACO_SR
Shengyun Zhong\and 
Mingyang Wu\and 
Renjie Li\and 
Yushen Zuo\and 
Zhengzhong Tu\and 
% Team AIOT_AI
Zongang Gao\and 
Guannan Chen\and 
Yuan Tian\and 
Wenhui Chen\and 
% Team JNU620
Weijun Yuan\and
Zhan Li\and
Yihang Chen\and
Yifan Deng\and
Ruting Deng\and
% Team LVGroup_HFUT
Yilin Zhang\and
Huan Zheng\and
Yanyan Wei\and
Wenxuan Zhao\and
Suiyi Zhao\and
Fei Wang\and
Kun Li\and
% Team YG
Yinggan Tang\and
Mengjie Su\and
% Team NanoSR
% TODO\and
% TODO\and
% Team MegastudyEdu_Vision_AI
Jae-hyeon Lee\and
Dong-Hyeop Son\and
Ui-Jin Choi\and
% Team XUPTBoys
% Jing Wei\and
% Mengyang Wang\and
% Ruilong Guo\and
% Qian Wang\and
% Team MILA
Tiancheng Shao\and
Yuqing Zhang\and
Mengcheng Ma\and
% Team AiMF_SR
Donggeun Ko\and
Youngsang Kwak\and
Jiun Lee\and
Jaehwa Kwak\and
% Team BVIVSR
Yuxuan Jiang\and
Qiang Zhu\and
Siyue Teng\and
Fan Zhang\and
Shuyuan Zhu\and
Bing Zeng\and
David Bull\and
% Team CUIT_HTT
Jing Hu\and
Hui Deng\and
Xuan Zhang\and
Lin Zhu\and
Qinrui Fan\and
% Team GXZY_AI
Weijian Deng\and
Junnan Wu\and
Wenqin Deng\and
Yuquan Liu\and
Zhaohong Xu\and
% Team IPCV
Jameer Babu Pinjari\and
Kuldeep Purohit\and
% Team X-L
Zeyu Xiao\and
Zhuoyuan Li\and
% Team Quantum\_Res
Surya Vashisth\and
Akshay Dudhane\and
Praful Hambarde\and
Sachin Chaudhary\and
Satya Naryan Tazi\and
Prashant Patil\and
Santosh Kumar Vipparthi\and
Subrahmanyam Murala\and
% Team SylabSR
Wei-Chen Shen\and
I-Hsiang Chen\and
% Team NJUPCA
Yunzhe Xu\and
Chen Zhao\and
Zhizhou Chen\and
% Team DepthIBN
Akram Khatami-Rizi\and
Ahmad Mahmoudi-Aznaveh\and
% Team Cidaut_AI
Alejandro Merino\and
Bruno Longarela\and
Javier Abad\and
Marcos V. Conde\and
% Team IVL
Simone Bianco\and
Luca Cogo\and
Gianmarco Corti\and
}
\maketitle
\let\thefootnote\relax\footnotetext{
$^*$ 
B. Ren (bin.ren@unitn.it, University of Pisa \& University of Trento, Italy), 
H. Guo (cshguo@gmail.com, Tsinghua University), 
L. Sun (lei.sun@insait.ai,INSAIT, Sofia University''St. Kliment Ohridski''), 
Z. Wu (zongwei.wu@uni-wuerzburg.de, University of W\"urzburg, Germany),
R. Timofte (Radu.Timofte@uni-wuerzburg.de,
University of W\"urzburg, Germany), and
Y. Li (yawei.li@vision.ee.ethz.ch, ETH Z\"urich, Switzerland) 
were the challenge organizers, while the other authors participated in the challenge.\\ 
Appendix~\ref{sec:teams} contains the authors' teams and affiliations.\\
NTIRE 2025 webpage: \url{https://cvslai.net/ntire/2025/}.\\ 
Code: \url{https://github.com/Amazingren/NTIRE2025_ESR/}.
}

%%%%%%%%% ABSTRACT
\begin{abstract}
This paper presents a comprehensive review of the NTIRE 2025 Challenge on Single-Image Efficient Super-Resolution (ESR). The challenge aimed to advance the development of deep models that optimize key computational metrics, i.e., runtime, parameters, and FLOPs, while achieving a PSNR of at least 26.90 dB on the $\operatorname{DIV2K\_LSDIR\_valid}$ dataset and 26.99 dB on the $\operatorname{DIV2K\_LSDIR\_test}$ dataset. A robust participation saw \textbf{244} registered entrants, with \textbf{43} teams submitting valid entries. This report meticulously analyzes these methods and results, emphasizing groundbreaking advancements in state-of-the-art single-image ESR techniques. The analysis highlights innovative approaches and establishes benchmarks for future research in the field.
\end{abstract}

%%%%%%%%% BODY TEXT
\section{Introduction}
\label{sec:introduction}
Single image super-resolution (SR) is designed to reconstruct a high-resolution (HR) image from a single low-resolution (LR) image, typically affected by blurring and down-sampling. The standard degradation model in traditional SR, bicubic down-sampling, allows for consistent benchmarks and systematic comparisons among different SR methods. This framework also serves as a platform to highlight the advances in SR technologies. SR techniques are widely used in fields such as satellite imaging, medical image enhancement, and surveillance, where improved image quality is essential for accurate interpretation and analysis.

State-of-the-art deep neural networks for image super-resolution (SR) often suffer from overparameterization, intensive computation, and high latency, making their deployment on mobile devices for real-time SR applications challenging. To address these limitations, extensive research has focused on improving network efficiency through techniques such as network pruning, low-rank filter decomposition, network quantization, neural architecture search, state space modeling, diffusion priors, and knowledge distillation~\cite{vim,liu2018rethinking,ren2024sharing,zheng2025distilling,ren2023masked,yu2017compressing,ma2024shapesplat,zhao2024denoising}. These compression methods, successfully applied to image SR, optimize both the computational footprint and the operational speed~\cite{xie2025mat,ren2024ninth,cao2023ciaosr}.

Efficient SR is particularly crucial for edge computing and mobile devices, where processing power, energy availability, and memory are limited. The enhanced efficiency of SR models ensures that these devices can execute high-quality image processing in real-time without exhausting system resources or draining battery life rapidly. Metrics like runtime, parameter count, and computational complexity (FLOPs) are vital for assessing the suitability of SR models for edge deployment. These parameters are key in maintaining a balance between performance and resource use, ensuring that mobile devices can deliver advanced imaging capabilities efficiently. This balance is critical for the widespread adoption of advanced SR techniques in everyday applications, driving the development of AI-enabled technologies that are both powerful and accessible.

In collaboration with the 2025 New Trends in Image Restoration and Enhancement (NTIRE 2025) workshop, we organize the challenge on single-image efficient super-resolution. The challenge's goal is to super-resolve an LR image with a magnification factor of $\times 4$ using a network that reduces aspects such as runtime, parameters, FLOPs, of EFDN~\cite{EFDN}, 
while at least maintaining the 26.90 dB on the DIV2K\_LSDIR\_valid dataset, and 26.99 dB on the DIV2K\_LSDIR\_test dataset. 
This challenge aims to discover advanced and innovative solutions for efficient SR, benchmark their efficiency, and identify general trends for the design of future efficient SR networks.

% % The cross-cite needs to be updated later from Radu.
% This challenge is one of the NTIRE 2025 Workshop~\footnote{\url{https://cvlai.net/ntire/2025/}} series of challenges on: night photography rendering~\cite{shutova2023ntire_night}, HR depth from images of specular and transparent surfaces~\cite{zama2023ntire_depth}, image denoising~\cite{li2023ntire_dn50}, video colorization~\cite{kang2023ntire_vc}, shadow removal~\cite{vasluianu2023ntire_isr}, quality assessment of video enhancement~\cite{liu2023ntire}, stereo super-resolution~\cite{wang2023ntire_ssr}, light field image super-resolution~\cite{wang2023ntire_lfsr}, image super-resolution ($\times4$)~\cite{zhang2023ntire}, 360° omnidirectional image and video super-resolution~\cite{cao2023ntire}, lens-to-lens bokeh effect transformation~\cite{conde2023ntire_bokeh}, real-time 4K super-resolution~\cite{conde2023ntire_rtsr}, HR nonhomogenous dehazing~\cite{ancuti2023ntire}, low light enhancement \cite{ntire2024lowlight}, efficient super-resolution~\cite{li2023ntire_esr,ren2024ninth}.

%% cross-referencing NTIRE 2025 associated challenges

This challenge is one of the NTIRE 2025~\footnote{\url{https://www.cvlai.net/ntire/2025/}} Workshop associated challenges on: 
ambient lighting normalization~\cite{ntire2025ambient}, 
reflection removal in the wild~\cite{ntire2025reflection}, 
shadow removal~\cite{ntire2025shadow}, 
event-based image deblurring~\cite{ntire2025event}, 
image denoising~\cite{ntire2025denoising}, 
XGC quality assessment~\cite{ntire2025xgc}, 
UGC video enhancement~\cite{ntire2025ugc},
night photography rendering~\cite{ntire2025night},
image super-resolution (x4)~\cite{ntire2025srx4}, 
real-world face restoration~\cite{ntire2025face}, 
efficient super-resolution~\cite{ntire2025esr}, 
HR depth estimation~\cite{ntire2025hrdepth}, 
efficient burst HDR and restoration~\cite{ntire2025ebhdr}, 
cross-domain few-shot object detection~\cite{ntire2025cross}, 
short-form UGC video quality assessment and enhancement~\cite{ntire2025shortugc,ntire2025shortugc_data}, 
text to image generation model quality assessment~\cite{ntire2025text}, 
day and night raindrop removal for dual-focused images~\cite{ntire2025day}, 
video quality assessment for video conferencing~\cite{ntire2025vqe}, low light image enhancement~\cite{ntire2025lowlight}, 
light field super-resolution~\cite{ntire2025lightfield}, 
restore any image model (RAIM) in the wild~\cite{ntire2025raim}, 
raw restoration and super-resolution~\cite{ntire2025raw} and 
raw reconstruction from RGB on smartphones~\cite{ntire2025rawrgb}.

\section{NTIRE 2025 Efficient Super-Resolution Challenge}
The goals of this challenge include: 
(i) promoting research in the area of single-imae efficient super-resolution, 
(ii) facilitating comparisons between the efficiency of various methods, and 
(iii) providing a platform for academic and industrial participants to engage, discuss, and potentially establish collaborations. This section delves into the specifics of the challenge.

\subsection{Dataset}
The DIV2K~\cite{agustsson2017ntire} dataset and LSDIR~\cite{li2023lsdir} dataset are utilized for this challenge. The DIV2K dataset consists of 1,000 diverse 2K resolution RGB images, which are split into a training set of 800 images, a validation set of 100 images, and a test set of 100 images. The LSDIR dataset contains 86,991 high-resolution high-quality images, which are split into a training set of 84,991 images, a validation set of 1,000 images, and a test set of 1,000 images. In this challenge, the corresponding LR DIV2K images are generated by bicubic downsampling with a down-scaling factor of 4$\times$. 
The training images from DIV2K and LSDIR are provided to the participants of the challenge. During the validation phase, 100 images from the DIV2K validation set and 100 images from the LSDIR validation set are made available to participants. 
During the test phase, 100 images from the DIV2K test set and another 100 images from the LSDIR test set are used. 
Throughout the entire challenge, the testing HR images remain hidden from the participants.

\subsection{EFDN Baseline Model}
\label{sec:baseline_model}

The Edge-Enhanced Feature Distillation Network (EFDN)~\cite{EFDN} serves as the baseline model in this challenge. 
The aim is to improve its efficiency in terms of runtime, number of parameters, and FLOPs, while at least maintaining 26.90 dB on the DIV2K\_LSDIR\_valid dataset and 26.99 dB on the DIV2K\_LSDIR\_test dataset. 

The main idea within EFDN is a combination of block composing, architecture searching,
and loss designing to obtain a trade-off between performance and light-weighting. Especially, 
For block composing, EFDN sum up the re-parameterization methods~\cite{ding2019acnet,ding2021diverse,zhang2021edge} and designs a more effective and complex edge-enhanced diverse branch block. 
In detail, they employ several reasonable
reparameterizable branches to enhance the structural information extraction, and then they integrate them into a vanilla convolution to maintain the inference performance.
To ensure the effective optimization of parallel branches in EDBB, they designed an edge-enhanced gradient-variance loss (EG) based on the gradient-variance loss~\cite{abrahamyan2022gradient}. 
The proposed loss enforces minimizing the difference between the computed variance maps, which is helpful to restore sharper edges.
The gradient maps calculated by different filters and the corresponding EG loss. 
In addition, the NAS strategy of DLSR is adopted to search for a robust backbone.

The baseline EFDN emerges as the 1st place for the overall performance of the NTIRE2023 Efficient SR Challenge~\cite{EFDN}. The quantitative performance and efficiency metrics of EFDN are summarized as follows:
(1) The number of parameters is 0.276 M. 
(2) The average PSNRs on validation (DIV2K 100 valid images and LSDIR 100 valid images) and testing (DIV2K 100 test images and LSDIR 100 test images) sets of this challenge are 26.93 dB and 27.01 dB, respectively. 
(3) The runtime averaged to 22.18ms on the validation and test set with PyTorch 2.0.0+cu118, and a single NVIDIA RTX A6000 GPU. 
(4) The number of FLOPs for an input of size $256\times256$ is 16.70 G.

\subsection{Tracks and Competition}
The aim of this challenge is to devise a network that reduces one or several aspects such as runtime, parameters, and FLOPs, while at least maintaining the 26.90 dB on the DIV2K\_LSDIR valid dataset, and 26.99 dB on the DIV2K\_LSDIR test dataset. 

\medskip
\noindent{\textbf{Challenge phases: }}
\textit{(1) Development and validation phase}: Participants were given access to 800 LR/HR training image pairs and 200 LR/HR validation image pairs from the DIV2K and the LSDIR datasets. 
An additional 84,991 LR/HR training image pairs from the LSDIR dataset are also provided to the participants. 
The EFDN model, pre-trained parameters, and validation demo script are available on GitHub \url{https://github.com/Amazingren/NTIRE2025_ESR}, allowing participants to benchmark their models' runtime on their systems. 
Participants could upload their HR validation results to the evaluation server to calculate the PSNR of the super-resolved image produced by their models and receive immediate feedback. 
The corresponding number of parameters, FLOPs, and runtime will be computed by the participants.
\textit{(2) Testing phase}: In the final test phase, participants were granted access to 100 LR testing images from DIV2K and 100 LR testing images from LSDIR, while the HR ground-truth images remained hidden. Participants submitted their super-resolved results to the Codalab evaluation server and emailed the code and factsheet to the organizers. 
The organizers verified and ran the provided code to obtain the final results, which were then shared with participants at the end of the challenge.

\medskip
\noindent{\textbf{Evaluation protocol: }}
Quantitative evaluation metrics included validation and testing PSNRs, runtime, FLOPs, and the number of parameters during inference. 
PSNR was measured by discarding a 4-pixel boundary around the images. 
The average runtime during inference was computed on the 200 LR validation images and the 200 LR testing images.
The average runtime on the validation and testing sets served as the final runtime indicator. 
FLOPs are evaluated on an input image of size 256$\times$256. 
Among these metrics, runtime was considered the most important. 
Participants were required to maintain a PSNR of at least 26.90 dB on the DIV2K\_LSDIR valid dataset, and 26.99 dB on the DIV2K\_LSDIR test dataset during the challenge.  
The constraint on the testing set helped prevent overfitting on the validation set.
It's important to highlight that methods with a PSNR below the specified threshold (\ie 26.90 dB on DIV2K\_LSDIR\_valid and, 26.99 dB on DIV2K\_LSDIR\_test) will not be considered for the subsequent ranking process. It is essential to meet the minimum PSNR requirement to be eligible for further evaluation and ranking.
A code example for calculating these metrics is available at \url{https://github.com/Amazingren/NTIRE2025_ESR}.

To better quantify the rankings, we followed the scoring function from NTIRE2024 ESR~\cite{ren2024ninth} for three evaluation metrics in this challenge: 
runtime, 
FLOPs, and 
parameters. 
This scoring aims to convert the performance of each metric into corresponding scores to make the rankings more significant. Especially, the score for each separate metric (\ie Runtime, FLOPs, and parameter) for each sub-track is calculated as:
\begin{equation}
    \begin{aligned}
    Score\_{Metric} = \dfrac{\text{Exp} (2 \times Metric_{TeamX})}{Metric_{Baseline}},
    \end{aligned}
    \label{equ:score1}
\end{equation}
based on the score of each metric, the final score used for the main track is calculated as:
\begin{equation}
    \begin{aligned}
    Score\_{Final} & = w_{1} \times Score\_Runtime \\
    & + w_{2} \times Score\_FLOPs \\
    & + w_{3} \times Score\_Params,
    \end{aligned}
    \label{equ:score2}
\end{equation}
where $w_{1}$, $w_{2}$, and $w_{3}$ are set to 0.7, 0.15, and 0.15, respectively. This setting is intended to incentivize participants to design a method that prioritizes speed efficiency while maintaining a reasonable model complexity.

%-----------------------%
\section{Challenge Results}
\label{sec:esr_results}
%-----------------------%
\begin{table*}[!ht]
% \scriptsize
\centering
\setlength{\extrarowheight}{0.7pt}
\setlength{\tabcolsep}{16pt}
\caption{Results of Ninth NTIRE 2025 Efficient SR Challenge. The performance of the solutions is compared thoroughly from three perspectives including the runtime, FLOPs, and the number of parameters. The underscript numbers associated with each metric score denote the ranking of the solution in terms of that metric. For runtime, ``Val.'' is the runtime averaged on DIV2K\_LSDIR\_valid validation set. ``Test'' is the runtime averaged on a test set with 200 images from DIV2K\_LSDIR\_test set, respectively. ``Ave.'' is averaged on the validation and test datasets. 
``\#Params'' is the total number of parameters of a model. ``FLOPs'' denotes the floating point operations. 
Main Track combines all three evaluation metrics.
The ranking for the main track is based on the score calculated via Eq.~\ref{equ:score2}, and the ranking for other sub-tracks is based on the score of each metric via Eq.~\ref{equ:score1}. Please note that \textbf{this is not a challenge for PSNR improvement. The ``validation/testing PSNR'' is not ranked. For all the scores, the lower, the better.}
}
\label{tbl:final_results}
\begin{threeparttable}
    \resizebox{\linewidth}{!}
    {
    \begin{tabular}{@{\extracolsep{\fill}} 
                                    l|
                                    |
                                    S[table-format=2.2]
                                    S[table-format=2.2]
                                    |
                                    S[table-format=3.3] 
                                    S[table-format=3.3] 
                                    S[table-format=3.3]
                                    |
                                    S[table-format=1.3] 
                                    S[table-format=2.2] 
                                    |
                                    S[table-format=3.2$_{(3)}$] 
                                    S[table-format=3.2$_{(3)}$] 
                                    S[table-format=2.2$_{(3)}$]
                                    |
                                    S[table-format=3.2$_{(3)}$]
                                    S[table-format=2]}
                                    
    \toprule[1.0pt]
    \multirow{2}{*}{Teams}      & \multicolumn{2}{c|}{PSNR [dB]} & \multicolumn{3}{c|}{Runtime [ms]}  & {\#Params}  &   {FLOPs}  & \multicolumn{3}{c|}{Sub-Track Scores}  &\multicolumn{2}{c}{Main-Track}   \\ \cline{2-6} \cline{9-13}
    &{Val.} & {Test} & {Val.} & {Test} & {Ave.}  & {[M]} & {[G]} & {Runtime} & {\#Params} & {FLOPs} & {Overall Score} & {Ranking}  \\ 
    \midrule[0.6pt]
    \textbf{EMSR} & 26.92 & 26.99 & 10.268 & 9.720 & 9.994 & 0.131 & 8.54 & 2.46$_{(5)}$ & 2.58$_{(6)}$ & 2.78$_{(7)}$ & 2.53 & \textbf{1}\\
    \textbf{XiaomiMM} & 26.92 & 27.00 & 9.958 & 9.132 & 9.545 & 0.148 & 9.68 & 2.36$_{(4)}$ & 2.92$_{(11)}$ & 3.19$_{(13)}$ & 2.57 & \textbf{2} \\
    \textbf{ShannonLab} & 26.90 & 27.00 & 8.938 & 8.302 & 8.620 & 0.172 & 11.23 & 2.18$_{(1)}$ & 3.48$_{(17)}$ & 3.84$_{(18)}$ & 2.62 & \textbf{3} \\
    TSSR & 26.90 & 27.02 & 9.812 & 8.898 & 9.355 & 0.164 & 10.69 & 2.32$_{(2)}$ & 3.28$_{(15)}$ & 3.60$_{(16)}$ & 2.66 & 4 \\
    Davinci & 26.92 & 27.00 & 11.426 & 9.876 & 10.651 & 0.146 & 9.55 & 2.61$_{(6)}$ & 2.88$_{(9)}$ & 3.14$_{(11)}$ & 2.73 & 5 \\
    SRCB & 26.92 & 27.00 & 11.412 & 9.960 & 10.686 & 0.146 & 9.55 & 2.62$_{(7)}$ & 2.88$_{(9)}$ & 3.14$_{(11)}$ & 2.74 & 6 \\
    Rochester & 26.94 & 27.01 & 11.934 & 10.454 & 11.194 & 0.158 & 10.30 & 2.74$_{(8)}$ & 3.14$_{(14)}$ & 3.43$_{(14)}$ & 2.91 & 7 \\
    mbga & 26.90 & 27.00 & 9.822 & 9.208 & 9.515 & 0.192 & 12.56 & 2.36$_{(3)}$ & 4.02$_{(19)}$ & 4.50$_{(20)}$ & 2.93 & 8 \\
    IESR & 26.90 & 26.99 & 13.760 & 12.582 & 13.171 & 0.143 & 8.32 & 3.28$_{(10)}$ & 2.82$_{(7)}$ & 2.71$_{(6)}$ & 3.12 & 9 \\
    ASR & 26.90 & 27.00 & 13.864 & 11.984 & 12.924 & 0.154 & 9.06 & 3.21$_{(9)}$ & 3.05$_{(12)}$ & 2.96$_{(8)}$ & 3.15 & 10 \\
    VPEG\_O & 26.90 & 26.99 & 16.356 & 13.926 & 15.141 & 0.145 & 9.42 & 3.92$_{(12)}$ & 2.86$_{(8)}$ & 3.09$_{(9)}$ & 3.63 & 11 \\
    mmSR & 26.95 & 27.05 & 14.450 & 12.036 & 13.243 & 0.212 & 13.85 & 3.30$_{(11)}$ & 4.65$_{(21)}$ & 5.25$_{(23)}$ & 3.80 & 12 \\
    ChanSR & 26.92 & 27.03 & 16.738 & 15.592 & 16.165 & 0.210 & 11.59 & 4.29$_{(16)}$ & 4.58$_{(20)}$ & 4.01$_{(19)}$ & 4.29 & 13 \\
    Pixel Alchemists & 26.90 & 27.01 & 17.322 & 14.608 & 15.965 & 0.213 & 12.93 & 4.22$_{(14)}$ & 4.68$_{(22)}$ & 4.70$_{(21)}$ & 4.36 & 14 \\
    MiSR & 26.90 & 27.02 & 17.056 & 14.988 & 16.022 & 0.213 & 13.86 & 4.24$_{(15)}$ & 4.68$_{(22)}$ & 5.26$_{(24)}$ & 4.46 & 15 \\
    LZ & 26.90 & 27.01 & 16.980 & 15.450 & 16.215 & 0.252 & 16.42 & 4.31$_{(17)}$ & 6.21$_{(25)}$ & 7.15$_{(25)}$ & 5.02 & 16 \\
    Z6 & 26.90 & 26.99 & 20.362 & 16.184 & 18.273 & 0.303 & 18.70 & 5.19$_{(20)}$ & 8.99$_{(27)}$ & 9.39$_{(27)}$ & 6.39 & 17 \\
    TACO\_SR & 26.94 & 27.05 & 17.828 & 15.652 & 16.740 & 0.342 & 20.03 & 4.52$_{(18)}$ & 11.92$_{(30)}$ & 11.01$_{(30)}$ & 6.61 & 18 \\
    AIOT\_AI & 26.90 & 27.00 & 19.836 & 18.158 & 18.997 & 0.301 & 19.56 & 5.54$_{(21)}$ & 8.86$_{(26)}$ & 10.41$_{(28)}$ & 6.77 & 19 \\
    JNU620 & 26.90 & 27.01 & 20.688 & 18.282 & 19.485 & 0.325 & 20.31 & 5.79$_{(22)}$ & 10.54$_{(29)}$ & 11.39$_{(31)}$ & 7.34 & 20 \\
    LVGroup\_HFUT & 26.96 & 27.07 & 16.394 & 14.876 & 15.635 & 0.426 & 27.87 & 4.09$_{(13)}$ & 21.91$_{(33)}$ & 28.15$_{(34)}$ & 10.38 & 21 \\
    SVM & 26.92 & 27.04 & 30.610 & 28.134 & 29.372 & 0.251 & 13.39 & 14.13$_{(23)}$ & 6.16$_{(24)}$ & 4.97$_{(22)}$ & 11.56 & 22 \\
    YG & 26.92 & 27.04 & 33.658 & 31.614 & 32.636 & 0.093 & 5.82 & 18.96$_{(24)}$ & 1.96$_{(5)}$ & 2.01$_{(5)}$ & 13.87 & 23 \\
    NanoSR & 26.97 & 27.08 & 17.930 & 16.300 & 17.115 & 0.551 & 36.02 & 4.68$_{(19)}$ & 54.20$_{(35)}$ & 74.72$_{(35)}$ & 22.61 & 24 \\
    MegastudyEdu Vision AI & 27.01 & 27.13 & 39.376 & 37.528 & 38.452 & 0.169 & 10.63 & 32.03$_{(25)}$ & 3.40$_{(16)}$ & 3.57$_{(15)}$ & 23.47 & 25 \\
    XUPTBoys & 26.91 & 27.03 & 50.564 & 35.012 & 42.788 & 0.072 & 3.39 & 47.36$_{(26)}$ & 1.68$_{(3)}$ & 1.50$_{(2)}$ & 33.63 & 26 \\
    MILA & 26.90 & 27.02 & 44.362 & 42.034 & 43.198 & 0.087 & 4.93 & 49.14$_{(27)}$ & 1.88$_{(4)}$ & 1.80$_{(4)}$ & 34.95 & 27 \\
    AiMF\_SR & 26.98 & 27.10 & 46.594 & 43.092 & 44.843 & 0.180 & 9.48 & 57.00$_{(28)}$ & 3.69$_{(18)}$ & 3.11$_{(10)}$ & 40.92 & 28 \\
    EagleSR & 27.04 & 27.16 & 47.730 & 45.192 & 46.461 & 0.352 & 21.89 & 65.95$_{(29)}$ & 12.82$_{(31)}$ & 13.76$_{(32)}$ & 50.15 & 29 \\
    BVIVSR & 26.97 & 26.99 & 49.488 & 46.798 & 48.143 & 0.155 & 10.79 & 76.75$_{(30)}$ & 3.07$_{(13)}$ & 3.64$_{(17)}$ & 54.73 & 30 \\
    HannahSR & 26.90 & 27.02 & 58.286 & 41.422 & 49.854 & 0.060 & 3.75 & 89.55$_{(31)}$ & 1.54$_{(2)}$ & 1.57$_{(3)}$ & 63.15 & 31 \\
    VPEG\_C & 26.90 & 27.00 & 60.046 & 40.950 & 50.498 & 0.044 & 3.13 & 94.90$_{(32)}$ & 1.38$_{(1)}$ & 1.45$_{(1)}$ & 66.86 & 32 \\
    CUIT\_HTT & 27.09 & 27.20 & 62.038 & 59.106 & 60.572 & 0.309 & 19.75 & 235.36$_{(33)}$ & 9.39$_{(28)}$ & 10.65$_{(29)}$ & 167.76 & 33 \\
    GXZY AI & 27.01 & 27.13 & 102.924 & 99.102 & 101.013 & 0.428 & 25.88 & 9.02$e3_{(34)}$ & 22.23$_{(34)}$ & 22.18$_{(33)}$ & 6.32$e3$ & 34 \\
    SCMSR & 26.92 & 27.00 & 133.866 & 114.088 & 123.977 & 0.393 & 17.62 & 7.15$e4_{(35)}$ & 17.25$_{(32)}$ & 8.25$_{(26)}$ & 5.01$e4$ & 35 \\
    IPCV & 27.27 & 27.40 & 366.924 & 357.268 & 362.096 & 0.866 & 65.66 & 1.51$e14_{(36)}$ & 531.32$_{(37)}$ & 2.60$e3_{(36)}$ & 1.05$e14$ & 36 \\
    X-L & 27.07 & 27.21 & 525.966 & 479.346 & 502.656 & 0.966 & 70.83 & 4.81$e19_{(37)}$ & 1.10$e3_{(38)}$ & 4.83$e3_{(37)}$ & 3.36$e19$ & 37 \\
    Quantum Res & 27.29 & 27.40 & 574.632 & 558.934 & 566.783 & 0.790 & 76.09 & 1.56$e22_{(38)}$ & 306.32$_{(36)}$ & 9.07$e3_{(38)}$ & 1.09$e22$ & 38 \\
    
    \midrule[0.6pt]
    \multicolumn{13}{c}{The following methods are not ranked since their validation/testing PSNR (underlined) is not on par with the threshold.}\\ 
    \midrule[0.6pt]
    SylabSR & 24.36 & 24.46 & 28.580 & 24.826 & 26.703 & 0.072 & 7.90 & 11.11 & 1.68 & 2.58 & 8.41 & {-} \\
    NJUPCA & 26.70 & 26.80 & 70.202 & 52.932 & 61.567 & 2.308 & 30.11 & 257.45 & 1.83$e7$ & 36.82 & 2.75$e6$ & {-} \\
    DepthIBN & 26.56 & 26.66 & 39.154 & 36.876 & 38.015 & 0.121 & 7.71 & 30.80 & 2.40 & 2.52 & 22.30 & {-} \\
    Cidaut AI & 26.86 & 26.95 & 27.220 & 24.974 & 26.097 & 0.210 & 12.83 & 10.52 & 4.58 & 4.65 & 8.75 & {-} \\
    IVL & 26.66 & 26.76 & 18.746 & 16.944 & 17.845 & 0.240 & 15.64 & 5.00 & 5.69 & 6.51 & 5.33 & {-} \\
    \midrule[0.6pt]
    Baseline & 26.93 & 27.01 & 23.912 & 20.454	& 22.183	& 0.276	& 16.7 & 7.39 & 7.39 & 7.39 & 7.39 & {-}\\
    \bottomrule[1pt]
    \end{tabular}
    }
\end{threeparttable}
\end{table*}

The final challenge results and the corresponding rankings are presented in Tab.~\ref{tbl:final_results}
The table also includes the baseline method EFDN~\cite{EFDN} for comparison. 
In Sec.\ref{sec:methods_and_teams}, the methods evaluated in Tab.~\ref{tbl:final_results} are briefly explained, while the team members are listed in~\ref{sec:teams}. 
The performance of different methods is compared from four different perspectives, including 
the runtime, 
FLOPs, 
the parameters, 
and the overall performance.
Furthermore, in order to promote a fair competition emphasizing efficiency, the criteria for image reconstruction quality in terms of test PSNR are set to 26.90 and 26.99 on the DIV2K\_LSDIR\_valid and DIV2K\_LSDIR\_test sets, respectively.

\noindent\textbf{Runtime. }In this challenge, runtime stands as the paramount evaluation metric. \textbf{ShannonLab}'s solution emerges as the frontrunner with the shortest runtime among all entries in the efficient SR challenge, securing its top-3 ranking position. 
Following closely, the TSSR and mbga claim the second and third spots, respectively. 
Remarkably, the average runtime of the top three solutions on both the validation and test sets remains below 10 ms. 
Impressively, the first 13 teams present solutions with an average runtime below 16 ms, showcasing a continuous enhancement in the efficiency of image SR networks. 
Despite the slight differences in runtime among the top three teams, the challenge retains its competitive edge. 
An additional distinction from previous challenges worth noting is that this year, runtime performance no longer predominantly dictates the overall rankings as it has in the past, where the top three solutions in terms of runtime were also the top performers in the main track (\eg from NTIRE ESR 2024~\cite{ren2024ninth}). This shift indicates that participants are now emphasizing a more balanced approach, focusing not only on runtime optimization but also on improving the comprehensive performance of their models

\noindent\textbf{Parameters.} Model complexity was further evaluated by considering the number of parameters, as detailed in Table~\ref{tbl:final_results}. In this sub-track, \textbf{VEPG\_C} achieved the top position with only 0.044M parameters, closely followed by HannahSR and XUPTBoys with 0.060M and 0.072M parameters, respectively. The minimal disparity among the top three methods highlights their competitive edge and efficiency in managing model complexity. They were scored at 1.38, 1.54, and 1.68, respectively, indicating a tight competition. However, it is noteworthy that these models also exhibited relatively high runtimes, suggesting an area for potential improvement in future iterations.

\noindent\textbf{FLOPs.} The number of floating-point operations (FLOPs) is another critical metric for assessing model complexity. Within this sub-track,\textbf{ VEPG\_C}, XUPTBoys, and HannahSR secured the top three positions with FLOPs of 3.13G, 3.39G, and 3.75G, respectively. The competitiveness of this sub-track is further confirmed by the close scores of 1.45, 1.50, and 1.57, aligned with the parameter evaluation results. Remarkably, the same models top both the parameters and FLOPs evaluations, demonstrating consistent performance across different complexity metrics. Similar to the parameters sub-track, the extended runtimes of these methods point to a need for further research and optimization. Key implications include:
i) \textit{Efficiency vs. Performance Trade-off}: The close competition among the top models in terms of parameters and FLOPs suggests a significant trade-off between model efficiency and performance. Despite achieving minimal parameter counts and FLOPs, the high runtimes indicate that these models might be optimizing computational complexity at the expense of execution speed. This raises important considerations for future research in balancing efficiency with real-world usability, especially in applications where inference speed is critical.
ii) \textit{Potential for Model Optimization}: The consistency in ranking between the parameters and FLOPs sub-tracks reveals that models which are optimized for one aspect of computational efficiency tend to perform well in others. However, the noted high runtimes across these models suggest an untapped potential for holistic model optimization. Future work could focus on integrating more advanced optimization techniques or exploring novel architectural innovations to enhance both the computational efficiency and runtime performance.

\noindent\textbf{Overall Evaluation.} The final assessment of performance employs a comprehensive metric that synthesizes runtime, FLOPs, and the number of parameters into a unified score. In this rigorous evaluation, the \textbf{EMSR} Group excelled, claiming the prestigious top position, followed by XiaomiMM (the winner of the NTIRE ESR 2024 challenge) and ShannonLab in second and third places, respectively. This achievement highlights the sophisticated engineering and innovative approaches implemented by these groups.

Contrasting with the previous year, where runtime heavily influenced overall rankings, this year presents a shift. The best performer in runtime only secured third place in the overall competition. Specifically, EMSR, the overall winner, ranked fifth in runtime, sixth in parameters, and seventh in FLOPs. Similarly, XiaomiMM, which came second overall, was fourth in runtime, eleventh in parameters, and thirteenth in FLOPs. This demonstrates that:
i) A balanced approach to model design, optimizing across multiple metrics rather than focusing on a single aspect, is becoming crucial in competitive evaluations.
ii) Achieving top performance in one metric does not guarantee similar success in overall rankings, underscoring the complexity of model optimization in real-world scenarios.
This year's goal was to encourage a balanced pursuit of speed and efficiency, a challenge that has evidently led to significant innovations and advancements in model design.

\noindent\textbf{PSNR.} Team \textbf{Quantum Res}, IPCV, X-L, and CUIT\_HTT demonstrate superior PSNR values, a critical evaluation metric in super-resolution. Specifically, Quantum Res and IPCV lead with an exceptional 27.40 dB, closely followed by X-L with 27.21 dB, and CUIT\_HTT at 27.20 dB on the DIV2K\_LSDIR\_test set. Despite these impressive performances, it is essential to emphasize that the primary focus of this challenge is on \textit{efficiency in super-resolution}. Accordingly, we have adjusted the PSNR criteria, setting rigorous lower thresholds of 26.90 dB and 26.99 dB for the DIV2K\_LSDIR\_valid and DIV2K\_LSDIR\_test sets, respectively. This adjustment is designed to prioritize a balance between high performance and computational efficiency.
A commendable total of 38 teams met this adjusted benchmark, demonstrating their capability to effectively balance image quality with efficiency. However, teams like IVL, Cidaut AI, SylabSR DepthIB, and NJUPCA, while notable for their efficiency, did not achieve the required PSNR levels. This highlights the ongoing challenge of optimizing super-resolution processes that meet both efficiency and performance standards, underscoring the complex nature of advancements in this field.

\subsection{Main Ideas}
Throughout this challenge, several techniques have been proposed to enhance the efficiency of deep neural networks for image super-resolution (SR) while striving to maintain optimal performance. The choice of techniques largely depends on the specific metrics that a team aims to optimize. Below, we outline some typical ideas that have emerged:

\begin{itemize}
    \item \textbf{Distillation is an effective manner to maintain the PSNR performance without increasing computation cost during inference}. The team EMSR added only the ConvLora-Like~\cite{aleem2024convlora} operation into the base model. Similarly, team ESPAN  also proposed to use the self-distillation for progressive learning strategy validated from~\cite{rife}.
    \item \textbf{Re-parameterization}~\cite{ding2021repvgg} ~\cite{du2022parameter,yang2021simam} \textbf{is commonly used in this challenge}. Usually, a normal convolutional layer with multiple basic operations (3 × 3 convolution, 1 × 1 operation, first and second-order derivative operators, skip connections) is parameterized during training. During inference, the multiple operations that reparameterize a convolution could be merged back into a single convolution. \eg Some top teams (\ie XiaomiMM, mmSR, HannahSR, etc) used this operation in their methods.
    \item \textbf{Parameter-free attention mechanism is validated as a useful technique to enhance computational efficiency}~\cite{du2022parameter,yang2021simam}. Specifically, XiaomiMM proposed a swift parameter-free attention network based on parameter-free attention, which achieves the lowest runtime while maintaining a decent PSNR performance.
    \item \textbf{Incorporating multi-scale information and hierarchical module design are proven strategies for effectively fusing critical information}. For instance, solutions such as HannahSR, XuPTBoys, and ChanSR have successfully utilized multi-scale residual connections and hierarchical module designs to enhance their performance.
    \item \textbf{Network pruning plays an important role}. It is observed that a couple of teams (\ie ASR, Davinci) used network pruning techniques to slightly compress a network. This leads to a more lightweight architecture without a heavy performance drop. 
    \item \textbf{Exploration with new network architectures is conducted.} Besides the common CNN or Transformers, the state space model (\ie vision mamba~\cite{mamba,guo2024mambairv2}) was tried by GXZY\_AI in this challenge, which was also validated in the last NTIRE ESR challenge~\cite{ren2024ninth}.
    \item \textbf{Various other techniques are also attempted}. Some teams also proposed solutions based on neural architecture search, vision transformers, frequency processing, multi-stage design, and advanced training strategies.
\end{itemize}

\subsection{Fairness}
To ensure the integrity and fairness of the Efficient SR Challenge, we meticulously established a set of rules focusing on the permissible datasets for training the models. Participants were allowed to augment their training with external datasets, such as Flickr2K, to promote diverse and comprehensive model training experiences. However, to guarantee an unbiased evaluation, the use of additional DIV2K and LSDIR validation sets, which include both high-resolution (HR) and low-resolution (LR) images, was explicitly prohibited during the training phase. This restriction aimed to maintain the validation set's integrity as a vital benchmark for assessing the proposed networks' performance and generalizability. Moreover, using LR images from the DIV2K and LSDIR test sets for training was strictly forbidden, ensuring the test dataset's purity and upholding the evaluation process's integrity. Lastly, the adoption of advanced data augmentation techniques during training was encouraged as a fair practice, allowing participants to enhance their models within the defined rules and guidelines.

\subsection{Conclusions}
The analysis of the submissions to this year's Efficient SR Challenge allows us to draw several important conclusions:

$\bullet$ Firstly, the competition within the image super-resolution (SR) community remains intense. This year, the challenge attracted \textbf{244} registered participants, with \textbf{43} teams making valid submissions. All proposed methods have enhanced the state-of-the-art in efficient SR. Notably, the competition among the top three teams has intensified, with last year's winner ranking second this year.

$\bullet$ Secondly, unlike in previous challenges, dominance in runtime no longer characterizes the top-ranking teams. Instead, more balanced solutions that consider all aspects of performance are proving to be more beneficial.

$\bullet$ Thirdly, consistent with the success of deep learning techniques like DeepSeek, the distillation approach has significantly contributed to performance improvements without adding computational complexity.

$\bullet$ Fourthly, re-parameterization and network compression have emerged as crucial techniques in enhancing efficiency in SR. Ongoing exploration in these areas is encouraged to further boost efficiency.

$\bullet$ Fifthly, the use of large-scale datasets, such as the one described in \cite{li2023lsdir}, for pre-training has been shown to enhance accuracy significantly. Typically, training incorporates multiple phases, gradually increasing the patch size and decreasing the learning rate, optimizing the training process.

$\bullet$ Sixthly, this year's challenge saw the introduction of the state space model, presenting a novel approach that may influence future research directions in the field.

Overall, by considering factors like runtime, FLOPs, and parameter count simultaneously, it is feasible to design models that optimize across multiple evaluation metrics.
Finally, as computational capabilities continue to evolve, the focus on optimizing models for runtime, FLOPs, and parameter efficiency becomes increasingly vital. With advancements in both hardware and software, we expect the development of more sophisticated and efficient models in the super-resolution domain. The pursuit of efficiency in SR is likely to remain a key driver of innovation, promising exciting advancements and continual progress in the field.
\section{Challenge Methods and Teams}
\label{sec:methods_and_teams}
% ---- 1st - 3rd of the Main Track 
\subsection{EMSR}

\begin{figure*}[t]
  \centering
   \includegraphics[width=\linewidth]{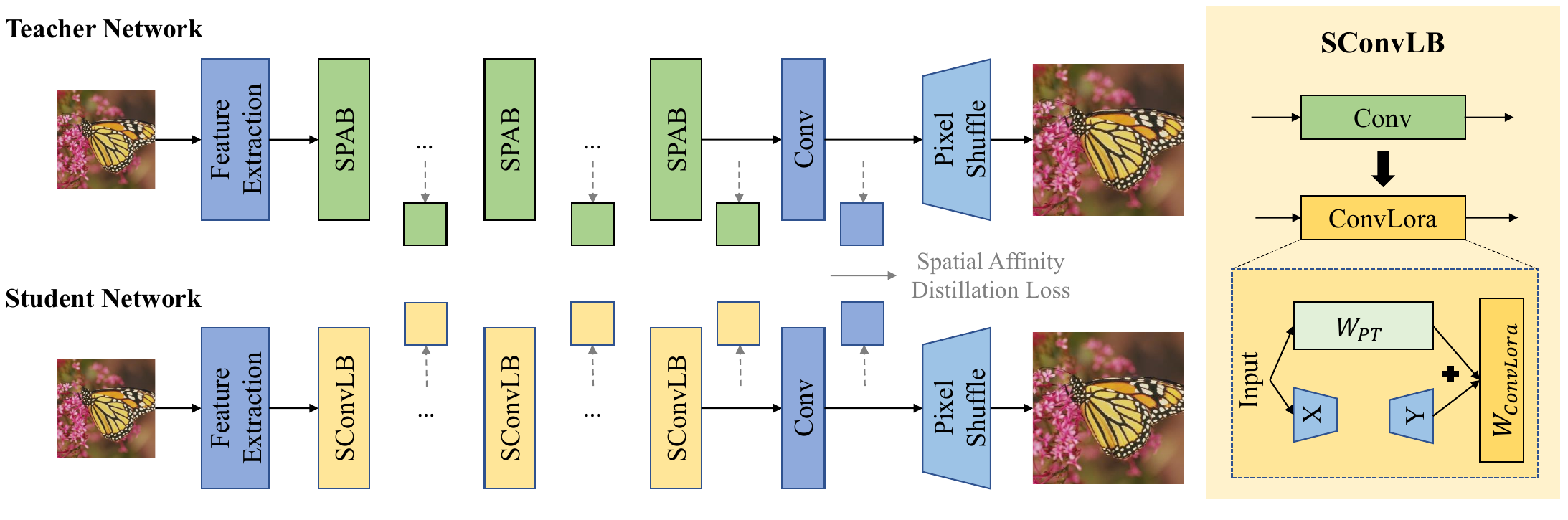}
   \caption{\textit{Team EMSR}: The team incorporates ConvLoras into the network to increase the performance without adding extra complexity.}
   \label{fig:overview}
   \vspace{-1em}
\end{figure*}

\noindent \textbf{Method.} The overall architecture of the team EMSR is shown in \cref{fig:overview}, which is based on the leading efficient super-resolution method SPAN \cite{wan2024swift}. Inspired by ConvLora \cite{aleem2024convlora}, the team proposes SconvLB, which incorporates ConvLora into SPAB to improve performance without increasing computation complexity. Specifically, given a pre-trained convolutional layer in SPAB, they update it by adding Lora layers, and representing it with a low-rank decomposition:
\begin{equation}
    W_{ConvLora} = W_{PT}+XY,
\end{equation}
where $W_{ConvLora}$ denotes the updated weight parameters of the convolution, $W_{PT}$ denotes the original pre-trained parameters of the convolution, $X$ is initialized by random Gaussian distribution, and $Y$ is zero in the beginning of training. Note that the Lora weights can be merged into the main backbone. Therefore, ConvLoras don't introduce extra computation during inference.

They adopt the pre-trained SPAN-Tiny model \cite{wan2024swift} with 26 channels. They replace the SPAB in SPAN with our proposed SconvLB, and also add ConvLora into the pixel shuffle block and the convolution before it. During training, they freeze the original weight and bias of the convolution and only update the Lora parameters. 

\noindent \textbf{Optimization.} To supervise the optimization of SconvLB, they adopt a knowledge-based distillation training strategy. They adopt spatial affinity-based knowledge distillation \cite{he2020fakd} to transfer second-order statistical info from the teacher model to the student model by aligning spatial feature affinity matrices at multiple layers of the networks. Given a feature $F_l \in R^{B\times C\times W\times H}$ extracted from the $l$-th layer of the network, they first flatten the tensor along the last two dimensions and calculate the affinity matrix $A_{spatial}$. Then the spatial feature affinity-based distillation loss can be formulated as:
\begin{equation}
    L_{AD}=\frac{1}{|A|}\sum_{l=1}^{n} ||A_l^S-A_l^T||_1,
\end{equation}
where $A_l^S$ and $A_l^T$ are the spatial affinity matrix of student and teacher networks extracted from the feature maps of the $l$-th layer, respectively. $|A|$ denotes the number of elements in the affinity matrix. Specifically, the team applys the distillation loss after each SconvLB. 

Except for the distillation loss in the feature space, the team applys a pixel-level distillation loss:
\begin{equation}
    L_{TS}=||\mathcal{T}(I_{LR})-\mathcal{S}(I_{LR})||_1,
\end{equation}
where $\mathcal{T}$ and $\mathcal{S}$ denote the teacher network and the student network, respectively. $I_{LR}$ denotes the LR image. 

They also apply the $L_2$ loss:
\begin{equation}
    L_{rec} = \| I_{HR} - \mathcal{S}(I_{LR}) \|_2^2,
\end{equation}
where $I_{HR}$ denotes the ground truth high-resolution image.
The overall loss is:
\begin{equation}
    L_{total} = \lambda_1L_{rec}+\lambda_2L_{TS}+\lambda_3L_{AD}.
\end{equation}

\noindent \textbf{Training Details.} The team uses DIV2K and LSDIR for training. Random flipping and random rotation are used for data augmentation. The training process is divided into two stages. 
\begin{enumerate}
    \item Stage One: HR patches of size $192 \times 192$ are randomly cropped from HR images, and the mini-batch size is set to 8. The model is trained by minimizing the $L_{total}$ mentioned above with the Adam optimizer. The learning rate is $1 \times 10^{-4}$. A total of $ 30k $ iterations are trained.
    \item Stage Two: In the second stage, the team increases the size of the HR image patches to $256 \times 256$, with other settings remaining the same as in the first stage.
\end{enumerate}

Throughout the entire training process, they employ an Exponential Moving Average (EMA) strategy to enhance the robustness of training.
\subsection{XiaomiMM}

\textbf{Method Details.} The team proposes an accelerated variant of the Swift Parameter-free Attention Network (SPAN)~\cite{wan2024swift}, called \textbf{SPANF}, which is built upon the fundamental SPAB block. To enhance the inference speed, SPANF introduces several key modifications compared to the original SPAN model. Firstly, they remove the last SPAB block, which reduces computational complexity without significantly impacting performance. Secondly, they increase the number of channels to 32, providing a better balance between model capacity and speed. Thirdly, they replace the first convolution layer with a nearest neighbor upsampling operation, which is computationally less intensive and accelerates the upsampling process. Lastly, they implement simple modifications to the shortcut connections within the network to further streamline computations. These changes collectively enable SPANF to achieve faster inference speeds while maintaining competitive image quality. The evaluations on multiple benchmarks demonstrate that SPANF not only upholds the efficiency of SPAN's parameter-free attention mechanism but also offers superior speed, making it highly suitable for real-world applications, particularly in scenarios with limited computational resources.

\begin{figure}[!t]
    \centering
    \includegraphics[width=\linewidth]{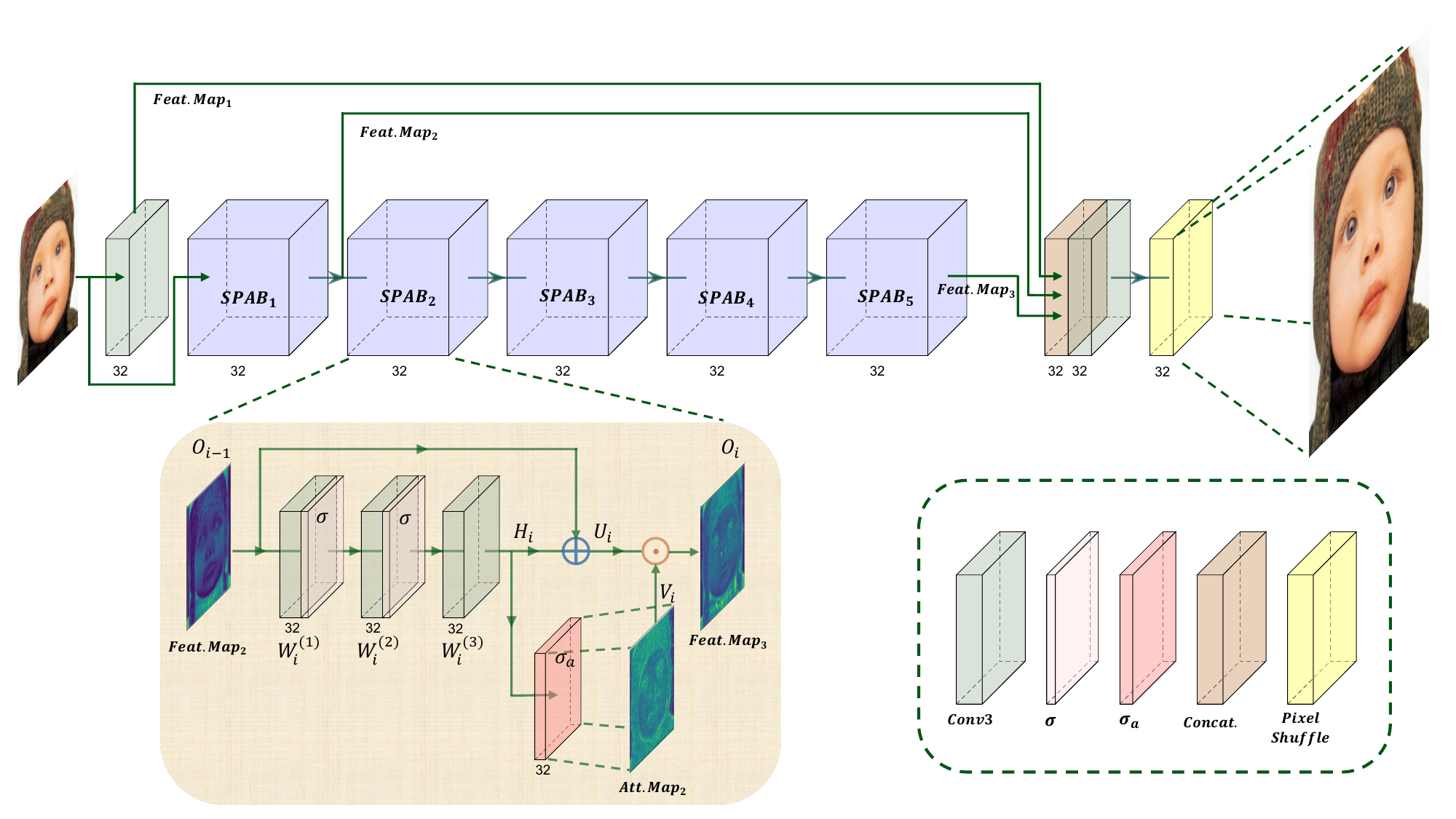}
    \caption{The proposed SPANF architecture. The main structure is basically the same as SPAN~\cite{wan2024swift}, but one SPAB module is reduced, and the number of channels is 32.} 
    \label{fig:team24_pipeline}
\end{figure}

\textbf{Implementation Details.} The dataset utilized for training comprises of DIV2K and LSDIR. During each training batch, 64 HR RGB patches are cropped, measuring $256 \times 256$, and subjected to random flipping and rotation. The learning rate is initialized at $5\times 10^{-4}$ and undergoes a halving process every $2\times10^5$ iterations. The network undergoes training for a total of $10^6$ iterations, with the L1 loss function being minimized through the utilization of the Adam optimizer~\cite{kingma2014adam}. They repeated the aforementioned training settings four times after loading the trained weights. Subsequently, fine-tuning is executed using the L1 and L2 loss functions, with an initial learning rate of $1\times10^{-5}$ for $5\times10^5$ iterations, and HR patch size of 512. They conducted finetuning on four models utilizing both L1 and L2 losses, and employed batch sizes of 64 and 128. Finally, they integrated these models' parameters to obtain our ultimate model.

\subsection{ShannonLab}
\begin{figure}
    \centering
    \includegraphics[width=1\linewidth]{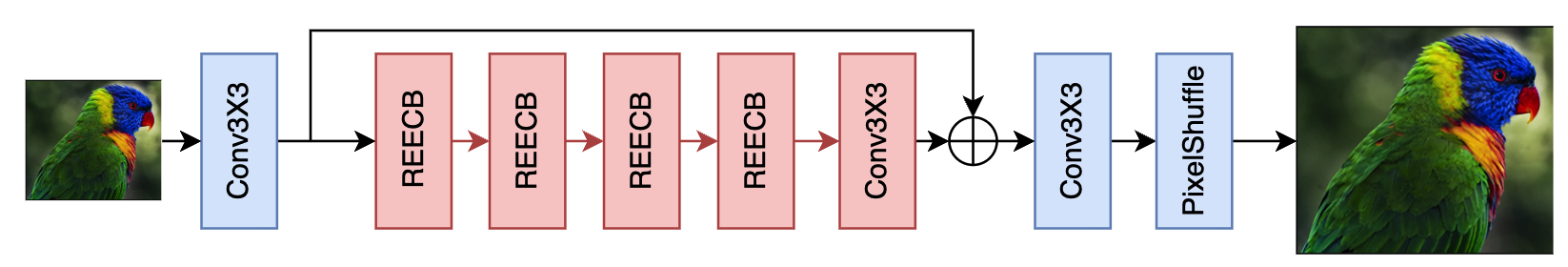}
    \caption{\textit{Team ShannonLab}: The pipeline of TSR.}
    \label{fig:TSR}
\end{figure}

\noindent \textbf{Method.} The method proposed by the team draws inspiration from ECBSR and SPAN. First, they optimized the ECB module by introducing a 1x1 convolutional layer for channel expansion before the input tensor enters the ECB module. After processing, another 1x1 convolution restores the original channel dimensions, while incorporating residual connections. During inference, these components can be merged into a standard 3x3 convolution through re-parameterization, thereby enhancing the ECB module's effectiveness without increasing computational overhead.As illustrated in Fig.\ref{fig:TSR}, The complete model architecture of TSR  comprises a shallow feature extraction convolution, a reconstruction convolution, a PixelShuffle module, and four REECB block which made of stacked optimized ECB.

\noindent \textbf{Training Details.} The model is trained on the DIV2K and LSDIR train dataset with random flipping and rotation applied for data augmentation. The Adam optimizer is consistently employed throughout the training process. The entire training process is divided into five steps.

1. HR patches of size 256×256 are randomly cropped from HR images, and the mini-batch size is set to 32. L1 loss is used and the initial learning rate is set to 5e-4, with a cosine learning rate decay strategy. The total iterations is 500k.

2. HR patches of size 256×256 are randomly cropped from HR images, and the mini-batch size is set to 32. L1 and L2 loss is used and the initial learning rate is set to 5e-4, with a cosine learning rate decay strategy. The total iterations is 1000k.

3. HR patches of size 512×512 are randomly cropped from HR images, and the mini-batch size is set to 64. L2 loss is used and the initial learning rate is set to 2e-4, with a cosine learning rate decay strategy. The total iterations is 1000k.

4. HR patches of size 512×512 are randomly cropped from HR images, and the mini-batch size is set to 64. L2 loss is used and the initial learning rate is set to 1e-4, with a cosine learning rate decay strategy. The total iterations is 1000k.

5. HR patches of size 512×512 are randomly cropped from HR images, and the mini-batch size is set to 64. L2 loss is used and the initial learning rate is set to 1e-5, with a cosine learning rate decay strategy. The total iterations is 1000k.

% ---- 1st - 3rd of the Sub-Track: Runtime
% 1st Sub-Track: Runtime
% \input{teams/team31_ShannonLab/main}
% 2nd Sub-Track: Runtime
\subsection{TSSR}
\textbf{Method.} They combined the ideas of reparameterization and attention mechanism to design a model that can capture image information in the network and effectively achieve image super-resolution.

\noindent
\textbf{Training Details.} The training process is divided into three steps.

1. HR patches of size 256×256 are randomly cropped from HR images, and the mini-batch size is set to 64. L1 loss with AdamW optimizer is used and the initial learning rate is set to 0.0005  and halved at every 100k iterations. The total iterations is 500k. 

2. HR patches of size 256×256 are randomly cropped from HR images, and the mini-batch size is set to 64. L1 and L2 loss with AdamW optimizer is used and the initial learning rate is set to 0.0002  and halved at every 100k iterations. The total iterations is 1000k. 

3. HR patches of size 512×512 are randomly cropped from HR images, and the mini-batch size is set to 64. L2 loss with AdamW optimizer is used and the initial learning rate is set to 0.0001  and halved at every 100k iterations. The total iterations is 1000k. 
% 3rd Sub-Track: Runtime
\subsection{mbga}

\noindent
\textbf{Architecture.}
The team proposes the ESPAN, which is based on SPAN~\cite{span}. Through evaluations of depth-channel combinations in SPAN on an A6000 GPU, they determined that setting the number of channels to 32 yields higher efficiency than 28 channels. To reduce parameters and FLOPs, a depth of 6 was adopted. Additionally, a 9×9 convolution replaced the conventional 3×3 convolution at the network's input stage since they find that 9×9 convolution is faster than 3x3 convolution on A6000.

\begin{figure}
    \centering
    \includegraphics[width=0.3\textwidth]{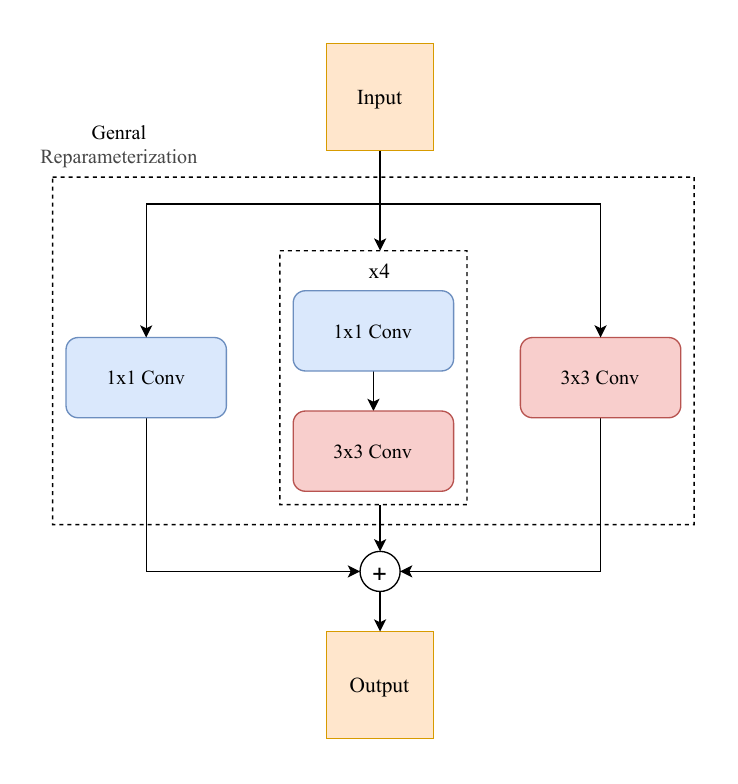}
    \caption{\textit{Team mbga:} General Reparameterization.} 
    \label{rep}
\end{figure}

\begin{figure}
    \centering
    \includegraphics[width=1.0\linewidth]{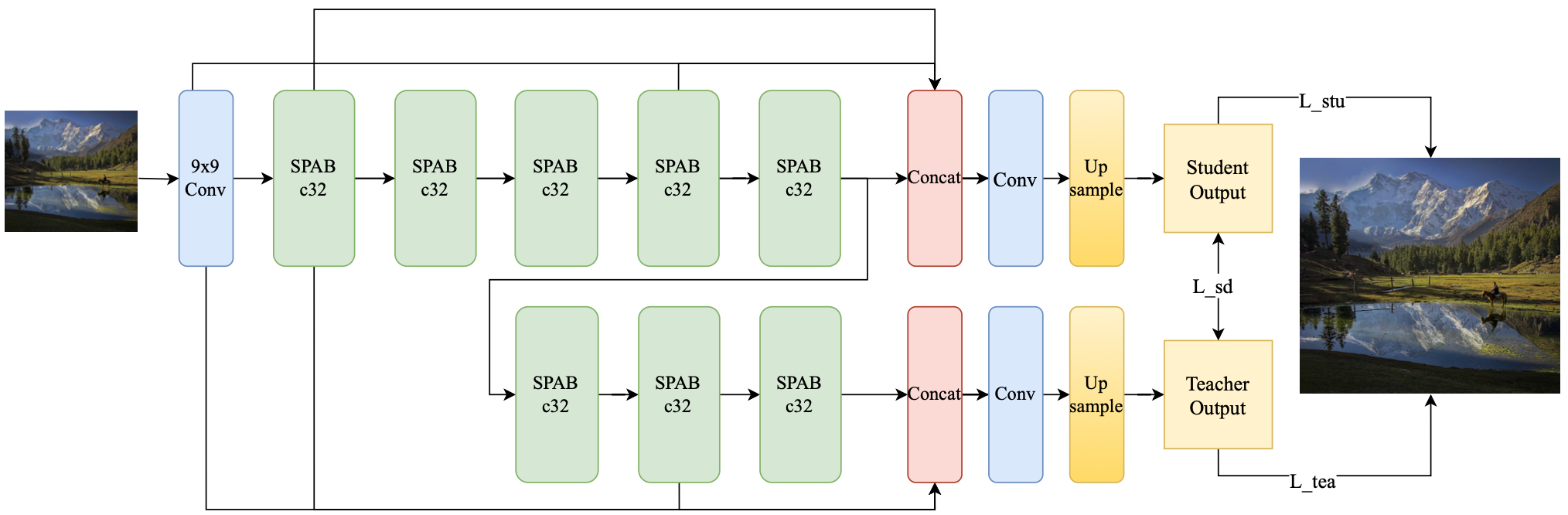}
    \caption{\textit{Team mbga:} ESPAN with self distillation.} 
    \label{selfdistill} 
\end{figure}

\noindent
\textbf{General Reparameterization.} Inspired by MobileOne~\cite{mobileone} and RepVGG~\cite{repvgg}, the team proposes a generalized reparameterization block(~\cref{rep}). The block consists of four 1×1-3×3 convolution branches, one  1×1 convolution branch, and one 3×3 convolution branch. Skip connections are omitted due to empirical observations of training instability. While additional duplicated branches or 3×3-1×1 convolution branches are feasible, the current configuration is found to offer superior performance consistency during optimization.

\noindent
\textbf{Self distillation and progressive learning.} Inspired by RIFE~\cite{rife}, self-distillation is incorporated into their training pipeline. The teacher model shares the identical backbone as the student model but includes three extra SPAB blocks appended to the student's backbone(~\cref{selfdistill}). A self-distillation loss similar to RIFE's formulation is adopted to co-train the teacher and student networks. This design enables the teacher model to learn robust backbone features. After the distillation phase, the student loss and distillation loss components are removed, and the entire teacher model is fine-tuned. Leveraging the pre-trained robust teacher, progressive learning is employed: the extra SPAB blocks are gradually removed from the teacher's backbone, finally resulting in an architecture identical to the original student model. 

\noindent
\textbf{Frequency-Aware Loss}. Since small models have limited parameters, during training, they should make the model focus more on important (or difficult) areas. In their methods, two types of frequency-aware losses are employed. The first type is the DCT loss. They use the discrete cosine transform (DCT) to convert the RGB domain to the frequency domain and then apply the L1 loss to calculate the difference. The other type is the edge loss. They add a blur to the image and then subtract the blurred image from the original one to obtain the high frequency area. Subsequently, the L1 loss is calculated on this high frequency area. 

\noindent
\textbf{Training details}: The training process contains two stages. And the training dataset is the DIV2K\_LSDIR\_train. General reparameterization is used on the whole process.

I. At the first stage, they use self distillation to train the teacher model.

\begin{itemize}
    \item Step1. The team first trains a 2x super-resolution model. HR patches of size 256x256 are randomly cropped from HR images, and the mini-batch size is set to 64. L1 loss and self distillation loss with AdamW optimizer are used and the initial learning rate is set to 0.0001 and halved at every 100k iterations. The total iterations is 500k. This step is repeated twice. And then they follow the same training setting and use 2x super-resolution model as pretrained model to train a 4x super-resolution model. This step is repeated twice.

    \item Step2. HR patches of size 512x512 are randomly cropped from HR images, and the mini-batch size is set to 16. MSE loss, frequency-aware loss and self distillation loss with AdamW optimizer are used and the initial learning rate is set to 0.0001 and halved at every 100k iterations. The total iterations is 500k. This step is also repeated twice.

    \item Step3. They only train the teacher model. HR patches of size 512x512 are randomly cropped from HR images, and the mini-batch size is set to 16. MSE loss and frequency-aware loss with AdamW optimizer are used and the initial learning rate is set to 0.00005 and halved at every 100k iterations. The total iterations is 500k. This step is also repeated twice.
\end{itemize}

II. At the second stage, they use progressive learning to get the final student model. 

\begin{itemize}
    \item Step4. They drop the additional SPAB block one by one. HR patches of size 512x512 are randomly cropped from HR images, and the mini-batch size is set to 16. L1 loss with AdamW optimizer are used and the initial learning rate is set to 0.0001 and halved at every 100k iterations. The total iterations is 500k.

    \item Step5. They repeat the following training process many times until convergence. HR patches of size 512x512 are randomly cropped from HR images, and the mini-batch size is set to 16. MSE loss and frequency-aware loss with AdamW optimizer are used and the initial learning rate is set to 0.00005 and halved at every 100k iterations. The total iterations is 500k. 
\end{itemize}

% ---- 1st - 3rd of the Sub-Track: FLOPs
% 1st Sub-Track: FLOPs
\subsection{VPEG\_C}
\begin{figure*}
	\centering
	\includegraphics[width=0.99\textwidth]{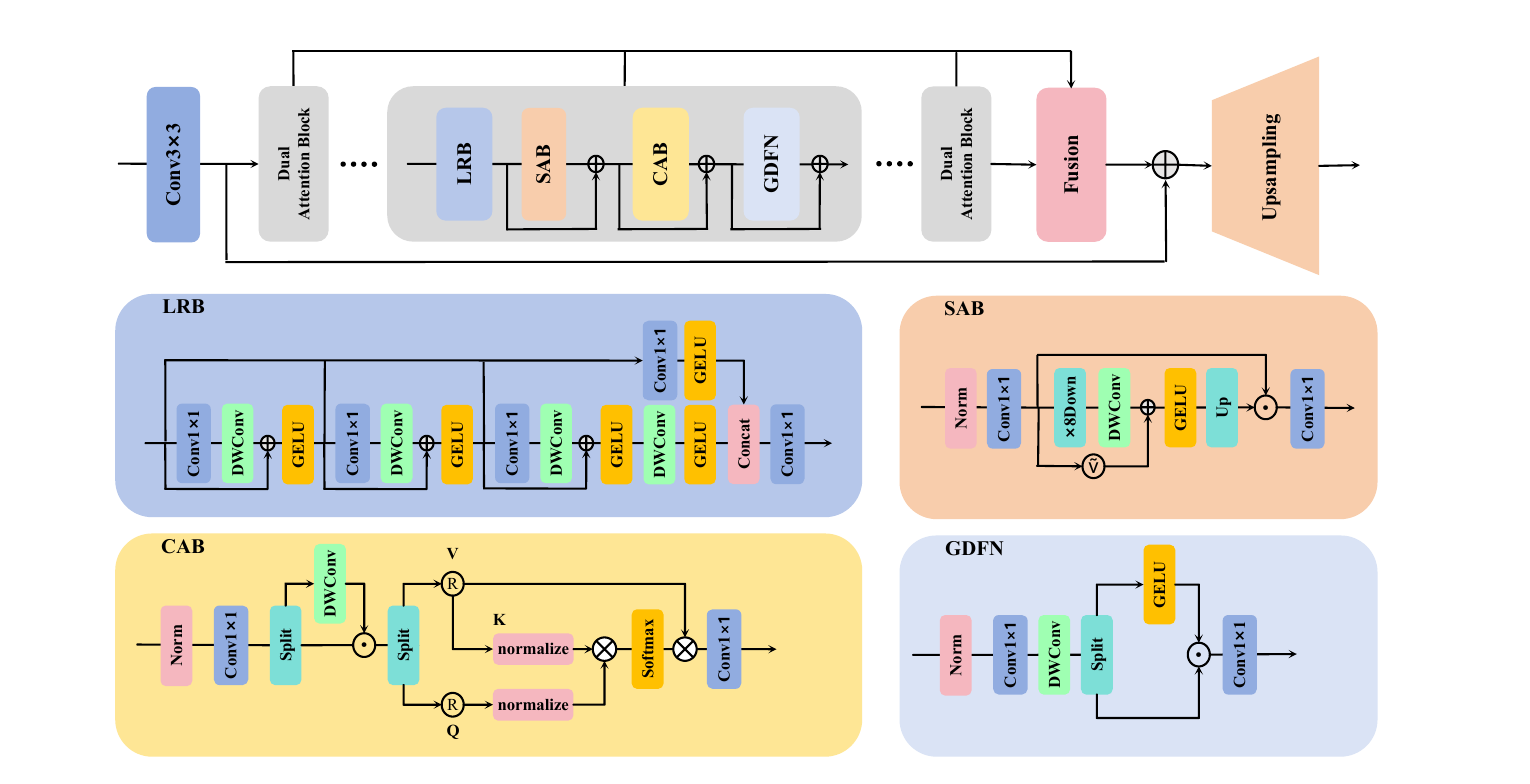}
	\caption{\textit{Team VPEG\_C}: An overview of the DAN.}  
	\label{fig:team20_framework}
\end{figure*}

\noindent
\textbf{General Method Description.}
As illustrated in ~\cref{fig:team20_framework}, they propose a Dual Attention Network (DAN) for the light-weight single-image super-resolution task. 
The core components of DAN consist of three parts: a Local Residual Block (LRB), a Spatial Attention Block (SAB), and a Channel Attention Block (CAB).

\noindent\textbf{Local Residual Block (LRB).} They leverage the $1\times1$ convolution layers followed by a $3\times3$ depthwise convolution as the basic unit, repeated three times. Specially, GELU activation is applied on each layers, and the features are passed in a densely connected manner. At the end of the block, feature maps from different levels are aggregated using channel concatenation, effectively capturing local image details.

\noindent\textbf{Spatial Attention Block (SAB).} They adopt the spatial attention design of SMFANet~\cite{smfanet}, which employs a variance-constrained feature modulation mechanism to aggregate spatial feature. This allows efficient spatial interaction with minimal computational cost.

\noindent\textbf{Channel Attention Block (CAB).} Global channel-wise information is modeled through a self-gating mechanism that enhances local representations and increases model non-linearity. This is followed by a key-value shared MDTA~\cite{Restormer} for global interaction and a GDFN~\cite{Restormer} for feature refinement.

\noindent
\textbf{Training Description.}
The proposed DAN consists of 6 feature mixing modules with 16 channels. The training process is divided into two stages:

\begin{itemize} 
\item[1.] \textbf{Pre-training Stage:} They pre-train DAN using 800 images from the DIV2K~\cite{NTIRE_2017} and the first 10K images of the LSDIR~\cite{li2023lsdir} datasets. The cropped LR image size is $72\times72$, and the mini-batch size is set to 64. The DAN is trained by minimizing L1 loss and the frequency loss\cite{MIMO} with Adam optimizer for total 800, 000 iterations. The initial learning rate is set to 2e-3 and halved at 200K, 400K, 600K, 700K.

\item[2.] \textbf{Fine-tuning Stage:} They fine-tune the model on the 800 images of DIV2K~\cite{NTIRE_2017} and the first 10K images of the  LSDIR~\cite{li2023lsdir} datasets. The cropped LR image size is $72\times72$, and the mini-batch size is set to 64. The DAN is trained by minimizing PSNR loss with the Adam optimizer for total 200, 000 iterations. They set the initial learning rate to 5e-4 and halve it at 50K, 100K, 150K, and 175 K.

\end{itemize}
% 2nd Sub-Track: FLOPs
\subsection{XUPTBoys}
\begin{figure*}[!tb]
    \centering
    \includegraphics[width=1.0\textwidth]{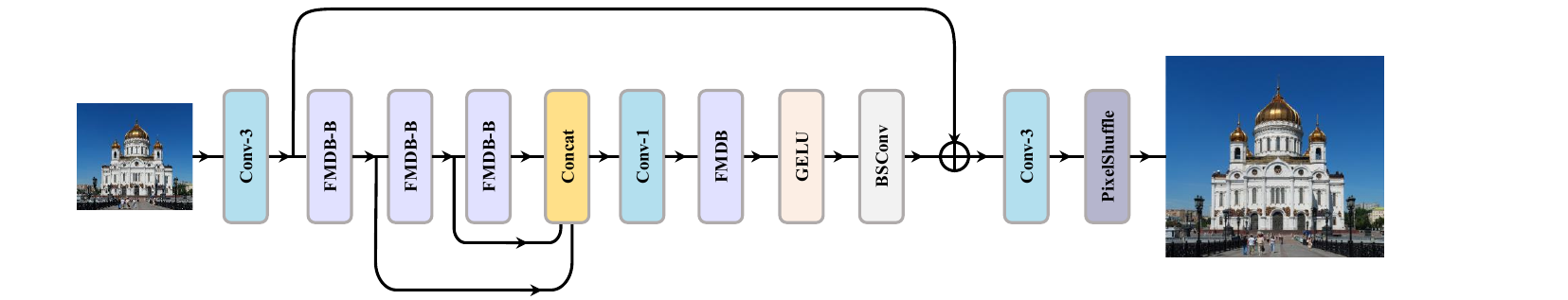}
    \caption{\textit{Team XUPTBoys}: The whole framework of Frequency-Guided Multi-level Dispersion Network (FMDN).}
    \label{fig:team26_framework}
\end{figure*}

\begin{figure*}[!tb]
    \centering
    \includegraphics[width=1.0\textwidth]{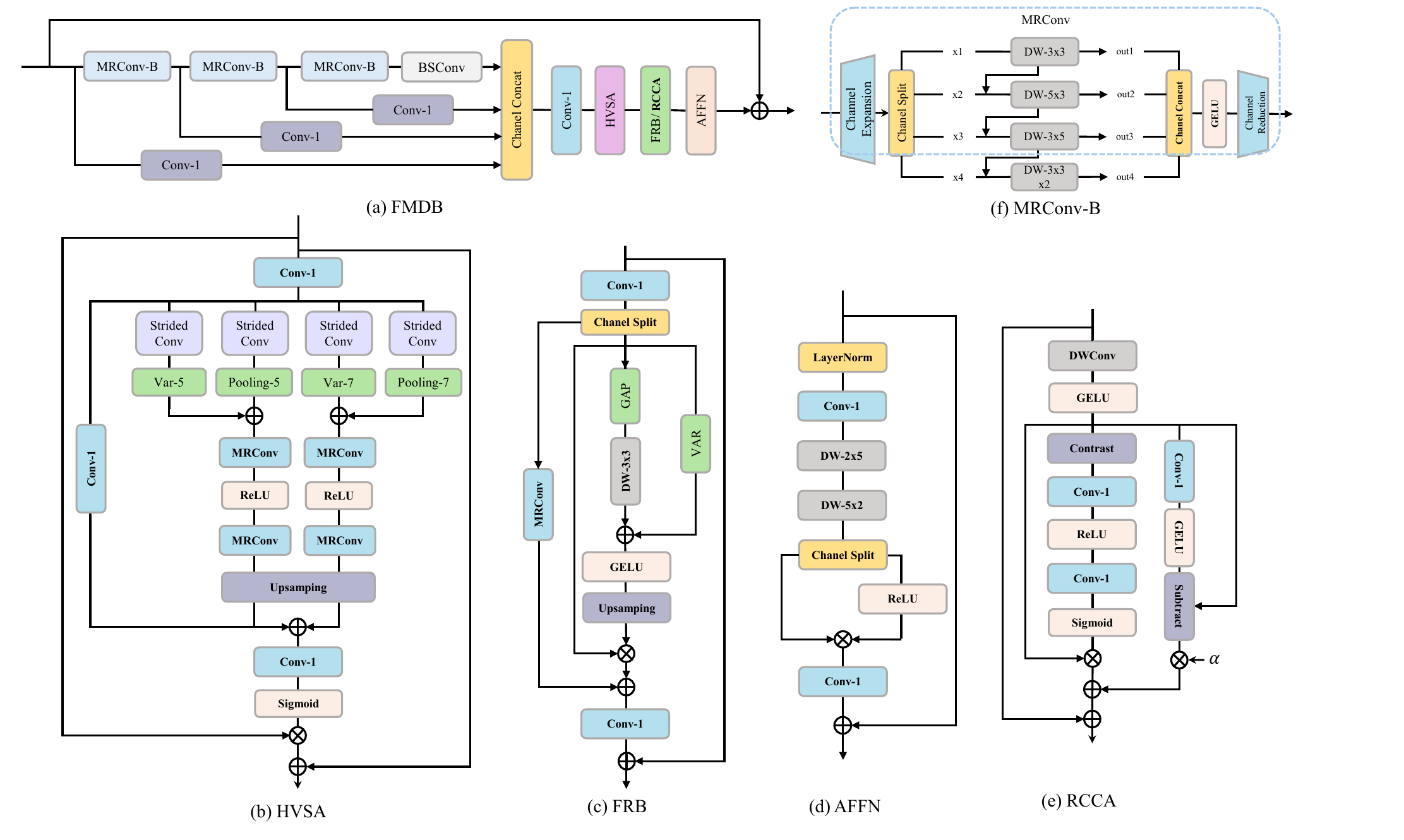}
    \caption{\textit{Team XUPTBoys}: The details of each component. 
    (a) FMDB: Frequency-Guided Multi-level Dispersion Block; (b) HVSA: Hierarchical Variance-guided Spatial Attention; (c) FRB: Frequency-Guided Residual Block; 
    (d) AFFN: Asymmetric FeedForward Network; 
    (e) RCCA: Reallocated Contrast-aware Channel Attention; (f) MRConv-B/MRConv: Multilevel Residual Convolution Basic and Multilevel Residual Convolution  }
    \label{fig:team26_details}
\end{figure*}

\noindent \textbf{General Method Description.}
The XUPTBoys team proposed the Frequency-Guided Multilevel Dispersion Network (FMDN), as shown in Fig.~\ref{fig:team26_framework}.FMDN adopts a similar basic framework to \cite{IMDN, RFDN, BSRN, MDRN}.

Based on the above analysis, they propose the new Frequency-Guided Multi-level Dispersion Block(FMDB) and the new Frequency-Guided Multi-level Dispersion Block Basic(FMDB-B) as the base block of FMDN. As shown in Fig.~\ref{fig:team26_details} they use Hierarchical Variance-guided Spatial Attention(HVSA), Reallocated Contrast-Aware Channel Attention (RCCA) as alternatives to Enhanced Spatial Attention (ESA)~\cite{ESA} and Contrast-Aware Channel Attention (CCA)~\cite{CCA}, Frequency-Guided Residual block (FRB), Asymmetric FeedForward Network (AFFN), Multilevel Residual Convolution (MRConv) and Multilevel Residual Convolution Basic(MRConv-B). The difference between FMDB and FMDB-B is that the former uses MRConv, while the latter uses MRConv-B.

In HVSA, the effects of multilevel branching and local variance on performance are examined. Small-window multilevel branches fail to capture sufficient information, while local variance within a single branch can create significant weight disparities. To address these issues, \cite{MDRN} was enhanced to introduce the D5 and D7 branches, which effectively utilize local variance to capture information-rich regions while balancing performance and complexity. In RCCA, this approach improves the traditional channel attention mechanism by not only reallocating weights across channels but also better managing shared information among them. Introduces complementary branches with 1×1 convolutions and GELU activation functions, which help redistribute complementary information, improving the uniqueness of each channel. In FRB, it enhances feature representation using convolutional layers and GELU activation. It normalizes input, extracts features with depthwise convolutions of different kernel sizes, and combines them through residual connections to preserve spatial information for effective image processing. In AFFN, it applies layer normalization and a 1x1 convolution to expand feature dimensions. It then uses two depthwise convolutions with different kernel sizes, combines the results with GELU activation, and projects the output back to the original dimension with a residual connection. In MRConv and MRConv-B, MRConv and MRConv-B use convolution kernels of different sizes for parallel convolution, and finally activate the output using GELU and combine it with residual connections, effectively preserving spatial information.

\noindent
\textbf{Training Description.}
The proposed FMDN has 3 FMDB-Basic blocks and 1 FMDB block, in which the number of feature channels is set to 24. The details of the training steps are as follows:
\begin{enumerate}
    \item Pretraining on the DIV2K~\cite{timofte2017ntire} and and Flickr2K~\cite{lim2017enhanced}. HR patches of size 256 × 256 are randomly cropped from HR images, and the mini-batch size is set to 64. The model is trained by minimizing the L1 loss function~\cite{l2loss} with the Adam optimizer~\cite{Adam}. The initial learning rate is set to $2 \times {10^{ - 3}}$ and halved at $\left\{ {100k, 500k, 800k,900k,950k} \right\}$-iteration. The total number of iterations is 1000k.
    \item Finetuning on 800 images of DIV2K and the first 10k images of LSDIR~\cite{li2023lsdir}. HR patch size and  mini-batch size are set to 384 × 384 and 64, respectively. The model is fine-tuned by minimizing L2 loss function~\cite{l2loss}. The initial learning rate is set to $5 \times {10^{ - 4}}$ and halved at $\left\{ {500k} \right\}$-iteration. The total number of iterations is 1000k.
\end{enumerate}
% 3rd Sub-Track: FLOPs
\subsection{HannahSR}

\textbf{General Method Description.}
\begin{figure*}[!ht]
    \centering
    \includegraphics[width=1.0\linewidth]{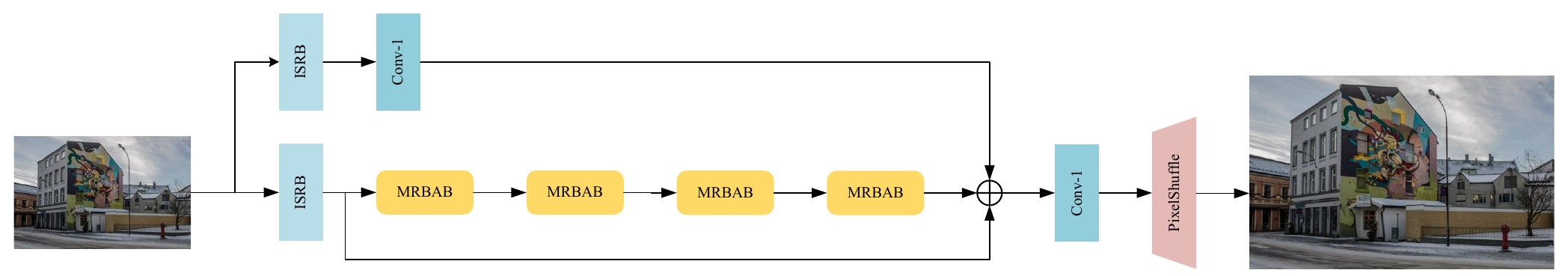}
    \caption{\textit{Team HannahSR:} The overall architecture of Multi-level Refinement and Bias-learnable Attention Dual Branch Network (MRBAN).}
    \label{hannahsr_fig1}
\end{figure*}
The architecture of the proposed network is depicted in \cref{hannahsr_fig1}, which is inspired by previous studies such as AGDN~\cite{10678094}, MDRN~\cite{10208430} and SPAN~\cite{10678610}. They propose a Multi-level Refinement and Bias-learnable Attention dual branch Network (MRBAN). More specifically, they build upon the AGDN framework by constructing another branch consisting of one $3\times3$ convolution layer (ISRB) and one $1\times1$ convolution layer to enhance the overall performance in a learnable way. Meanwhile, they replace the concat module in the AGDN with a direct element-wise summation, for the sake of harvesting significant savings of the parameters.

\begin{figure*}[!ht]
    \centering
    \begin{subfigure}[t]{0.85\textwidth}
        \vspace{0pt}
        \centering
        \includegraphics[width=\textwidth]{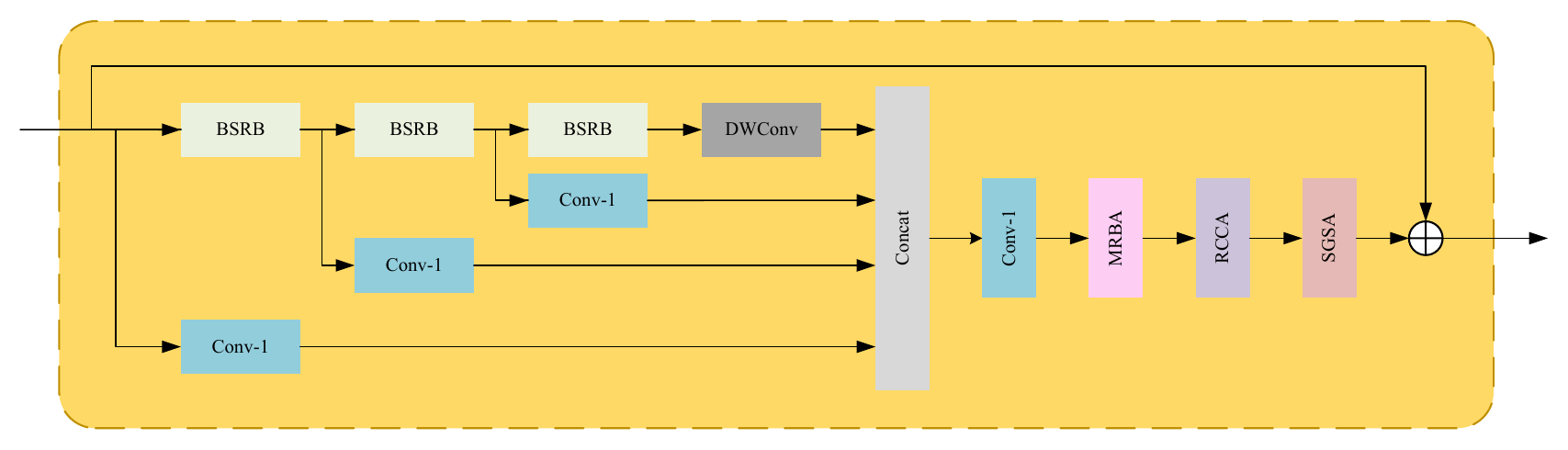}
        \caption{\textit{Team HannahSR:} The MRBAB architecture.}
        \label{hannahsr_fig2a}
    \end{subfigure}

    \vspace{1em} 

    \begin{subfigure}[t]{0.7\textwidth}
        \vspace{0pt}
        \centering
        \includegraphics[width=\textwidth]{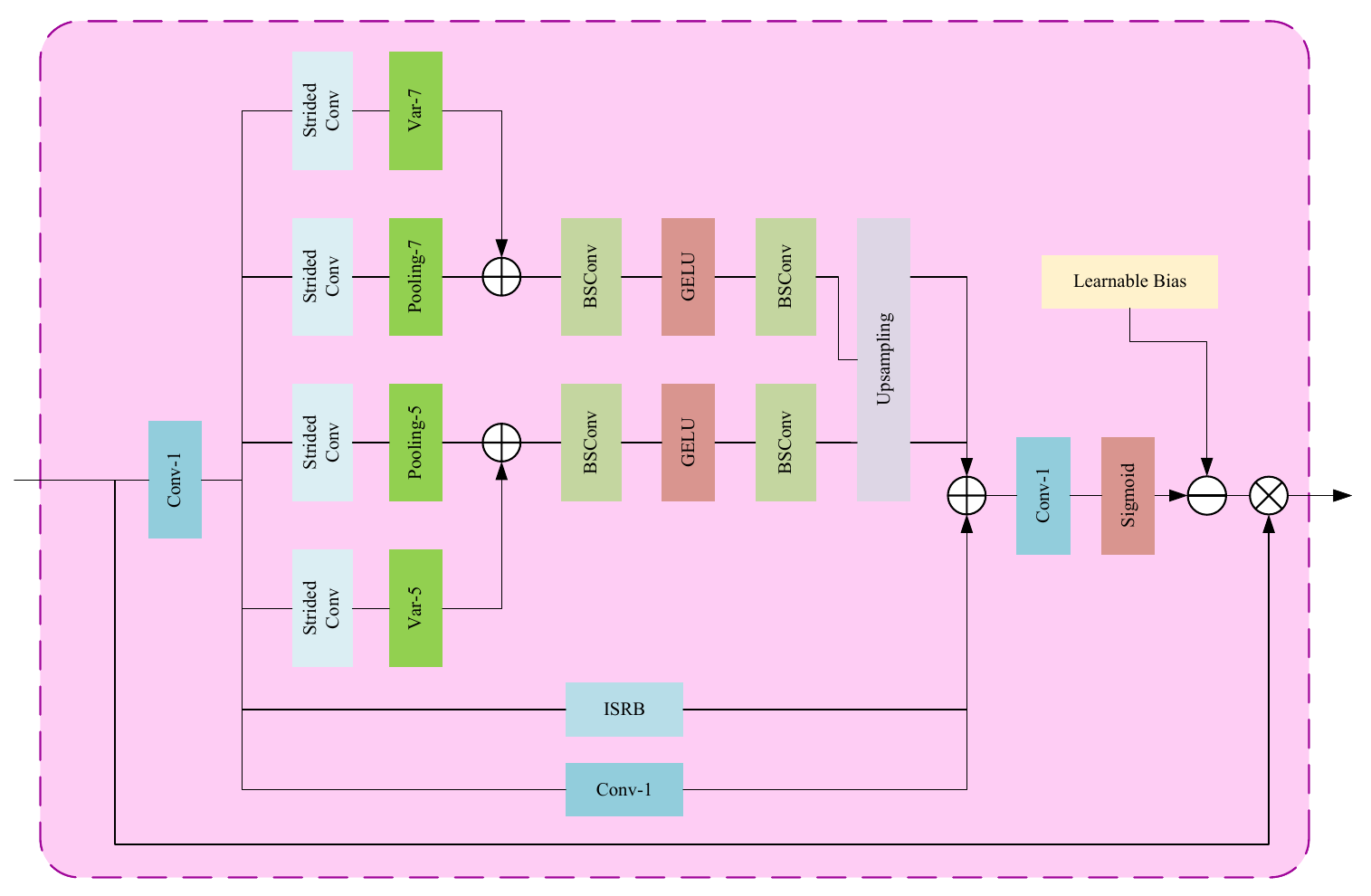}
        \caption{\textit{Team HannahSR:} The MRBA architecture.}
        \label{hannahsr_fig2b}
    \end{subfigure}
    \quad
    \vspace{0.5em} 
    \begin{subfigure}[t]{0.15\textwidth}
        \vspace{0pt}
        \centering
        \includegraphics[width=\textwidth]{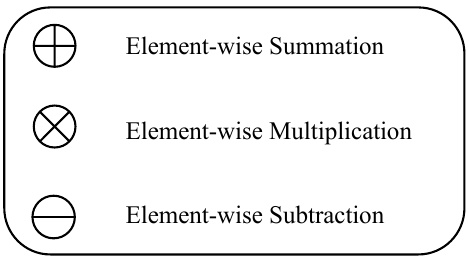}
        \label{hannahsr_fig2c}
    \end{subfigure}
    \caption{\textit{Team HannahSR:} The detailed architecture of the network MRBAN. (a) MRBAB: Multi-level Refinement and Bias-learnable Attention Block; (b) MRBA: Multi-level Refinement and Bias-learnable Attention; Other components: BSRB: Blueprint Shallow Residual Block~\cite{9857155}; BSConv: Blueprint Separable Convolution~\cite{9857155}; RCCA: Reallocated Contrast-aware Channel Attention~\cite{10678094}; SGSA: Sparse Global Self-attention~\cite{10678094}.}
    \label{hannahsr_fig2}
\end{figure*}

In addition, they propose the multi-level refinement and bias-learnable attention block (MRBAB) as the basic block of our network. As described in Figure \ref{hannahsr_fig2}, they attempt to minimize the information loss induced by Sigmoid module. When confronted with a negative input with a large absolute value, the output of the Sigmoid module will be approximately equal to zero, which results in remarkable information loss. To address this issue, SPAN~\cite{10678610} used an origin-symmetric activation function. They added a bias of $-0.5$ to the Sigmoid function, which allowed the information carried by negative inputs to be taken into account. However, when dealing with the larger positive inputs, their outputs would be approximately equal to 0.5. When compared with the original 1.0, they inevitably suffered from significant information loss. To tackle this issue, they set the negative bias as a learnable parameter so that it can be updated dynamically during the training process to optimally boost the accuracy performance.

Eventually, they adopt the reparameterization technique. They replace the first $3\times3$ convolution layer with identical scale reparameterization block to extract richer local features for supplying the following layers with more valuable information, while standardizing the number of channels to an identical scale for lightweight super resolution networks to prevent incurring inappropriate model capacity increments.

\noindent \textbf{Training Strategy.} The proposed MRBAN consists of 4 MRBAB, and the feature channel is set to 32. They adopt a four-step training strategy. The details of the training steps are as follows:
\begin{itemize}
    \item[1.] Pretraining on the DIV2K~\cite{8014884} and Flickr2K~\cite{8014885} datasets with the patch size of $256\times 256$ and the mini-batch size is set to 64. The MRBAN is trained by minimizing the L1 loss function with the Adam optimizer. The initial learning rate is set to $3 \times 10^{-3}$ and halved at \{100k, 500k, 800k, 900k, 950k\}-iteration. The number of iterations is 1000k.
    \item[2.] Initial fine-tuning on DIV2K and the first 10K images of LSDIR~\cite{li2023lsdir}. The patch size is $384\times384$ and the mini-batch size is set to 32. The model is trained by minimizing the MSE loss function. The initial learning rate is set to $1.5 \times 10^{-3}$ and halved at \{100k, 500k, 800k, 900k, 950k\}-iteration. The number of iterations is 1000k.
    \item[3.] Advanced training on the DIV2K and the whole LSDIR datasets.  The patch size is $384\times384$ and the mini-batch size is set to 64. The model is trained by minimizing the MSE loss function. The initial learning rate is set to $8 \times 10^{-4}$ and halved at \{100k, 500k, 800k, 900k, 950k\}-iteration. The number of iterations is 1000k. This stage can be repeated twice.
    \item[4.] Final fine-tuning on the DIV2K and the whole LSDIR datasets.  The patch size is $448\times448$ and the mini-batch size is set to 128. The model is trained by minimizing the MSE loss function. The initial learning rate is set to $5 \times 10^{-6}$ and halved at \{100k, 500k, 800k, 900k, 950k\}-iteration. The number of iterations is 1000k.
\end{itemize}

% ---- 1st - 3rd of the Sub-Track: Params
% 1st Sub-Track: Params
% \input{teams/team20_VPEG_C/main}
% 2nd Sub-Track: Params
% \input{teams/team13_HannahSR/main}
% 3rd Sub-Track: Params
% \input{teams/team26_XUPTBoys/main}

% --- Other Ranked Teams via Main Track order:
% 4th of Main Track
% \input{teams/team58_TSSR/main}    % Incomplete
% 5th of Main Track
\subsection{Davinci}

\textbf{Final Solution Description.}
They chose the Swift Parameter-free Attention Network~\cite{wan2024swift} as their base model, the winner of the NTIRE2024 ESR track. After trying the evolution pipeline mentioned in SwinFIR~\cite{swinfir}, the content decoupling strategy proposed in CoDe~\cite{gu2024code}, the pre-training fine-tuning paradigm, and the model compression techniques such as model pruning and knowledge distillation discussed in Ref~\cite{jiang2024compressing} respectively, they employ the model \textbf{P}runing of the \textbf{la}st la\textbf{yer} with $l_2$ norm of the baseline and introducing the mixup \textbf{Aug}mentation as their final proposal to preserve the original parameter distributions as much as possible, termed \textbf{PlayerAug}.

\noindent \textbf{Training Details.} 
After pruning the SPAN, they train it on the DIV2K\_LSDIR mixed training set, cropping the patch size to 512. The random rotation and flip are configured for data augmentation. The Adam~\cite{kingma2014adam} optimizer with $\beta_1 = 0.9$ and $\beta_2 = 0.99$ and the L1 loss function are adopted to optimize the models, and the mini-batch size is set to 32. All the experiments are conducted on 8 L40S GPUs.
% 6th of Main Track
% \input{teams/team46_SRCB/main}        % Incomplete
% 7th of Main Track
\subsection{Rochester}

\begin{figure}[!tb]
\centering
\includegraphics[width=\linewidth]{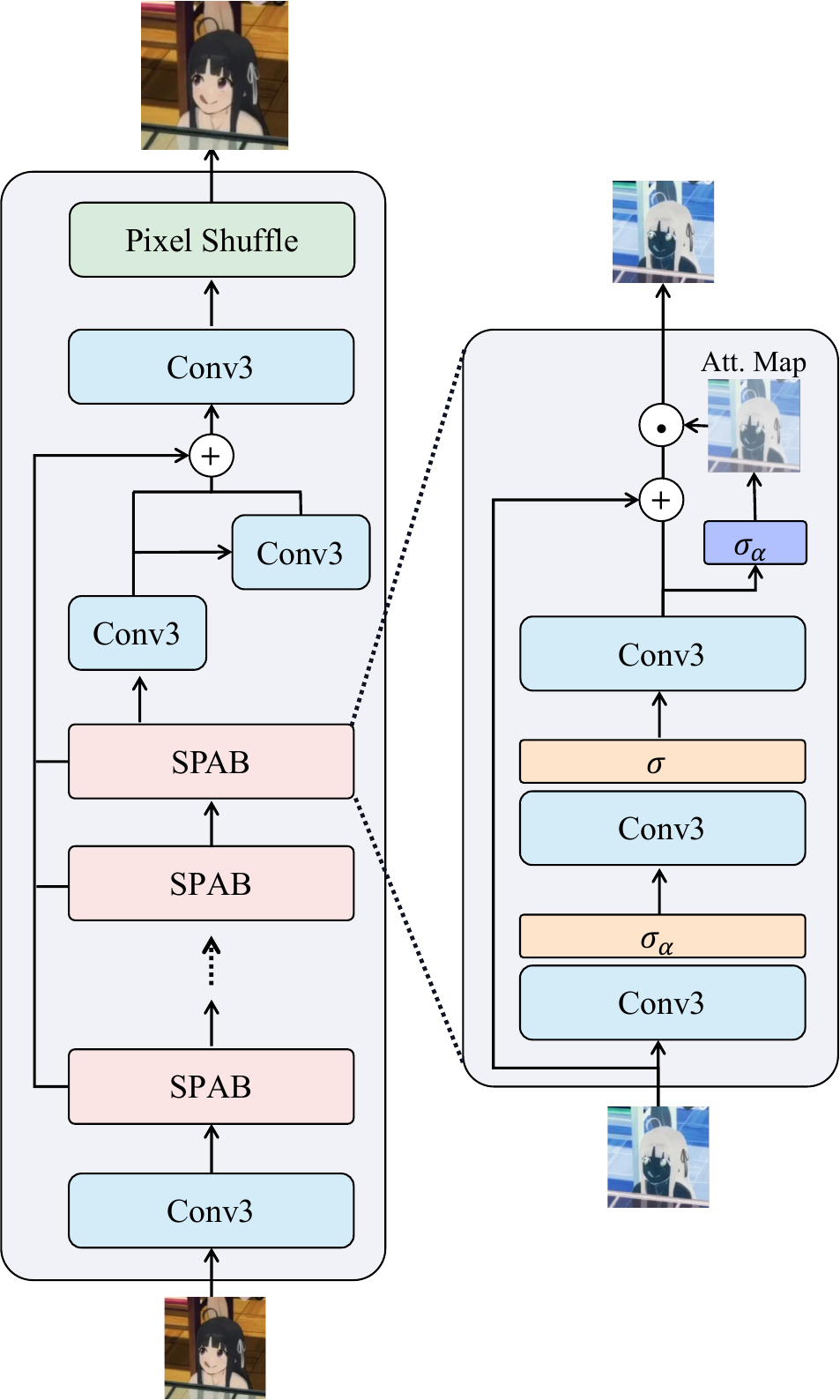}
\vspace{-0.5cm}
\caption{\textit{Team Rochester}: They reduce the channel dimension from 48 to 28 from the original design and introduce additional convolution to stabilize the attention feature maps from SPAB blocks. Example input and output are adapted from \cite{tang2025ai4anime}.}
 \vspace{-0.5cm}
\label{fig:model-team38}
\end{figure}

\paragraph{Method Details.} The proposed method, \textbf{ESRNet}, is an improved and more efficient variant of last year’s XiaomiMM SPAN network~\cite{wan2024swift}. The original SPAN network demonstrated strong generation quality but required complex training tricks and model fusion strategies, making it difficult to reproduce and computationally expensive. In contrast, ESRNet achieves similar performance with significantly reduced computational overhead, enhanced training stability, and improved inference speed.

\noindent \paragraph{Model Architecture.}
A key aspect of ESRNet’s design is its ability to maintain high performance while reducing computational costs. As shown in Fig.~\ref{fig:model-team38}, their modifications include:

\begin{itemize}
    \item Retaining the first six SPAN attention blocks as core feature extraction components while introducing a lightweight convolutional layer to refine the extracted feature maps before fusing them with the original features. This modification enhances feature representation while stabilizing the training process.
    \item Reducing the number of feature channels from 48 to 26, leading to a substantial decrease in both model parameters and floating-point operations (FLOPs). This reduction not only lowers GPU memory consumption but also improves inference efficiency without degrading performance.
    \item Improved validation speed, as ESRNet requires fewer computations per forward pass, making it more suitable for real-time applications compared with the baseline method.
\end{itemize}

Overall, ESRNet has approximately half the number of parameters and FLOPs compared to the baseline EFPN network, yet it maintains a high PSNR score, demonstrating that their modifications achieve an excellent trade-off between efficiency and performance.

\paragraph{Training Methodology.}
They train ESRNet on RGB image patches of size 256×256, applying standard augmentation techniques such as random flipping and rotation to enhance generalization. To ensure stable convergence and optimal performance, they adopt a three-stage training strategy:

\begin{enumerate}
    \item \textbf{Initial Feature Learning:} They train the model with a batch size of 64 using Charbonnier loss, a robust loss function that mitigates the effects of outliers. The Adam optimizer is used with an initial learning rate of \(2 \times 10^{-4}\), which follows a cosine decay schedule.
    \item \textbf{Refinement Stage:} They progressively decrease the learning rate linearly from \(2 \times 10^{-4}\) to \(2 \times 10^{-5}\), allowing the model to refine its learned features while maintaining stable gradients.
    \item \textbf{Fine-Tuning with L2 Loss:} In the final stage, they adopt L2 loss to fine-tune the model, further enhancing detail restoration. The learning rate is further reduced from \(2 \times 10^{-5}\) to \(1 \times 10^{-6}\) for smooth convergence.
\end{enumerate}

By structuring the training into these stages, they eliminate the need for complex training tricks used in previous approaches while achieving more stable and reliable optimization.

One of the most significant advantages of ESRNet is its improved validation time due to its optimized architecture. Compared to the original SPAN network, ESRNet achieves a similar PSNR score while reducing computational complexity. The model requires significantly fewer FLOPs and parameters, leading to a noticeable reduction in inference time and GPU memory usage. This makes ESRNet a practical solution for applications requiring both high-quality generation and efficient computation.
% 8th of Main Track
% \input{teams/team36_mbga/main}
% 9th of Main Track
\subsection{IESR}

\begin{figure*}[!ht]
\centering
\includegraphics[width=\textwidth]{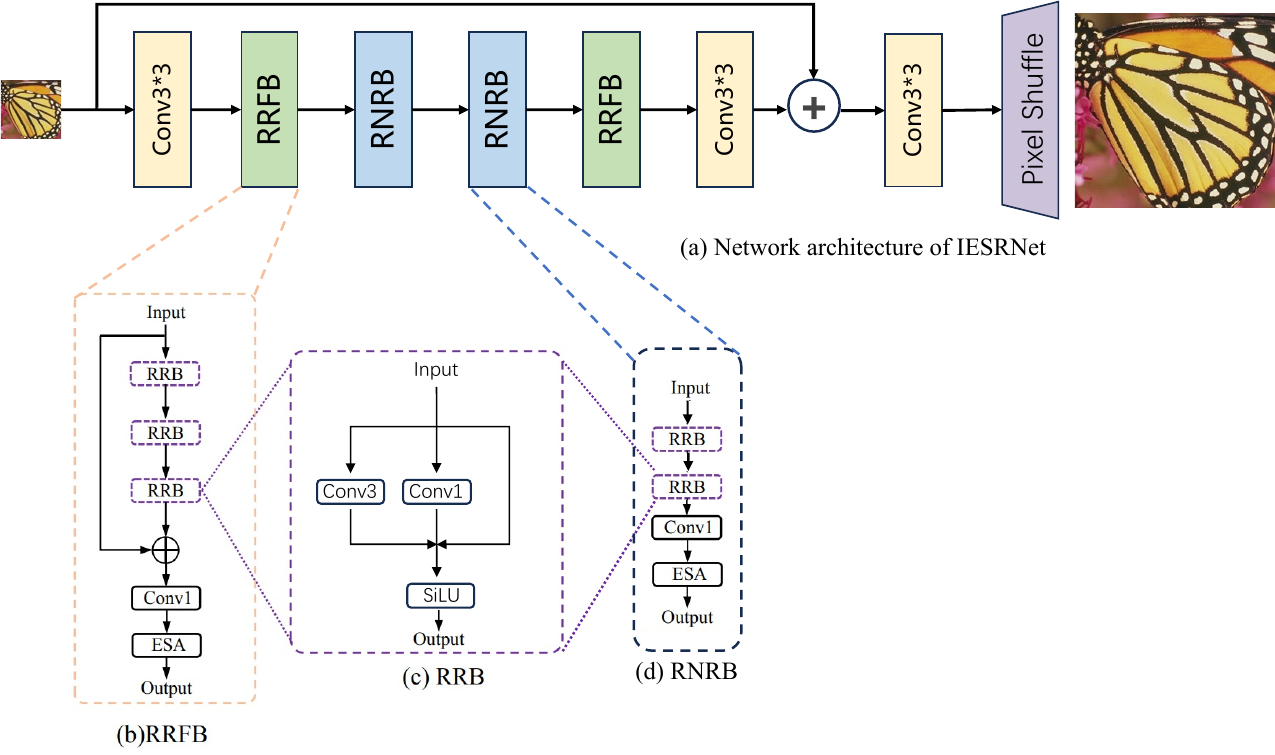}
\caption{\textit{Team IRSR:} The overview of the proposed IESRNet. The IESRNet is built based on DIPNet~\cite{yu2023dipnet}.}
\label{fig:arch}
\end{figure*}

\noindent
\textbf{Model Design.}
As for the Efficeint Super-Resolution competition, they proposed the Inference Efficient Super-Resolution Net (IESRNet). IESRNet is not a specific network, but a bag of tricks to make a Super-Resolution Network infer more Efficient on a GPU. They will apply these tricks based on DIPNet~\cite{yu2023dipnet}, which won the first place on the NTIRE2023 ESR challenge in runtime track~\cite{li2023ntire_esr}. The specific structure of IESRNet is shown in \cref{fig:arch}. They will describe the tricks they used in detail below.

1. Remove bias in Conv. The bias add of the convolution is a relatively inefficient operation in the convolution layer. It only occupies a small part of the FLOPs in the convolution, but occupies 15\% or more of the runtime. They removed the bias of all convolutional layers except the ESA module, and the PSNR loss was less than 0.01db.

2. Less Residual Connection. Although residual connection helps the model converge during training, too many residual structures will introduce many additional operations, reducing the inference efficiency of the model. Therefore, they replace the two middle RRFB in DIPNet with reparameterization no residual block(RNRB) to balance the trade-off between inference efficiency and model accuracy.

3. Standard number of Conv channels. Since the convolution operator has different performance optimizations for different configurations, generally, convolutions with a standard number of channels (such as 32, 48, and 64) are more deeply optimized and therefore occupy higher inference efficiency on the GPU. Based on NVIDIA V100 GPU testing, a 48-channel 3*3 convolution is even faster than a 30-channel convolution, although the FLOPs is over doubled. For this reason, they set the number of feature channels to 32, and the number of ESA channels to 16. 

4. Efficient activation function. They replace all activation functions in the network with SiLU~\cite{silu}, which performs well in super-resolution tasks and significantly outperforms the RELU. In addition to its great performance, SiLU is also very fast when inferring on GPUs due to its computational characteristics.

5. Reparameterization. They adopt re-parameterization to enhance the representation capabilities of the model. They use complex re-parameterization structures to train during training and merge them into regular convolutions during inference without incurring additional computational overhead. The specific rep-structure is shown in \cref{fig:arch}(c). 

\noindent\textbf{Implementation Details.}
The training dataset consists of DIV2K and the first 15,000 images of LSIDR~\cite{li2023lsdir}. Random flipping and rotation are adopt for Data Augmentation. They adopt a multi-stage training paradigm to train their super-resolution network. The details of training steps are as follows:

1. Initial training: HR patches of size 256 × 256 are randomly cropped from HR images. They set the mini-batch as 128. The model is trained by minimizing the PSNR loss with the Adam optimizer. The initial learning rate is set to 5e-4, and halved per 200k iterations. The total number of iterations is 1000k.

2. Warm-Start Training: Load the pre-trained weight and train it three times with the same setting.

3. Finetune with increasing patch size: In this process, the training patch size is progressively increased to improve the performance, which is selected from [384, 512, 640]. For each patch size, they finetune the network with 1000k iterations. And the initial learning  rate is correspondingly selected from [2e-4, 1e-4, 5e-5]. The batch size decreases to 64 for saving GPU memery. All experiments are conducted on 8 NVIDIA V100 GPUs.

% 10th of Main Track
\subsection{ASR}

\noindent
\textbf{Model Design.}
The network architecture is built based on DIPNet~\cite{yu2023dipnet}, which won the first place on the NTIRE2023 ESR challenge runtime track~\cite{li2023ntire_esr}. They made several modifications to make it more efficient while maintaining the excellent performance. They call it DIPNet\_slim.
% and its specific structure is shown in \cref{fig:arch}.

First of all, they did not use pruning as DIPNet dose. Although it can decrease the model parameters, it will degrade the inference speed of the model due to the irregular number of convolution channels. These operator configurations are not deeply optimized. For this reason, they set the number of feature channels to 32, and the number of ESA channels to 16. Second, they re-parameterize all 3x3 convolutional layers in the network. They adopt re-parameterization to enhance the expressiveness of the model. They use complex re-parameterization structures to train during training and merge them into regular convolutions during inference without incurring additional infer overhead. In addition, they changed the last convolution before the residual connection from 3x3 to 1x1, saving parameters while retaining the ability of feature normalization. Finally, they replace all activation functions in the network with SiLU~\cite{silu}, which performs well in super-resolution tasks and significantly outperforms the RELU.

\noindent\textbf{Implementation Details.}
The training dataset consists of DIV2K~\cite{div2k} and the first 15,000 images of LSIDR. The details of training steps are as follows:

1. Initial Training: HR patches of size 256 × 256 are randomly cropped from HR images. They set the mini-batch as 128. The model is trained by minimizing the PSNR loss with the Adam optimizer. The initial learning rate is set to 5e-4, and halved per 200k iterations. The total number of iterations is 1000k.

2. Warm-Start Training: Load the pre-trained weight and train it three times with the same setting.

3. Finetune with increasing patch size: In this process, the training patch size is progressively increased to improve the performance, which is selected from [384, 512, 640]. For each patch size, they finetune the network with 1000k iterations. And the initial learning  rate is correspondingly selected from [2e-4, 1e-4, 5e-4]. The batch size decreases to 64 for saving GPU memory.

% 11th of Main Track
\subsection{VPEG\_O}
\begin{figure*}
	\centering
	\includegraphics[width=0.99\textwidth]{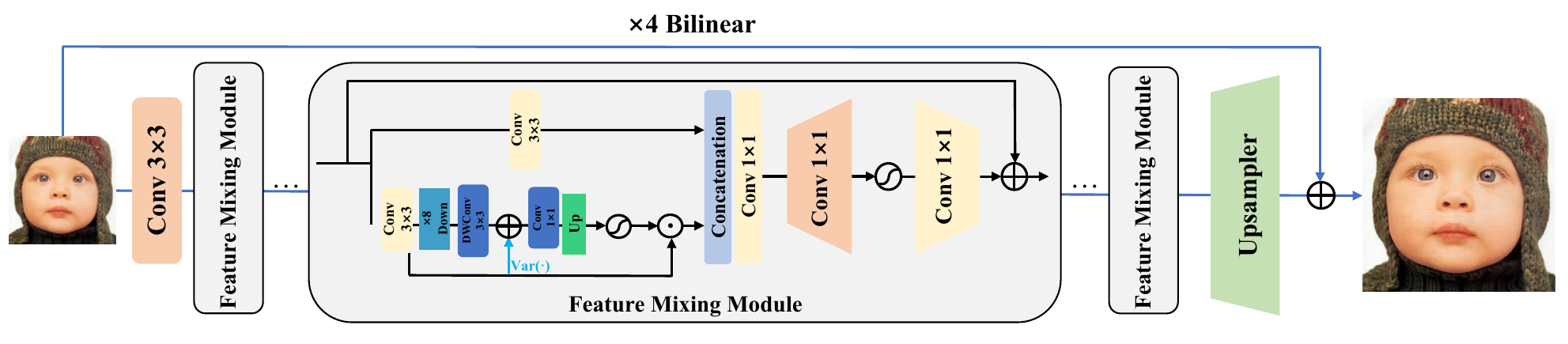}
	\caption{\textit{Team VPEG\_O}: An overview of the proposed SAFMNv3.}  
	\label{fig:framework}
\end{figure*}

\noindent
\textbf{General Method Description.}
They introduce SAFMNv3, an enhanced version of SAFMN~\cite{sun2023safmn} for solving real-time image SR.
This solution is mainly concentrates on improving the effectiveness of the spatially-adaptive feature modulation (SAFM)~\cite{sun2023safmn} layer.
Different from the original SAFMN, as shown in Fig~\ref{fig:framework}, the simplified SAFM layer is able to extract both local and non-local features simultaneously without channel splitting.
Within this module, they use two $3\times3$ convolutions to project the input and use variance-constrained feature modulation operator~\cite{smfanet} in branches with fewer channels, 
and finally aggregate these two parts of the feature, 
then refine the aggregated features via a feed-forward neural network.

\noindent
\textbf{Training Description.}
The proposed SAFMNv3 consists of 6 feature mixing modules, and the number of channels is set to 40.
They  rain the network on RGB channels and augment the training data with random flipping and rotation. 
Following previous methods, the training process is divided into three stages:
\begin{itemize}
    \item[1.] In the first stage, they randomly crop 256$\times$256 HR image patches from the selected LSIDR~\cite{li2023lsdir} dataset, with a batch size of 64. 
    The proposed SAFMNv3 is trained by minimizing L1 loss and the frequency loss\cite{MIMO} with Adam optimizer for total 800, 000 iterations
    The initial learning rate is set to 2e-3, with a Cosine Annealing scheme~\cite{SGDR}. 
    \item[2.] In the second stage, they increase the size of the HR image patches to 384$\times$384. 
    The model is fine-tuned on the DF2K~\cite{NTIRE_2017} by minimizing Charbonnier loss function.
    The initial learning rate is set to 5e-4, and the total iterations is 500k.
    \item[3.] In the third stage, the batch size is set to 64, and PSNR loss is adopted to optimize over 300k iterations. 
    The initial learning rate is set to 5e-5.  
\end{itemize}
Throughout the training process, they also employ an Exponential Moving Average (EMA) strategy to enhance the robustness of training.
% 12th of Main Track
\subsection{mmSR}
\begin{figure}  %
    \centering
    \includegraphics[width=\columnwidth]{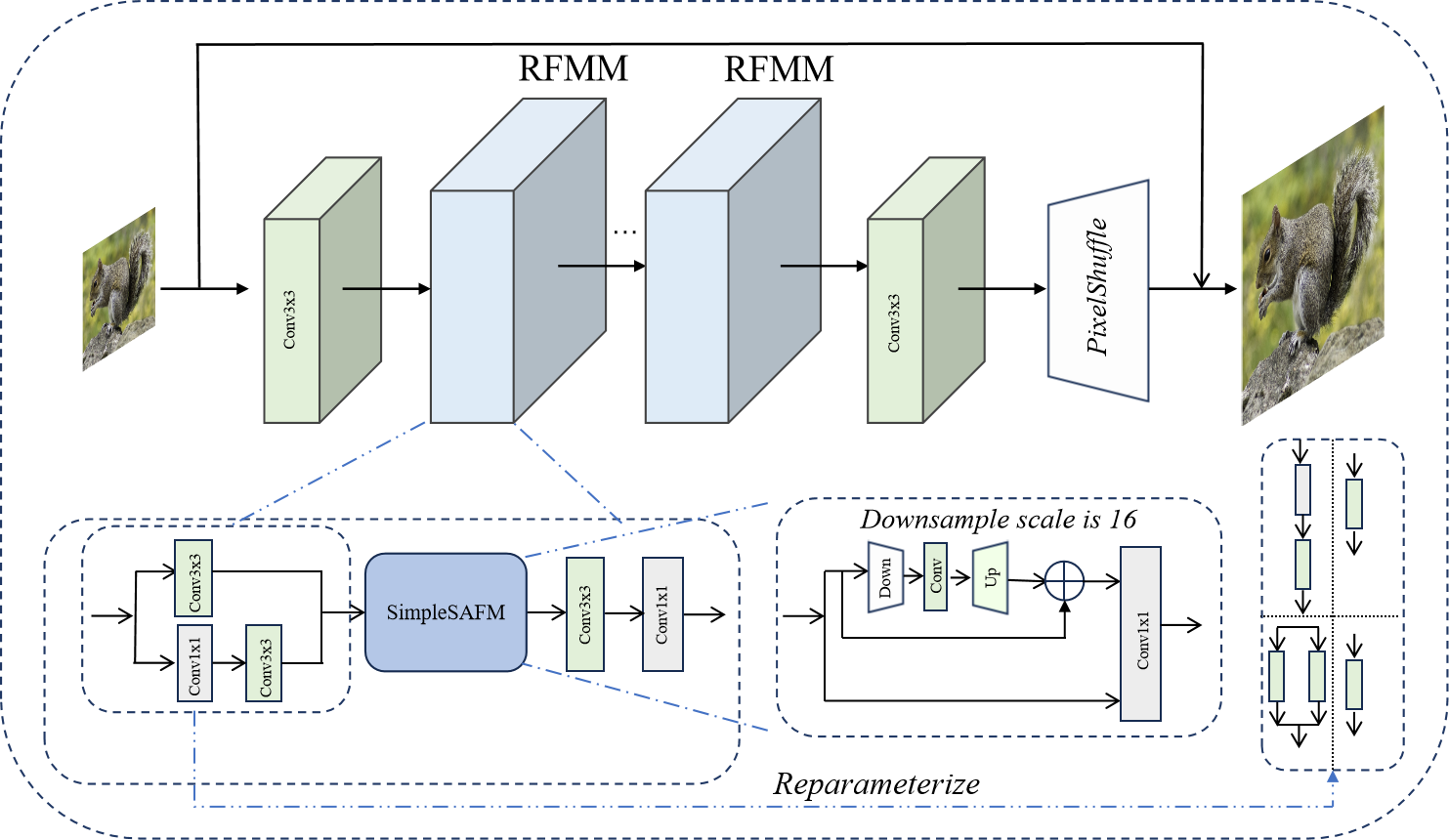}
    \caption{\textit{Team mmSR:} The overall network architecture of FAnet.}
    \label{fig:team08}
\end{figure}

\textbf{Method. } They improve the model based on SAFMN++ \cite{ren2024ninth} and name it FAnet as shown in Fig.~\ref{fig:team08}. Compared to SAFMN++, their model achieves a higher PSNR with a lower computational cost. Unlike the original SAFMN++ method, they introduce modifications in both the data and model structure. In terms of model structure, as shown in the figure, they improve the Feature Mixing Module of the original architecture and incorporate the concept of reparameterization, designing the RFMM. They modify the convolutional extraction network preceding the original module into a parallel structure to accommodate multi-granularity feature extraction and apply re-parameterization~\cite{repvgg} during inference. Furthermore, they adjust the downsampling factor in SimpleSAFM to 16 to achieve lower computational complexity. Regarding the data, in addition to utilizing the provided training dataset, they analyze the super-resolution results of the model and identify common issues in fine-detail generation. Given constraints on model parameters and computational resources, it is impractical for a lightweight model to generate details identical to the ground truth. Therefore, they shift their focus to expanding the training dataset. Specifically, they use 10,800 images from the training dataset as input and employ convolutional neural networks such as Omni-SR~\cite{wang2023omni} to generate new images. This additional data is incorporated into the training process to facilitate learning and mitigate the risk of learning bias caused by excessive learning difficulty.

\noindent
\textbf{Training Details.} They train their model on the DIV2K~\cite{NTIRE_2017}, Flickr2K~\cite{lim2017enhanced}, and LSDIR ~\cite{li2023lsdir} datasets. The cropped low-resolution (LR) image size is set to 64 × 64 and subjected to random flipping and rotation. The FAnet model is optimized using the Adam optimizer with L1 loss minimization in a multi-stage training scheme. During the training phase, they set the initial learning rate to \(2 \times 10^{-3}\) and the minimum learning rate to \(1 \times 10^{-6}\), training for 500,000 iterations with a mini-batch size of 512. In finetuning stage, Initialized with training phase weights, they fine-tune the model with the given training dataset and additional dataset which is proposed as above. They fine-tune the model using a learning rate of \(1 \times 10^{-4}\) and the minimum learning rate set to \(1 \times 10^{-6}\), with a mini-batch size of 64. 
% 13th of Main Track
\subsection{ChanSR}
\begin{figure*}[t]
	\centering
	\includegraphics[width=0.95\textwidth]{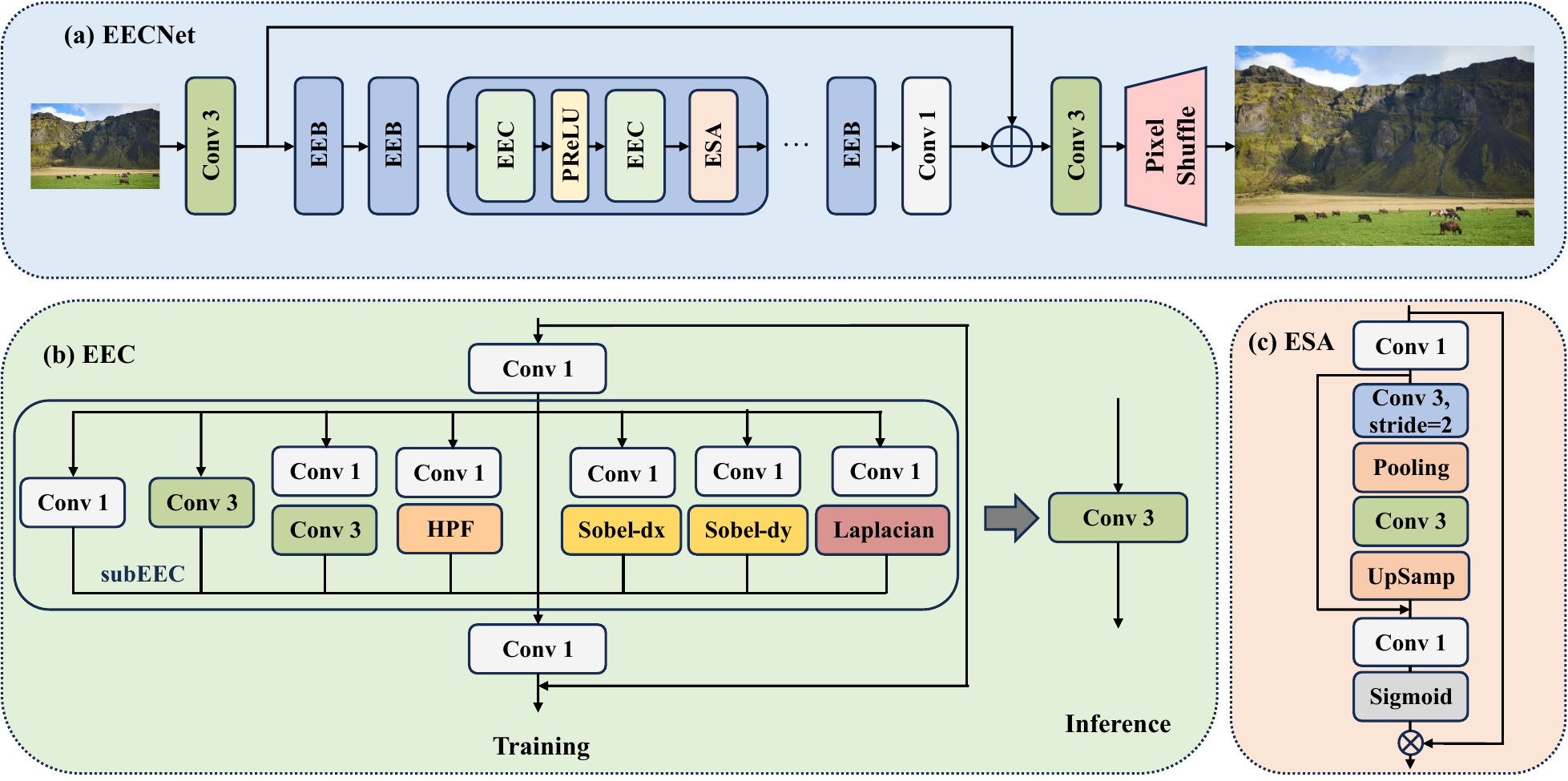}
	\caption{\textit{Team ChanSR:} Network architecture of the EECNet.}
	\label{team47:arch}
\end{figure*}

\noindent
\textbf{General Method Description.} They propose the Edge Enhanced Convolutional Network (EECNet) for the efficient super-resolution task. The network architecture is inspired by the design of SRN~~\cite{wang2023single}, while fully exploring the capacity of reparameterizable convolution. The whole architecture is shown in ~\cref{team47:arch}(a). They introduce a predefined High-Pass Filter (HPF) branch to explicitly capture edge details, formulated as:

\begin{equation}
\mathbf{K}_{hpf} =\frac{1}{16} \begin{bmatrix}
-1 & -2 & -1 \\
-2 & 12 & -2 \\
-1 & -2 & -1 
\end{bmatrix}.
\end{equation}

Then they integrate the proposed HPF into the EDBB~\cite{EFDN}, creating the subEEC module. As subEEC can be mathematically equivalent to a standard 3×3 convolution, they replace the original 3×3 convolution in RRRB~\cite{FMEN} with our subEEC to obtain the final EEC architecture, whose structure is shown in \cref{team47:arch}(b). Notably, to ensure valid re-parameterization, they initialize the bias of the first convolution layer as zero to compensate for the zero-padding operation in subEEC.

To better capture global spatial information, they adopt the simplified Efficient Spatial Attention mechanism from SRN~\cite{wang2023single}, whose structure is shown in Fig.~\ref{team47:arch}(c). Compared with the original ESA, this implementation removes the 1×1 convolution layer and reduces computational complexity by employing only a single 3×3 convolution in the convolutional group.

\noindent
\textbf{Training Description.} The proposed EECNet contains eight EEBs, in which they set the number of feature maps to 32. Also, the channel number of the ESA is set to 16 similar to ~\cite{kong2022residual}. 
Throughout the entire training process, they use the Adam optimizer ~\cite{kingma2014adam}, where $\beta{1}$ = 0.9 and $\beta{2}$ = 0.999. The model is trained for 1000k iterations in each stage.
Input patches are randomly cropped and augmented. Data augmentation strategies included horizontal and vertical flips, and random rotations of 90, 180, and 270 degrees.
Model training was performed using Pytorch 1.12.0 ~\cite{2019PyTorch} on RTX 3090.
Specifically, the training strategy consists of several steps as follows.

1. In the starting stage, they train the model from scratch on the 800 images of DIV2K~\cite{agustsson2017ntire} and the first 10k images of LSDIR~~\cite{li2023lsdir} datasets.
The model is trained for a total $10^6$ iterations by minimizing L1 loss and FFT loss~\cite{cho2021rethinking}. The HR patch size is set to 256$\times$256, while the mini-batch size is set to 64. They set the initial learning rate to 1 $\times$ $10^{-3}$ and the minimum one to 1 $\times$ $10^{-5}$, which is updated by the Cosine Annealing scheme.

2. In the second stage, they increase the HR patch size to 384, while the mini-batch size is set to 32. The model is fine-tuned by minimizing the L1 loss and the FFT loss. They set the initial learning rate to 5 $\times$ $10^{-4}$ and the minimum one to 1 $\times$ $10^{-6}$, which is updated by the Cosine Annealing scheme.

3. In the last stage, the model is fine-tuned with 480×480 HR patches, however, the loss function is changed to minimize the combination of L2 loss and FFT loss~\cite{cho2021rethinking}. Other settings are the same as Stage 2.
% 14th of Main Track
\subsection{Pixel\_Alchemists}

\begin{figure}[!ht]
    \centering
    \includegraphics[width=0.43\textwidth]{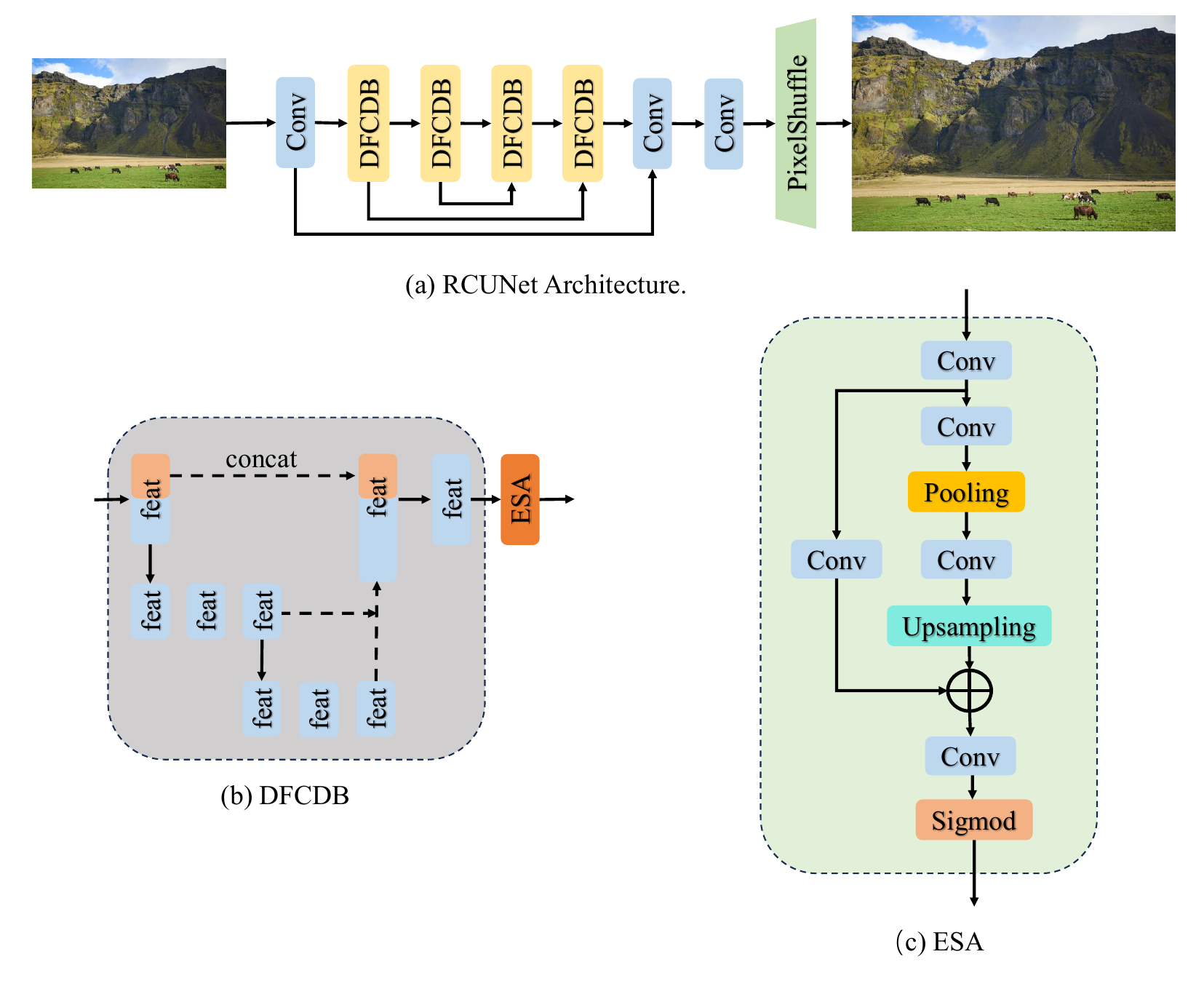}
    \caption{\textit{Team Pixel\_Alchemists:} RCUNet Architecture.}
    \label{fig:team37_arch}
\end{figure}

\textbf{Network Architecture.}
The overall architecture of team Pixel Alchemists is shown in ~\cref{fig:team37_arch}. They propose a novel architecture named resolution-consistent UNet (RCUNet). The proposed network consists of four deep feature complement and distillation blocks (DFCDB). Inspired by \cite{ghostnet, ghostSR}, the input feature map is split along the channel dimension in each block. Then, four convolutional layers process one of the split feature maps to generate complementary features. The input features and complementary features are concatenated to avoid loss of input information and distilled by a conv-1 layer. Besides, the output feature map of DFCDB is further enhanced by the ESA layer~\cite{RLFN}.

\begin{figure}[!ht]
    \centering
    \includegraphics[width=0.43\textwidth]{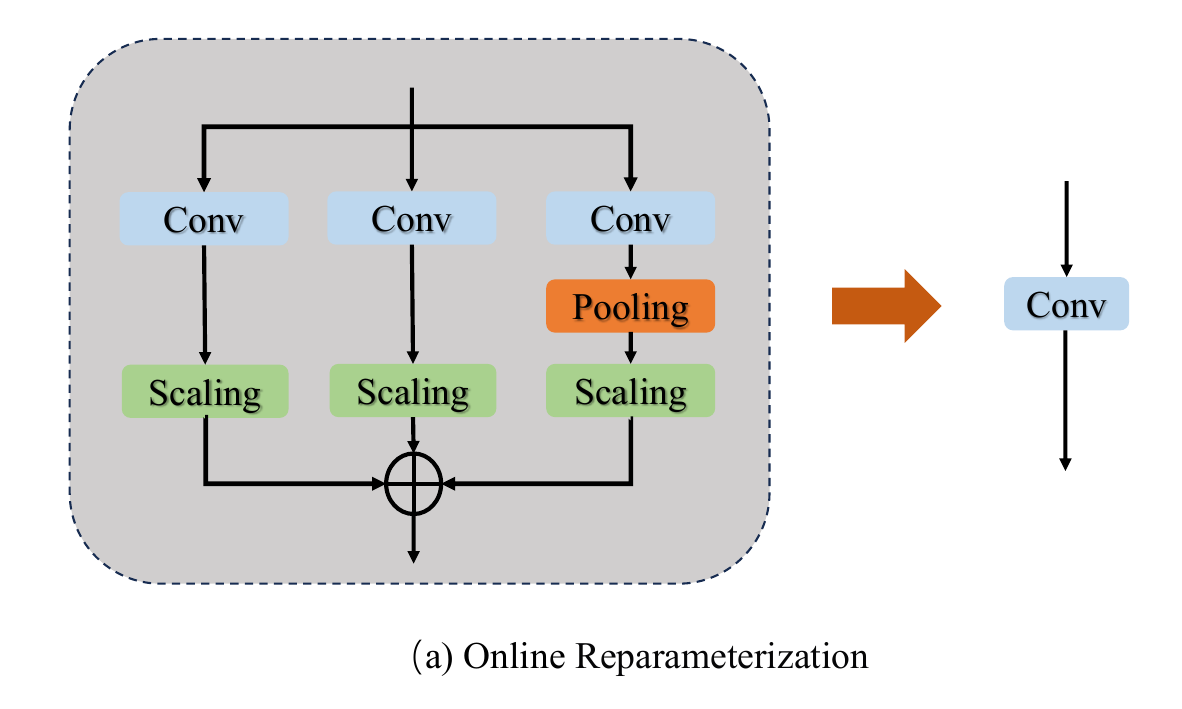}
    \caption{\textit{Team Pixel\_Alchemists:} Online re-parameterization. }
    \label{fig:team37_reparam}
\end{figure}

\noindent
\textbf{Online Convolutional Re-parameterization.} Re-parameterization~\cite{ECBSR} has improved the performance of image restoration models without introducing any inference cost. However, the training cost is large because of complicated training-time blocks. To reduce the large extra training cost, online convolutional re-parameterization~\cite{Online_Convolutional_Re-parameterization} is employed by converting the complex blocks into a single convolutional layer during the training stage. The architecture of RepConv is shown in Fig.~\ref{fig:team37_reparam}. It can be converted to a $3\times3$ convolution during training, which saves the training cost.

\noindent
\textbf{Training Details.}
The proposed RCUNet has four DFCDBs. The number of features is set to 48, and the number of ESA channels is set to 16. 

DIV2K~\cite{agustsson2017ntire} and LSDIR~\cite{li2023lsdir} datasets are used for training. The training details are as follows:

\begin{itemize}
  \item [1.] The model is first trained from scratch with $256\times256$ patches randomly cropped from HR images from the DIV2K and LSDIR datasets. The mini-batch size is set to 64. The L1 loss and pyramid loss are minimized with the Adam optimizer. The initial learning rate is set to 1e-3 with a cosine annealing schedule. The total number of iterations is 1000k.
  \item [2]
  Then the model is initialized with the pre-trained weights of Stage 1. The MSE loss and pyramid loss is used for fine-tuning with $512 \times 512$ HR patches and a learning rate of 1e-5 for 500k iterations. 
\end{itemize}

% 15th of Main Track
% \input{teams/team39_MiSR/main}      % NO Update
% 16th of Main Track
\subsection{LZ}

\begin{figure}[!ht]
    \centering
    \includegraphics[width=1.0\linewidth]{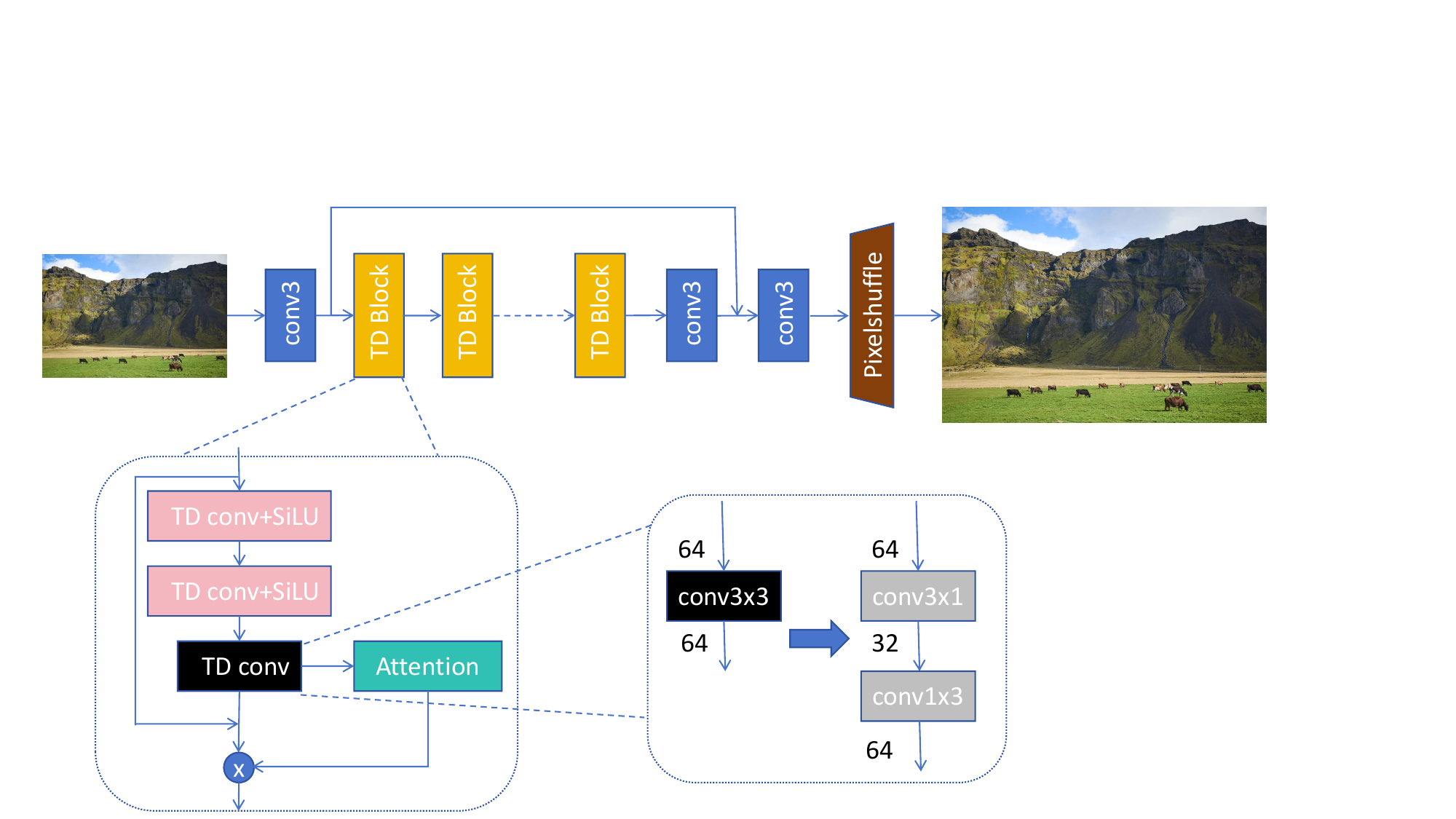}
    \caption{\textit{Team LZ:} Detailed architecture of TDESR.}
    \label{net}
\end{figure}

\noindent
\textbf{General Method Description.} To enhance model complexity without increasing computational overhead, they focus on designing structurally simple yet expressively powerful components, notably through re-parameterization techniques. Drawing inspiration from ECBSR~\cite{Zhang2021EdgeorientedCB}, their TDESR framework strategically implements re-parameterization to improve super-resolution performance while preserving training efficiency. Following the re-parameterization phase, they employ tensor decomposition for light-weight network design, where standard 3×3 convolutions are factorized into sequential 3×1 and 1×3 convolutional operations.

As illustrated in \cref{net}, their architecture comprises five TD Blocks interspersed with three standard 3×3 convolutions, implementing a skip connection through element-wise addition between the input features (processed by a 3×3 convolution) and intermediate feature maps. The network maintains 64 channels throughout, with tensor decomposition intermediate channels reduced to 32 for computational efficiency. They integrate insights from Swift-SR's parameter-free attention mechanism~\cite{wan2024swift} to enhance feature representation. The final reconstruction stage employs PixelShuffle with 48 input channels for high-quality image upsampling, completing their balanced design of performance and efficiency.

\noindent
\textbf{Training Details.} 
The training details of team LZ are as follows.

\begin{itemize}
\item \textbf{Base Training (×2 upscaling)} The model is initially trained for ×2 super-resolution using randomly cropped 96×96 HR patches with a batch size of 32. They employ the Adam optimizer to minimize the L1 loss, starting with an initial learning rate of $1\times10^{-4}$ that decays via MultiStepLR scheduler at the mid-training point. The training completes over 100 epochs, utilizing re-parameterization techniques throughout the process.

\item \textbf{Enhanced Resolution Training.} Building upon the ×2 pretrained weights, this phase increases the HR patch size to 128×128 while reducing the batch size to 16. All other hyperparameters (optimizer, learning rate schedule, and re-parameterization) remain consistent with Stage 1. The continued use of L1 loss maintains training stability during this resolution scaling phase.

\item \textbf{Convolutional Architecture Refinement.} They implement standard 3×3 convolutional layers in this optimization stage, replacing previous architectural components. The training objective shifts to L2 loss minimization for fine-tuning, while preserving the fundamental network structure and parameter initialization from earlier stages. This transition enhances edge preservation in super-resolved outputs.

\item \textbf{Tensor Decomposition Optimization.} The final refinement employs tensor decomposition techniques with dual loss supervision (L1 + L2). Training progresses with 256×256 HR patches using a reduced batch size of 16 and lower initial learning rate ($1\times10^{-5}$). They implement cosine annealing scheduling for smooth convergence, completing the multi-stage optimization process through L2-loss-focused fine-tuning..
\end{itemize}

% 17th of Main Track
\subsection{Z6}
 \begin{figure}[!tb]
 	\centering
 	\includegraphics[width=0.46\textwidth]{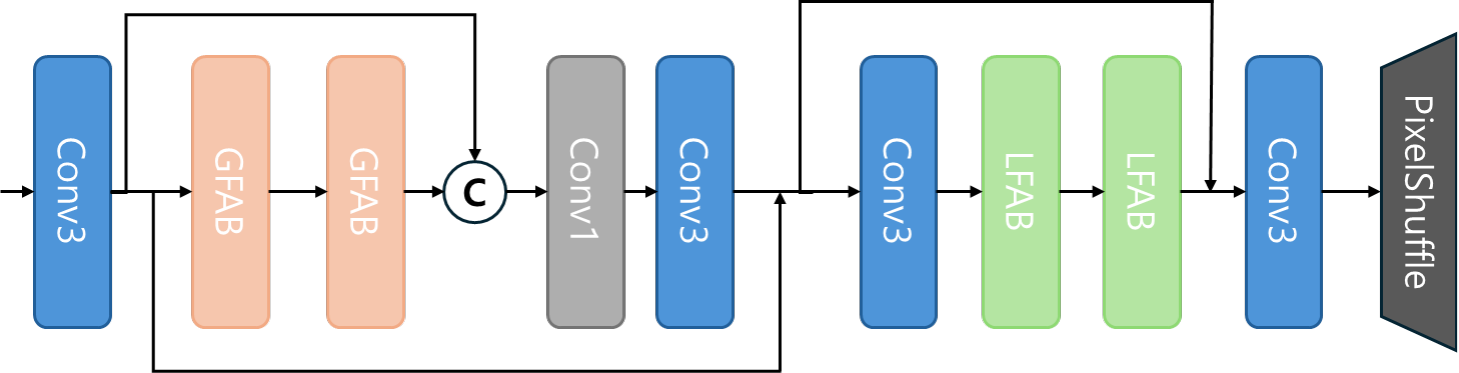}
 	\caption{\textit{Team Z6:} Network architecture of GloReNet.}
 	\label{team48:z6}
\end{figure}

\noindent
\textbf{General Method Description.} They introduce a lightweight and efficient image super-resolution (SR) network that leverages both global and local feature attention mechanisms to produce high-quality reconstructions. As depicted in \cref{team48:z6}, their network is divided into two main blocks named Global Feature Attention Block (GFAB) and Local Feature Attention Block (LFAB). 

GFAB is designed to capture large-scale context and dependencies across the entire image. Enhances globally significant features, helping the model learn the global information from input images. And LFAB can focus on refining fine-grained details and spatially localized information. Emphasizes subtle textural elements and sharp edges that are critical for upscaling. GFAB utilizes the parameter-free attention module (SPAN \cite{span}) and LFAB uses Efficient Spatial Attention (ESA) \cite{liu2020residual} to selectively highlight essential features. And all convolution layers applied reparameterization block \cite{yoon2024casr}. The network begins with a series of convolution layers to extract initial features, which then pass through GFAB units for global attention. Subsequently, the output is processed by LFAB units for local attention, and finally, a PixelShuffle layer upscales the features to the target resolution. By combining these two parts, their method effectively preserves global context and local details, achieving a balance between high-quality reconstruction and efficient low computation.

\noindent
\textbf{Training Description.} Their training process employs a scratch training stage and a fine-tuning stage. In the first scratch training stage, they use DIV2K datasets for the training dataset. In the fine-tuning stage, they use DIV2K and the first 10K LSDIR datasets for the training dataset. All experiments are carried out in the same experimental environment. The training process is executed using RTX A6000 GPUs. They use the Pytorch 1.13 version for all training steps.

\begin{itemize}
    \item Scratch train stage:
    In the first step, their model is trained from scratch. The LR patches were cropped from LR images with an 8 mini-batch of 256 x 256.  Adam optimizer is used with a learning rate of 0.0005 during scratch training. The cosine warm-up scheduler is used. The total number of epochs is set to 2000. They use the $l1$ loss.
    \item Fine-tuning stage:
    In the second step, the model is initialized with the weights trained in the first step. To improve precision, they used the loss method $l2$ loss. This stage improves the value of the peak signal-to-noise ratio (PSNR) by 0.05 $\sim$ 0.06 dB. In this step, The LR patches are cropped from LR images with 32 mini-batch 512 x 512 sizes. And the initial learning rate is set to 0.00005 and the Adam optimizer is used in conjunction with a cosine warm-up. The total epoch is set to 200 epochs. 
\end{itemize}

% 18th of Main Track
\subsection{TACO\_SR}
\begin{figure*}[t]
  \centering
  \includegraphics[width=\linewidth]{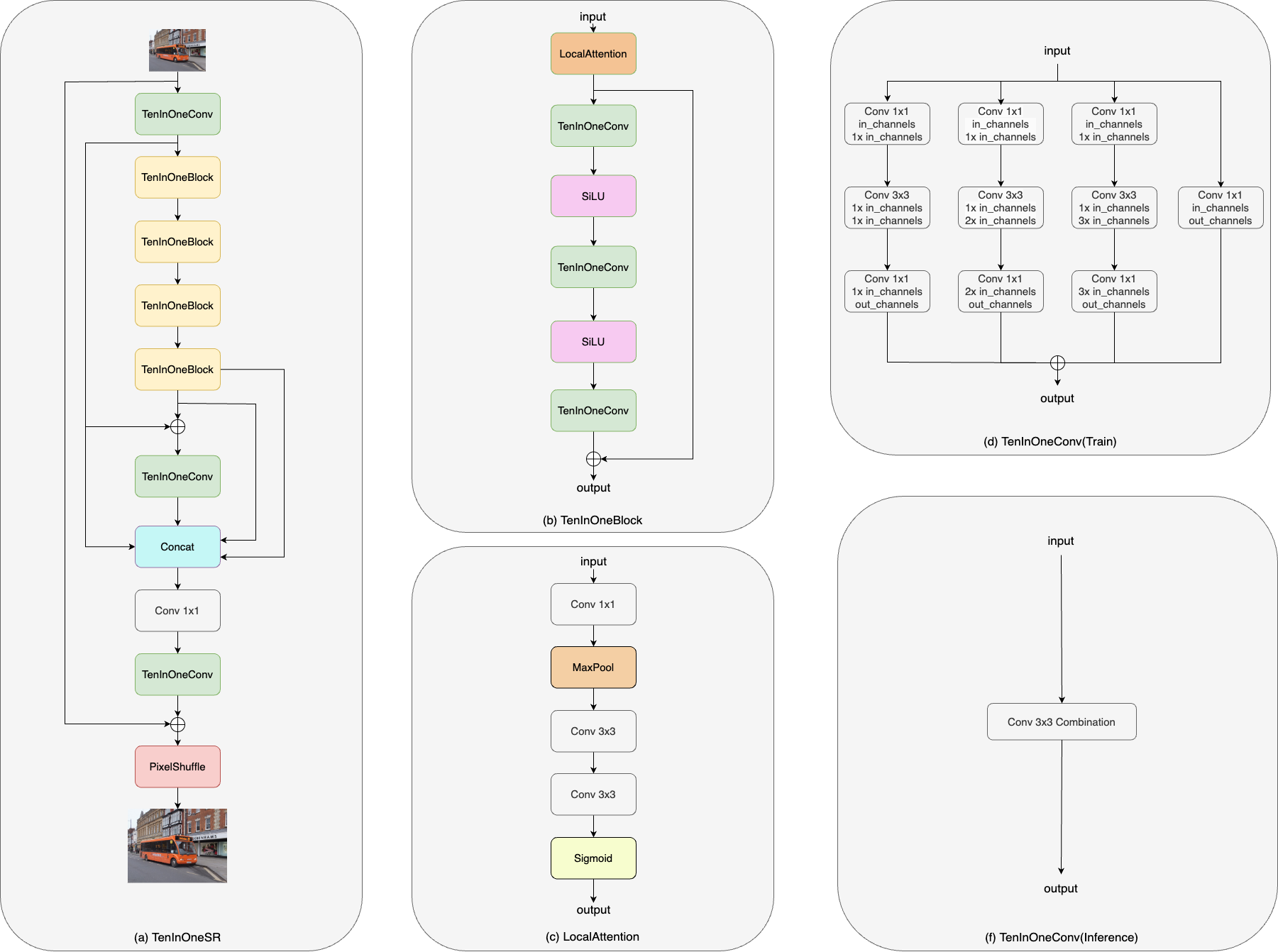}
  \caption{\textit{Team TACO\_SR}: The architecture of proposed TenInOneSR.}
  \label{fig:50architecture}
\end{figure*}

\textbf{General Method Description. } The overall architecture of their network is showed in~\cref{fig:50architecture}(a) , inspired by SPAN~\cite{SPAN2024} and PFDNLite~\cite{ren2024ninth}. Motivated by the design of the Conv3XC module in SPAN, they introduce two additional parallel branches with varying channel expansion ratios, resulting in a novel convolution module termed TenInOneConv, which fuses multiple convolution kernels into a single equivalent kernel to improve inference efficiency. Furthermore, to enhance the model’s capability in capturing local texture and detail features, the LocalAttention module, inspired by PFDNLite is integrated, allowing the network to better focus on informative regions within feature maps. 

TenInOneSR employs four TenInOneBlock modules. Each of these blocks (detailed in~\cref{fig:50architecture}(b)) begins with a LocalAttention module, which enhancing the network’s ability to capture fine details. Subsequently, each block applies three cascaded TenInOneConv layers, interleaved with the SiLU activation function, to perform hierarchical feature refinement. The block concludes with a residual connection, allowing better gradient flow.

Notably, the behavior of the TenInOneConv differs between the training and inference phases. During training (~\cref{fig:50architecture}(d)), TenInOneConv operates in a multi-branch configuration. It introduces three parallel convolutional branches with different channel expansion ratios (gains set as 1, 2, and 3), along with an additional skip connection. This multi-scale feature extraction enables the network to better aggregate complementary spatial features.

In the inference stage (~\cref{fig:50architecture}(f)), for computational efficiency and faster runtime, these multiple convolution kernels are fused into a single equivalent convolution kernel. Specifically, the parallel branches and skip connection weights are mathematically combined to form one unified 3×3 convolutional kernel, significantly accelerating inference without compromising performance.\\

\noindent \textbf{Training description. }  The proposed architecture is trained on two NVIDIA RTX Titan GPUs with a total of 48 GB memory. 
In the first training stage, the DIV2K dataset is augmented by a factor of 85$\times$ and registered into the LSDIR format, resulting in a large-scale training set containing 152{,}991 high-resolution RGB images. During this stage, training is conducted with 64 randomly cropped 256$\times$256 patches per batch, using common augmentations such as random flipping and rotation. The model is optimized using the Adam optimizer with L1 loss for a total of 100{,}000 iterations. The learning rate is initialized at 5$\times$10$^{-4}$ and decayed by half every 20{,}000 iterations. In the second stage, they keep the training strategy and hyperparameters unchanged, except for increasing the input patch size to 384$\times$384 and reducing the batch size to 32 to fit GPU memory. Then another 100{,}000 training iterations are conducted to further improve the model’s performance on higher-resolution textures.

% 19th of Main Track
\subsection{AIOT\_AI}
\begin{figure*}[!tb]
    \centering
    \includegraphics[width=\linewidth]{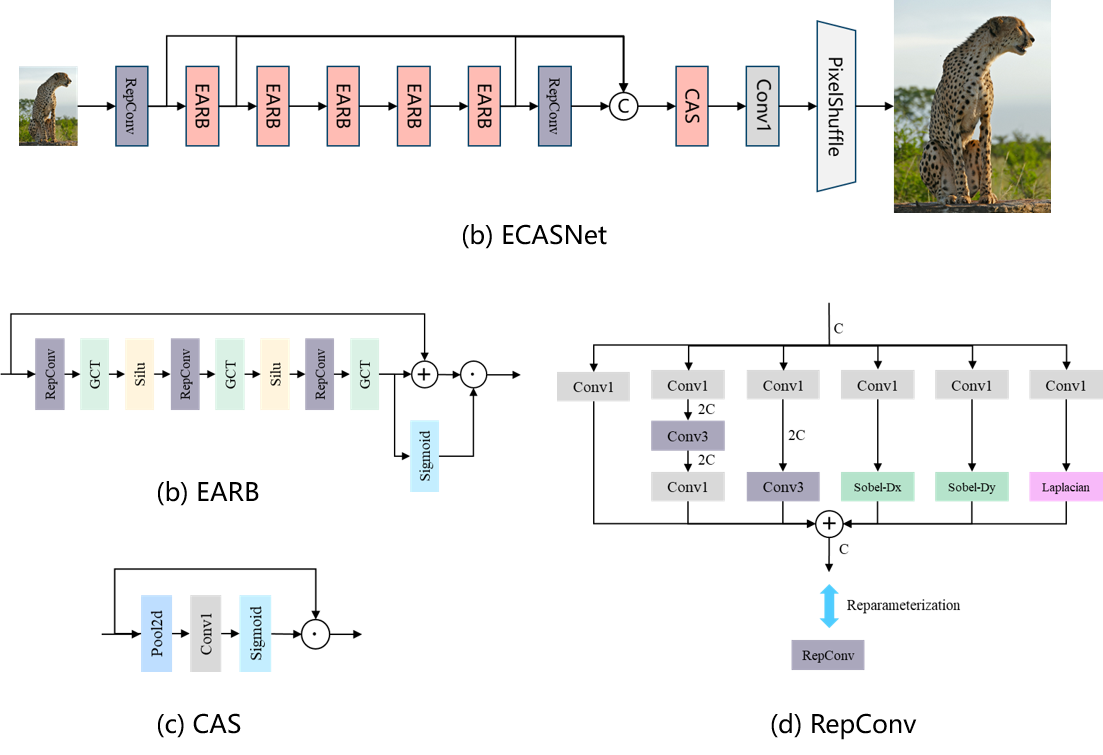}
    \caption{\textit{Team AIOT\_AI}: Detailed architecture of the proposed ECASNet.}
    \label{fig:team21q}
\end{figure*}

\textbf{Method}. The overall architecture of their network is shown in \cref{fig:team21q}(a), inspired by the previous leading methods SPAN\cite{wan2024swift} and ECBSR\cite{zhang2021edge}. They propose an Efficient channel attention super-resolution network acting on space (ECASNet). Specifically, on the basis of SPAB from SPAN, they combine edge-oriented convolution block (ECB) and regularization module (GCT) to form a new re-parameterized feature extraction module named enhanced attention and  re-parameterization block(EARB), as shown in 
\cref{fig:team21q}(b). In addition, unlike SPAN , they find that using channel attention after feature map concatenating can significantly improve performance. For the sake of lightweight design, they use an efficient channel attention module, called the efficient channel attention module which acts on space(CAS) , as shown in \cref{fig:team21q}(c).

\noindent\textbf{Training Detail.} The datasets used for training include DIV2K and LSDIR. Imitating the previous method, the training process is divided into two stages. In the first stage, they randomly crop 256x256 HR image blocks from the ground truth image, batch is 16, and randomly flipped and rotated them. Using Adam optimizer, set $\beta$1=0.9 and $\beta$2=0.999, and minimize L1 loss function. The initial learning rate is set to 5e-4, and the cosine learning rate attenuation strategy is adopted. Epoch is set to 200. In the second stage, they changed the loss function to L2, and other settings are the same as those in the first stage.

% 20th of Main Track
\subsection{JNU620}
\begin{figure*}[!tb]
% \begin{figure*}[htbp]
% \begin{figure}[htbp]
    \centering
    \includegraphics[width=0.98\textwidth]{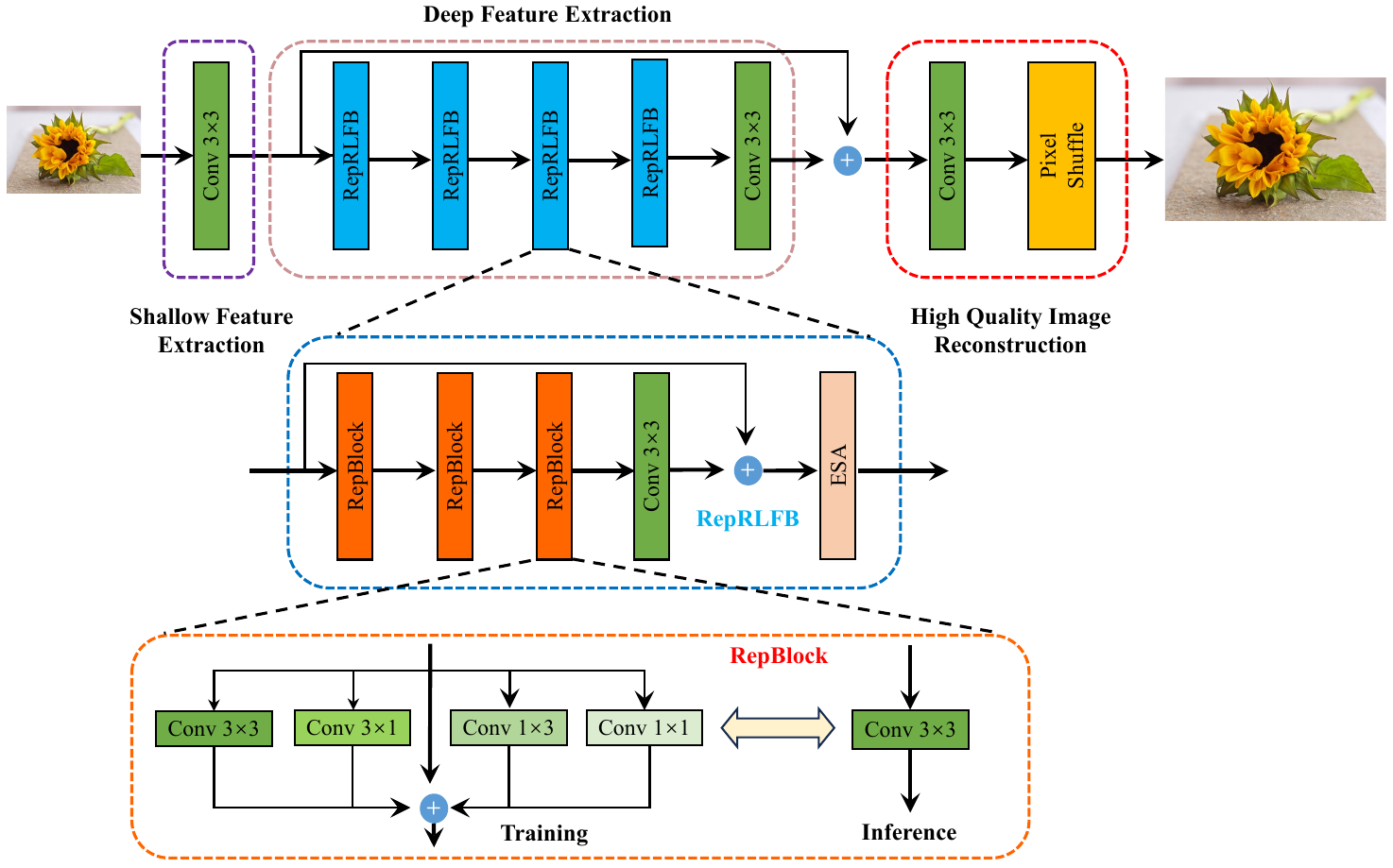}
%     \caption{The network architecture of the
% proposed improved SwinIR network}
    \caption{\textit{Team JUN620}: The network architecture of RepRLFN}
    \label{01network}
    \vspace{-.2mm}
\end{figure*}

% Method
\textbf{General Method Description.}
They propose a reparameterized residual local feature network (RepRLFN) for efficient image super-resolution, which is influenced by existing studies such as RepRFN~\cite{deng2023reparameterized} and RLFN~\cite{RLFN}. ~\cref{01network} illustrates the overall architecture of RepRLFN, which has been extensively validated in previous studies.

They replace the RLFB in RLFN~\cite{RLFN} with their reparameterized residual local feature block (RepRLFB). RepBlock is the main component of RepRLFB, which employs multiple parallel branch structures to extract the features of different receptive fields and modes to improve performance. At the same time, the structural re-parameterization technology is leveraged to decouple the training and inference phases to avoid the problem that computational complexity increases caused by the introduction of multi-branch.

% \begin{figure*}[!t]
% \end{figure}

% Training Strategy
\textbf{Training Strategy.}
The proposed RepRLFN consists of 4 RepRLFBs, with the number of feature channels set to 48. The details of the training steps are as follows:

1. In the first stage, the model is pre-trained on DIV2K~\cite{agustsson2017ntire}. HR patches of size 480×480 are randomly cropped from HR images, and the mini-batch size is set to 32. The model is trained by minimizing the L1 loss function using the Adam optimizer. The initial learning rate is set to 5e-4 and is halved every 200 epochs. The total number of epochs is 800.

2. In the second stage, the model is fine-tuned on 3450 images from DIV2K~\cite{agustsson2017ntire} and Flickr2k~\cite{flickr2k} (DF2K) and the first 10k images from LSDIR~\cite{li2023lsdir}. HR patches of size 640×640 are randomly cropped from HR images, and the mini-batch size is set to 32. The model is fine-tuned by minimizing the L2 loss function. The initial learning rate is set to 2e-4 and is halved every 5 epochs. The total number of epochs is 25.

3. In the third stage, the model is fine-tuned again on 3450 images from DF2K and the first 10k images from LSDIR~\cite{li2023lsdir}. The HR patch size and minibatch size are set to 640×640 and 32, respectively. The model is fine-tuned by minimizing the L2 loss function. The initial learning rate is set to 1e-4 and is halved every 5 epochs. The total number of epochs is 20.

4. In the fourth stage, the model is fine-tuned on 3450 images from DF2K and the first 10k images from LSDIR~\cite{li2023lsdir}. The HR patch size and minibatch size are set to 640×640 and 32, respectively. The model is fine-tuned by minimizing the L2 loss function. The learning rate is set to 5e-5, and the total number of epochs is 10. To prevent over-fitting, the model ensemble via stochastic weight averaging~\cite{izmailov2018averaging} (SWA) is performed during the last 8 epochs to obtain the final model for testing.
% 21st of Main Track
\subsection{LVGroup\_HFUT}

\begin{figure}[!tb]
    \centering
    \includegraphics[width=\linewidth]{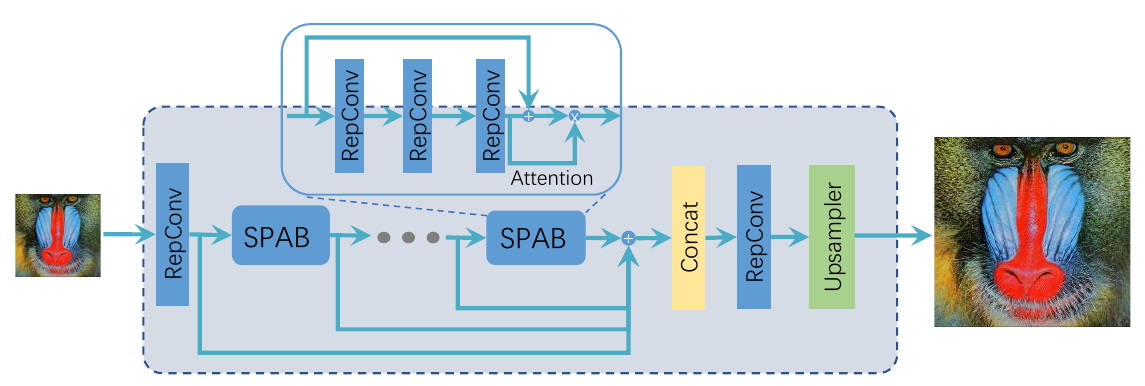}
    \caption{\textit{LVGroup\_HFUT}: The overall framework of SPAN.}
    \label{fig:lvgroup-hfut-solution}
\end{figure}

\noindent
\textbf{General Method Description.} The Swift Parameter-free Attention Network (SPAN) \cite{wan2024swift} introduces a novel parameter-free attention mechanism to address the trade-off between performance and computational complexity, as shown in \ref{fig:lvgroup-hfut-solution}. SPAN employs symmetric activation functions (e.g., shifted Sigmoid) applied to convolutional layer outputs to generate attention maps without learnable parameters, enhancing high-contribution features while suppressing redundant information. Residual connections within each Swift Parameter-free Attention Block (SPAB) mitigate information loss and preserve low-level features. The lightweight architecture with cascaded SPABs achieves fast inference by avoiding parameter-heavy attention computations while maintaining reconstruction quality through hierarchical feature aggregation and pixel-shuffle upsampling.

\noindent
\textbf{Training Details.} They trained the SPAN model \cite{wan2024swift} on a mixed dataset composed of DIV2K \cite{timoftediv2k} and LSDIR \cite{li2023lsdir}, setting feature\_channels to 48, where the crop size of images is 256x256. They used the Adam optimizer with L1 loss, an initial learning rate of 5e-4, and trained for a total of 1000k iterations, halving the learning rate every 200k iterations. Training was completed using a single NVIDIA RTX 4090 GPU.

% 22nd of Main Track
% \input{teams/team43_SVM/main}       % NO Update
% 23rd of Main Track
\subsection{YG}
\subsubsection{Method Details.}
The Primary idea of the proposed SGSDN is to explore non-local information in a SA-like manner while modeling local details for efficient image super-resolution. This section will start by introducing the overall architecture of SGSDN and then explain the SGM and ESD in detail.

\begin{figure}[!tb]
    \centering
    \includegraphics[width=1\linewidth]{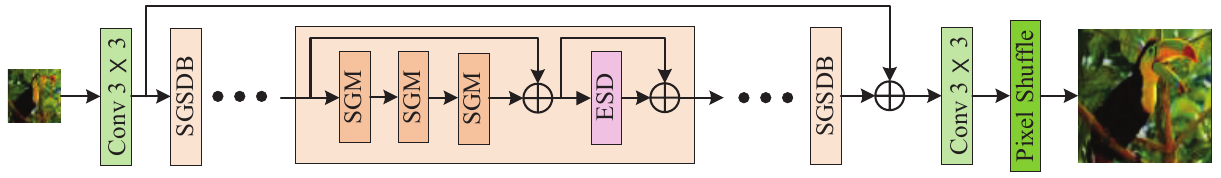} %1
    \caption{\textit{Team YG:} The Spatial-gate self-distillation network (SGSDN).}
    \label{fig:team18_fig1}
\end{figure}

\begin{figure}[!tb]
    \centering
    \includegraphics[width=0.7\linewidth]{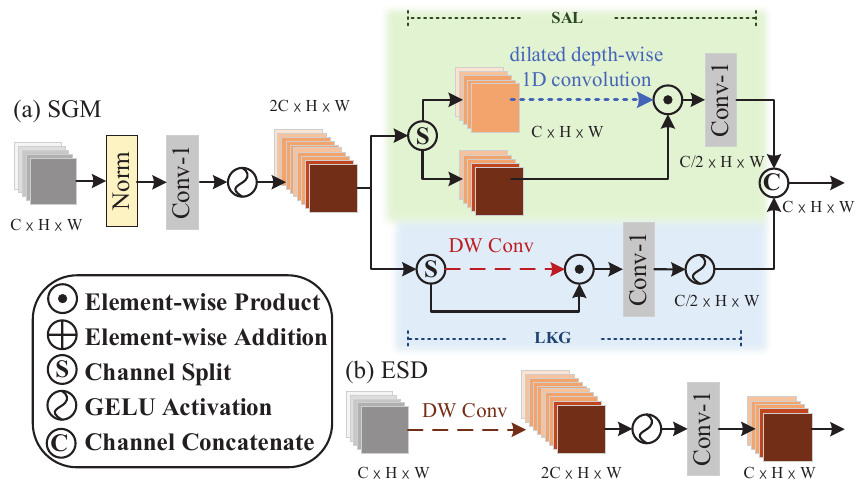} %1
    \caption{\textit{Team YG:} The details of each component. (a) SGM: Spatial-gate modulation module; (b) ESD: Enhanced self-distillation module.}
    \label{fig:team18_fig2}
\end{figure}

\noindent
\textbf{Network Architecture} 
\label{subsec:Network Architecture}
The overall structure of the SGSDN is shown in ~\cref{fig:team18_fig1}. It consists of three stages: shallow feature extraction, deep feature extraction, and image reconstruction. First, they use  a $3\times3$ convolutional layer to extract shallow features, which is expressed as:
\begin{equation}
 \mathbf{I}_s = F_{Conv3\times3}(\mathbf{I}_{LR}),
\end{equation}
where, $F_{Conv3\times3}$ represents the shallow feature extraction module using a $3 \times 3$ convolutional layer. The obtained shallow feature is denoted as $\mathbf{I}_s$. Subsequently, the extracted shallow features are fed to several stacked SGSDBs to produce deep representative features, This process can be expressed as:
\begin{equation}
 \mathbf{I}_k = F_{SGSDB}^k(\mathbf{I}_{k-1}),k=1,\cdots,n,
\end{equation}
where, $F_{SGSDB}^k\left ( \cdot  \right )$ represents the $k$-th SGSDB, $\mathbf{I}_{k-1}$ and $\mathbf{I}_k$ denote the input and output features of the $k$-th SGSDB, respectively. Each SGSDB consists of three SGMs and an ESD . Given an input feature $\mathbf{I}_t$, the mapping process of SGSDB can be represented as:
\begin{equation}
    \begin{aligned}
        \mathbf{I}_{d_1} & = F_{SGM}(\mathbf{I}_t),\\
        \mathbf{I}_{d_2} & = F_{SGM}(\mathbf{I}_{d_1}),\\
        \mathbf{I}_{d_3} & = F_{SGM}(\mathbf{I}_{d_2}) + \mathbf{I}_t,\\
        \mathbf{I}_{o} & = F_{ESD}(\mathbf{I}_{d_3}) + \mathbf{I}_{d_3}\\
	\end{aligned}
\end{equation}
where, $F_{SGM}$ represents the SGM, $F_{ESD}$ represents the ESD. After the deep feature extraction block, the representative features are processed by a 3 $\times$ 3 standard convolution layer and a pixel shuffle operation \cite{shi2016real} to reconstruct the high-quality SR image. To take advantage of high-frequency
information, they insert a long-distance residual connection before the image reconstruction module. The reconstruction stage is described as follows
\begin{equation}
	\textbf{I}_{SR}=F_{PixelShuffle}(F_{Conv3\times3}(\mathbf{I}_d + \mathbf{I}_s )),
\end{equation}
where $\mathbf{I}_d$ denotes the deep feature obtained by the stacked  SGSDBs, and $F_{Conv3\times3}(\cdot)$ indicates the 3 $\times$ 3 standard convolution layer. $F_{PixelShuffle}(\cdot)$ is used to upscale the final feature and output the SR reconstructed image $\textbf{I}_{SR}$.
\par
Finally, to train the network, they use the $L_1$ loss function to minimize the pixel-level difference between the ground truth image $\textbf{I}_{GT}$ and the reconstructed image $\textbf{I}_{SR}$, which can be expressed as:
\begin{equation}
	L_1=\left \|\textbf{I}_{SR} -\textbf{I}_{GT}\right \|_1,
\end{equation}
At the same time, they notice that only using the pixel-wise loss function can not effectively generate more high-frequency details \cite{cho2021rethinking}. Thus, they accordingly employ a frequency constraint to regularize network training. The adopted loss function for the network training is defined as
\begin{equation}
	L=L_{1} +\lambda \left \| \mathcal{F} \left (\textbf{I} _{SR}  \right )-\mathcal{F} \left ( \textbf{I}_{GT}  \right )   \right \|.
\end{equation}
where $\mathcal{F}$ represents the Fast Fourier Transform, and $\lambda$ is a weight parameter which is empirically set to 0.1.

\noindent
\textbf{Spatial-gate modulation module}\label{SGM}
Considering that the reason why the ViT-based model performs well is that SA explores non-local information and expands the effective receptive field of the model. 
% Inspired by \cite{guo2023visual}, the depth-wise convolution can be further decomposed. 
They develop a lightweight spatial-gate modulation (SGM) module to collaboratively extract representative features, where the SAL branch exploits non-local features in a larger receptive field by integrating the dilated depth-wise convolutional layers with horizontal and vertical 1-D kernels, and the LKG branch captures local features in parallel. Moreover, to avoid potential block artifacts aroused by dilation, they adopt the gate mechanism to recalibrate the generated feature maps adaptively, as shown in Fig.~\ref{fig:team18_fig2}.
\par
Given the input feature $\textbf{I}_{in} \in R ^ {C \times H \times W}$, where $H \times W$ denotes the spatial size and $C$ is the number of channels, Specifically, they first apply a normalization layer and a point-by-point convolution to normalize information and expand the channel.
 \begin{equation}
 \mathbf{I}_1 = F_{Conv1\times1}(F_{Norm}(\mathbf{I}_{in})),
\end{equation}
where, $F_{Norm}$ represents the $L_2$ normalization and $F_{Conv1\times1}$ denotes a $1 \times 1$ convolutional layer, $\mathbf{I}_1 \in R ^ {2C \times H \times W}$. Subsequently, the obtained features $\mathbf{I}_1$ are splitted into two parts along the channel dimension, this process can be expressed as:
\begin{equation}
	\mathbf{I}_{x}, \mathbf{I}_{y} = F_S(F_G (\mathbf{I}_1)),
\end{equation}
where $F_G$ denotes the GELU activation function \cite{hendrycks2016gaussian}, $F_S$ denotes a channel splitting operation, $\mathbf{I}_{x} \in R ^ {C \times H \times W}$ and $\mathbf{I}_{y} \in R ^ {C \times H \times W}$. They then process the features $\mathbf{I}_{x}$ and $\mathbf{I}_{y}$ in parallel via the SAL and LKG branches, producing the non-local feature $\mathbf{I}_{n}$ and local feature $\mathbf{I}_{l}$, respectively. It is worth mentioning that the SAL and LKG branches only need to be responsible for half the input signals, and the parallel processing is faster. Finally, they fuse the non-local feature $\mathbf{I}_{n}$ and local feature $\mathbf{I}_{l}$ together with channel concatenation to form a representative output of the SGM module. This process can be expressed as,
\begin{equation}
	\mathbf{I}_{SGM} = F_C(\mathbf{I}_{n}, \mathbf{I}_{l}),
\end{equation}
where, $\mathbf{I}_{DSG}$ is  the output feature and $F_C( \cdot )$ is the channel cascade operation. 

\noindent
\textbf{SA-like branch}
They exploit non-local features in a larger receptive field by integrating the dilated depth-wise convolutional layers with horizontal and vertical 1-D kernels.
\begin{equation}
    \begin{split}
		\mathbf{I}_{o} = F_{D^3WConv5\times11}(F_{DWConv5\times1} \\
(F_{D^3WConv1\times11}(F_{DWConv1\times5}({\mathbf{I}}_m))))
    \end{split}
\end{equation}
where $F_{DWConv1\times5}( \cdot  )$ denotes the DWConv layer with a kernel of size 1 $\times$ 5, $F_{D^3WConv1\times11}( \cdot  )$ signifies the DWConv layer with a kernel of size 1 $\times$ 11 and the dilated factor is set to 3, $F_{DWConv5\times1}( \cdot  )$ denotes the DWConv layer with a kernel of size 5 $\times$ 1, $F_{D^3WConv11\times1}( \cdot  )$ signifies the DWConv layer with a kernel of size 11 $\times$ 1 and the dilated factor is set to 3. Given that increasing the convolution kernel directly will greatly increase the parameter and computation amount, as well as increase the inference time of the model, whereas utilizing the dilated depth-wise convolutional layers with horizontal and vertical 1-D kernels will alleviate the problem. In this way, the information extraction capability of the convolutional layer is further enhanced without greatly increasing the number of computations. Moreover, to avoid potential block artifacts arising from dilation, they adopt the gate mechanism to recalibrate the generated feature maps adaptively. Finally, they use a $1\times1$ convolution to distill the output feature for extracting the representative structure information $\mathbf{I}_{n}$.
 \begin{equation}
		\mathbf{I}_{n} = F_{Conv1\times1}(\mathbf{I}_{o} \ast \mathbf{I}_{y})
\end{equation}
where $\ast$ represents the element-wise product operation.

\noindent
\textbf{Local spatial-gate branch}
Local details are important for the pleasing high-frequency reconstruction. As the SAL branch prioritizes non-local structure information exploration, they develop a simple local spatial-gate branch to capture local features simultaneously. In detail, a $3\times3$ depth-wise convolution is used to encode local information from the input features $\mathbf{I}_{x}$. Then, they use the gate mechanism to generate the enhanced local feature. Finally, they use a $1\times1$ convolution with a GELU activation to distill the output features for extracting the representative detail information $\mathbf{I}_{l}$, which is achieved by,
\begin{equation}
	\begin{aligned}
        \mathbf{I}_{o} & = F_{DWConv3\times3}({\mathbf{I}}_x)\ast \mathbf{I}_{y},\\	
        \mathbf{I}_{l} & = F_G(F_{Conv1\times1} (\mathbf{I}_{o}))
	\end{aligned}
\end{equation}
where $F_{DWConv3\times3}( \cdot  )$ denotes the DWConv layer with a kernel of size 3 $\times$ 3, $F_G$ represents GELU activation function.

\noindent
\textbf{Enhanced self-distillation module} \label{ESD}
They present an enhanced self-distillation (ESD) module to expand and refine the features derived from the SGM in spatial and channel dimensions further. 
% Fig.~\ref{fig_3}(b) shows that 
The ESD uses a $3 \times 3$ depth-wise convolutional to expand spatial and channel information. Then they use the GLUE activation function to introduce nonlinearity and extend the representation of the network. Finally, the output features are fed into a $1\times1$ convolution for further feature mixing and reducing the hidden channel back to the original input dimension. Given the input feature $\textbf{I}_{in} \in R ^ {C \times H \times W}$, this process can be formulated as,
\begin{equation}
	\begin{aligned}
        \mathbf{I}_{l} & =F_{Conv1\times1}(F_G(F_{DWConv3\times3} (\mathbf{I}_{in})))\\		
	\end{aligned}
\end{equation}

\noindent
\textbf{Training Details.}
Following previous works \cite{9857155}, they use the DF2K dataset, which consists of 800 images from DIV2K \cite{agustsson2017ntire} and 2650 images from Flickr2K \cite{lim2017enhanced} as the training dataset. A sliding window slicing operation is used to decompose each HR image into 480 $\times$ 480 patches for training. The LR images are obtained by downsampling the HR images using the MATLAB bicubic kernel function. 
\par
During the training, random rotation and horizontal flipping are used for data augmentation. The proposed  SGSDN has 8 SGSDBs, in which the number
of feature channels is set to 24. They start by pretraining the model on the DIV2K and Flickr2K datasets. The mini-batch size is set to 64. The model is trained by the ADAN optimizer \cite{xie2024adan} with $\beta _1=0.98$, $\beta _2=0.92$ and $\beta_3=0.99$, and the exponential moving average (EMA) is set to 0.999 to stabilize training. The initial and minimum learning rates are set to $5\times 10^{-3}$ and $1\times 10^{-6}$, respectively, and decay according to cosine learning rate. The model is optimized using a combination of the $L_1$ loss and an FFT-based frequency loss function \cite{cho2021rethinking} for a total of $1\times 10^6$ iterations. The size of the randomly cropped LR patches is $64\times64$. 
\par
They then conduct fine-tuning on the DIV2K dataset and the first 10k images from LSDIR \cite{li2023lsdir}. The input size is set to $96\times96$, with a batch size of 32. The fine-tuning process optimizes the model by starting with an initial learning rate of $3\times 10^{-3}$, while keeping the rest consistent with pretraining.  The fine-tuning phase encompasses a total of 100k iterations. They implemented our model on an NVIDIA RTX 3090 GPU using Pytorch.  
% 24th of Main Track
\subsection{NanoSR}

\begin{figure*}[!t]
    \centering
    \includegraphics[width=0.98\textwidth]{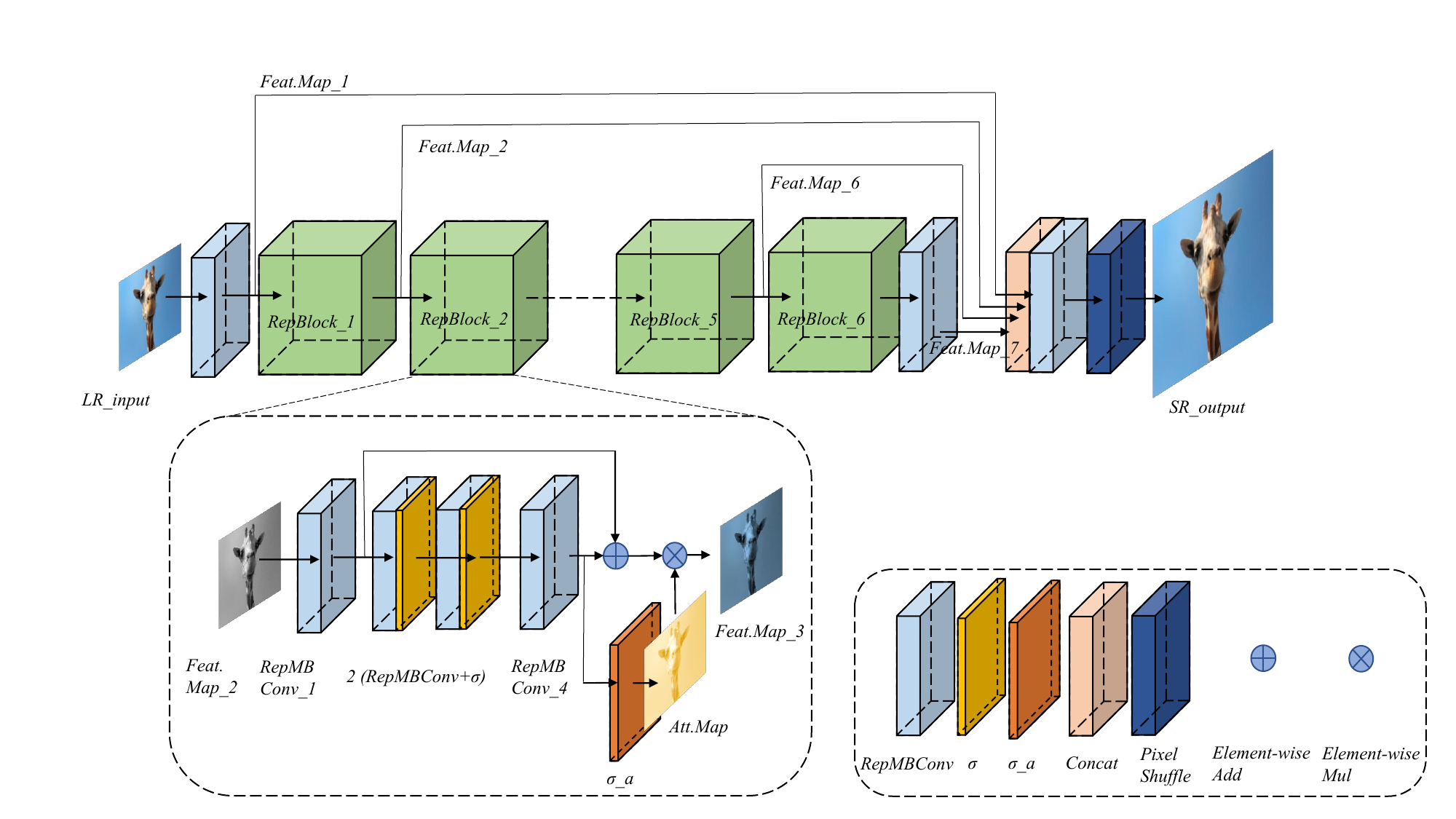}
    \caption{\textit{Team NanoSR}: The network architecture of RepRLFN}
    \label{fig:NanoSR}
\end{figure*}

\noindent \textbf{Network Architecture.}
Their network architecture is inspired by SPAN~\cite{wan2024swift} and PAN~\cite{zhao2020efficient}. While maintaining the overall design of SPAN, they replace the SPAB block with the RepBlock. The RepBlock consists of a feature extractor using reparameterized convolution and a reparameterized pixel attention module. During training, the RepBlock operates in a complex mode to achieve better quality performance but can be equivalently transformed into a simple mode with fewer parameters and FLOPs. The detailed network architecture is illustrated in ~\cref{fig:NanoSR}.

\begin{figure}[!tb]
    \centering
    \includegraphics[width=0.3\textwidth]{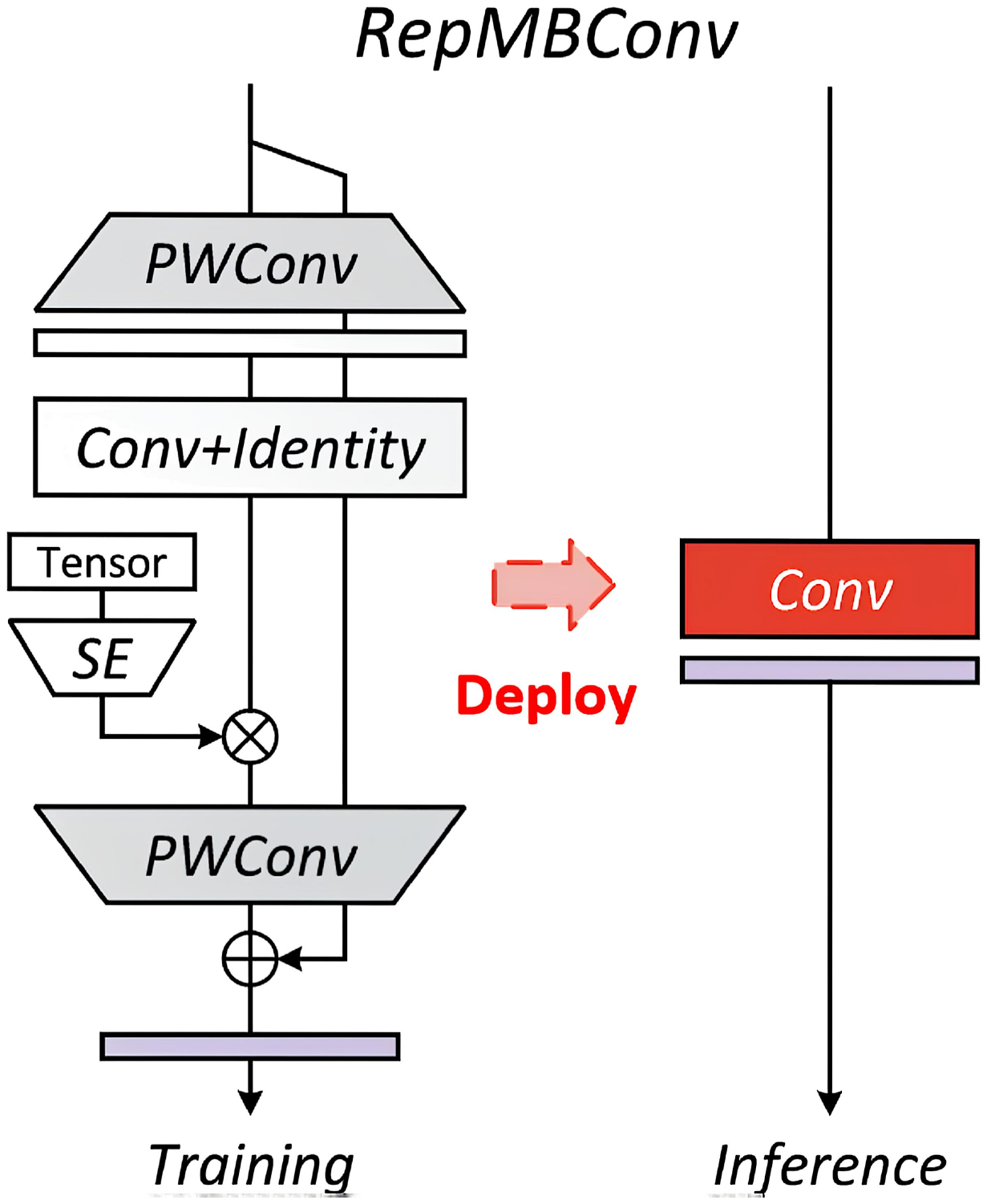}
    \caption{\textit{Team NanoSR}: The network architecture of RepRLFN}
    \label{fig:RepMBConv}
\end{figure}

\noindent \textbf{Reparameterized Convolution.}
Reparameterized convolution plays a crucial role in improving the performance of efficient CNN-based super-resolution networks. They employ the RepMBConv introduced in PlainUSR~\cite{wang2024plainusr}, and this RepMBConv forms all the convolutions in the RepBlock. In addition, RepMBConv is derived from MobileNetV3~\cite{howard2019searching} Block (MBConv). The architecture of RepMBConv is depicted in ~\cref{fig:RepMBConv}.

\noindent \textbf{Implementation Details.}
They train the model using all 85,791 image pairs from the DIV2K and LSDIR datasets. Each image pair is cropped into $480 \times 480$ sub-patches for training. During each training batch, 64 HR RGB patches of size $128 \times 128$ are randomly cropped and augmented with random flipping and rotation. The optimization objective is the $\ell_1$ loss, and they use the AdamW optimizer ($\beta_1 = 0.9$, $\beta_2 = 0.99$) to train NanoSR. The learning rate is initialized at $5 \times 10^{-4}$ and halved at $\{250\text{k}, 400\text{k}, 450\text{k}, 475\text{k}\}$ iterations within a total of 500k iterations. The proposed method is implemented using the PyTorch framework on a single NVIDIA RTX 4090 GPU.

% 25th of Main Track
\subsection{MegastudyEdu\_Vision\_AI}

\begin{figure*}[!tb]
    \centering
     \includegraphics[width=1\textwidth]{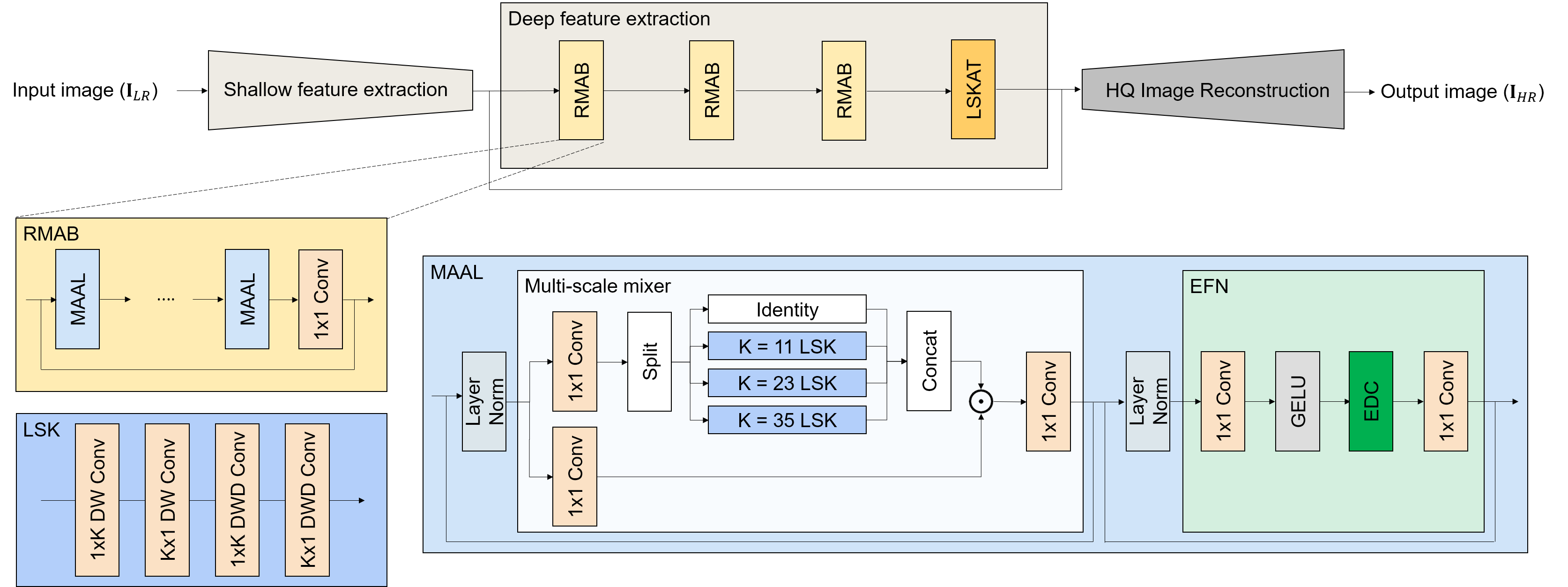}
    \caption{\textit{Team MegastudyEdu\_Vision\_AI}: Overview of multi-scale aggregation attention network.}
    \label{fig:maan}
\end{figure*}

\textbf{General Method Description.}
To effectively model long-range dependency and extensive receptive field, inspired by CFSR \cite{Wu_cfsr}, they propose the multi-scale aggregation attention network (MAAN), as illustrated in Fig. \ref{fig:maan}. MAAN reconstructs high-quality images through a shallow feature extractor, a stack of three residual multi-scale aggregation blocks (RMAB) composed of multi-scale aggregation attention layers (MAAL), a large separable kernel attention tail (LSKAT), and an image reconstruction module. Specially, MAAL captures global and local details via a multi-scale mixer and efficient feed-forward network (EFN) \cite{Wu_cfsr}. Given a low-resolution input image \( I_{LR} \in \mathbb{R}^{ 3 \times H \times W} \), shallow features such as edges, textures,  and fine details are extracted using a \( 3 \times 3 \) convolution in the shallow feature extraction stage and passed to the MAAL. As shown in Fig. \ref{fig:maan}, the MAAL processing pipeline begins with an input \(X\), applying layer normalization, followed by a \(1 \times 1\) convolution and splitting the feature map into four groups along the channel dimension:
\begin{equation}
\begin{aligned}
    V & = Conv_{1 \times 1}(X) ,\\
    F_{gate} & = Conv_{1 \times 1}(X) ,\\
    F_{id}, F_{gate1}, F_{gate2}, F_{gate3} &= \text{Split}(F_{gate}),\\
    &= F_{:g}, F_{g:2g}, F_{2g:3g}, F_{3g:}\\
\end{aligned}
\end{equation}
Here, \(F_{id}\)\ is the identity mapping without channel modification. The channel count used in convolution branches, denoted as \(g\), is determined by a ratio \(r_g\), computed as \(g = r_g C\). They set \(r_g\) to 0.25. Subsequently, each branch is processed using large separable kernel (LSK), inspired by large separable kernel attention (LSKA) \cite{lau2024large}:

\begin{equation}
\begin{aligned}
F_{id}^{'} &= F_{id}, \\
F_{gate1}^{'} &= LSK_{11,2}(F_{gate1}), \\
F_{gate2}^{'} &= LSK_{23,3}(F_{gate2}), \\
F_{gate3}^{'} &= LSK_{35,3}(F_{gate3}), \\
\end{aligned}
\end{equation}
where  \(LSK_{k,d}\) indicates the kernel size \(k\)  and dilation factor \(d\). Each LSK is composed of consecutive  \(1 \times k\) depth-wise convolution,  \(k \times 1\) depth-wise convolution,  \(1\times k\) dilated depth-wise convolution, and  \(k \times 1\) dilated depth-wise convolution. The distinct kernel sizes and dilation factors across branches effectively handle multi-scale features.

After concatenating the outputs from each branch, the combined result is integrated with \(V\) through an element-wise product. Subsequently,  \(1 \times 1\) convolution is applied to obtain the final output as follows:
\begin{equation}
F_{out} = Conv_{1 \times 1}(V \odot Concat(F_{id}^{'}, F_{gate1}^{'}, F_{gate2}^{'}, F_{gate3}^{'}))
\end{equation}
This  \(F_{out}\) is then fed into EFN \cite{Wu_cfsr}. For further EFN details, refer to CFSR \cite{Wu_cfsr}.

While CFSR \cite{Wu_cfsr} employs a 3×3 convolution tail for deep feature extraction, it has limitations in establishing long-range connections, restricting the representational capability of reconstructed features. To overcome this, they propose LSKAT inspired by the large kernel attention tail(LKAT) \cite{wang2024multi}, as depicted in Fig. \ref{fig:maan}.

\noindent
\textbf{Training Details.}
Their approach leverages DIV2K\cite{div2k}, Flickr2K\cite{lim2017enhanced}, and the first 10K portion of LSDIR\cite{li2023lsdir}. In each RMAB, the number of channels, RMABs, and MAALs are set to 48, 3, and 2-3-2, respectively. During training, they used 256 HR RGB patches with a batch size of 64. Data augmentation included random flips and rotations. Parameters are optimized using the L1 loss and the Adam optimizer\cite{kingma2014adam}. The learning rate started at \(1 \times 10^{-3}\) and decreasing to \(1 \times 10^{-6}\) using a cosine annealing scheduler. The network is trained for 1,000K iterations, implemented in PyTorch, and executed on an NVIDIA RTX 3090 GPU.

% 26th of Main Track
% \input{teams/team26_XUPTBoys/main}
% 27th of Main Track
\subsection{MILA}

\begin{figure*}
	\centering
	\includegraphics[width=0.99\textwidth]{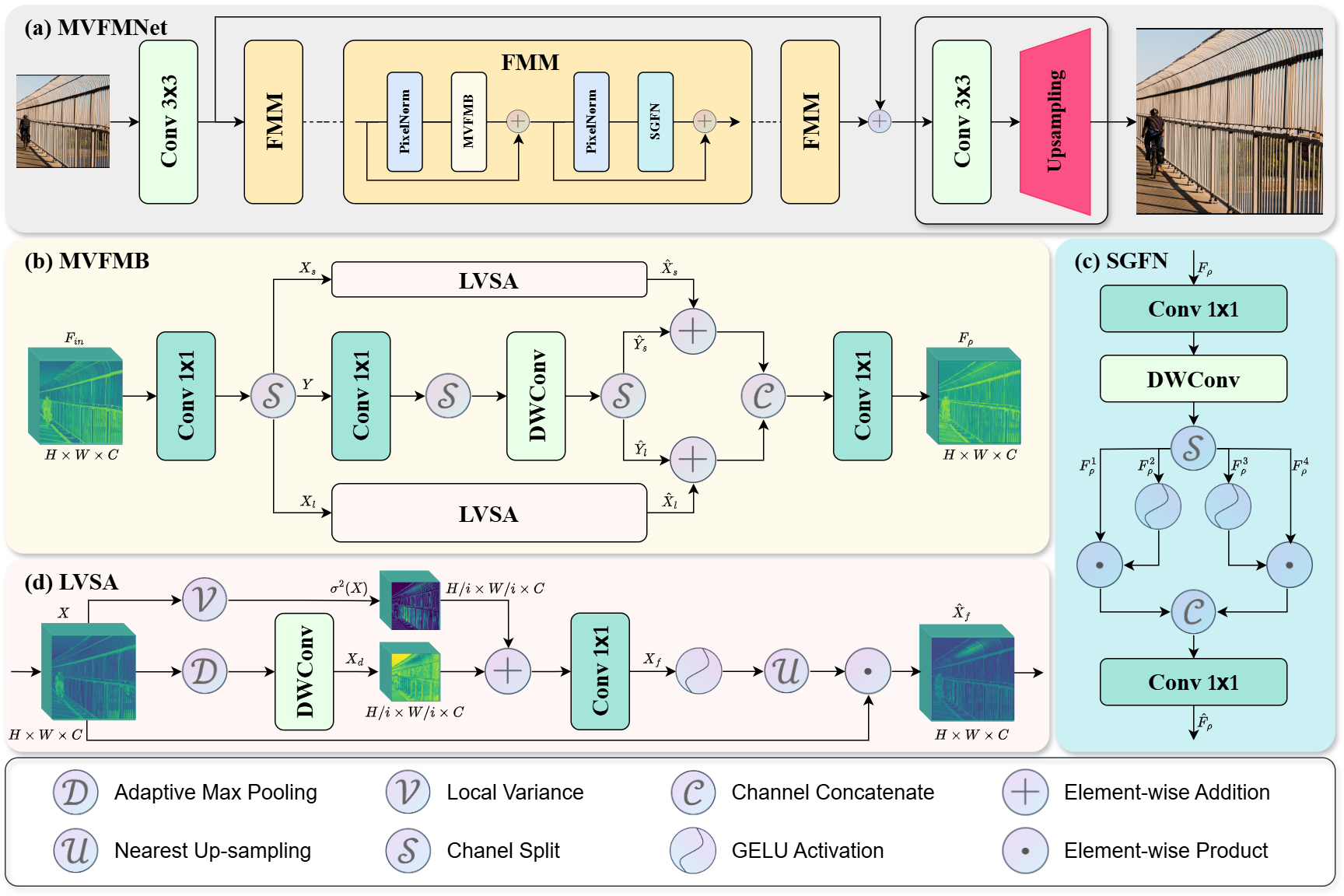}
	\caption{\textit{Team MILA:} Network architecture of the proposed MVFMNet.}  
	\label{fig:MVFMNet}
\end{figure*}

\noindent
\textbf{General Method Description.}
As shown in Figure \ref{fig:MVFMNet}, inspired by the efficient approximation of self-attention (EASA) \cite{smfanet}, they introduce local variance and design LVSA. 
Additionally, inspired by MDRN \cite{MDRN} and AGDN \cite{10678094}, they consider the impact of multi-level branches on performance. 
Therefore, they design a multi-level variance feature modulation block that incorporates non-local information with local variance perception at two different levels.
This design aims to better leverage the interplay between local and non-local features while balancing performance and model complexity.

The gated-dconv feed-forward network (GDFN) \cite{Restormer} introduces gating mechanism and depth-wise convolutions to encode information from spatially adjacent pixel positions, which is highly useful for learning local image structures to achieve effective restoration.
However, the single gating structure is relatively simple and cannot effectively capture and blend local contextual information. Therefore, they propose the symmetric gated feed-forward network.
 
\noindent
\textbf{Training Description.}
The proposed MVFMNet has 6 FMMs, in which the number of feature channels is set to 26. The details of the training steps are as follows:
\begin{itemize}
    \item[1.] Pretraining on the DF2K and the first 1k images of LSDIR datasets. HR patches of size 256 $\times$ 256 are randomly cropped from HR images, and the mini-batch size is set to 64. The model is trained by minimizing L1 loss and the frequency loss \cite{MIMO} with Adam optimizer for total 100k iterations. They set the initial learning rate to $1\times10^{-3}$ and the minimum one to $1\times10^{-6}$, which is updated by the Cosine Annealing scheme~\cite{SGDR}. 

    \item[2.] Finetuning on the DF2K and the first 1k images of LSDIR datasets. HR patch size and mini-batch size are set to 256 $\times$ 256 and 64, respectively. The model is fine-tuned by minimizing the L2 loss function. The learning rate is initialized at $2\times 10^{-5}$ and gradually decreased to $1\times 10^{-8}$ over 500k iterations using the Cosine Annealing scheme.
\end{itemize}

% 28th of Main Track
\subsection{AiMF\_SR}

\begin{figure*}
    \centering
    \includegraphics[width=0.95\linewidth]{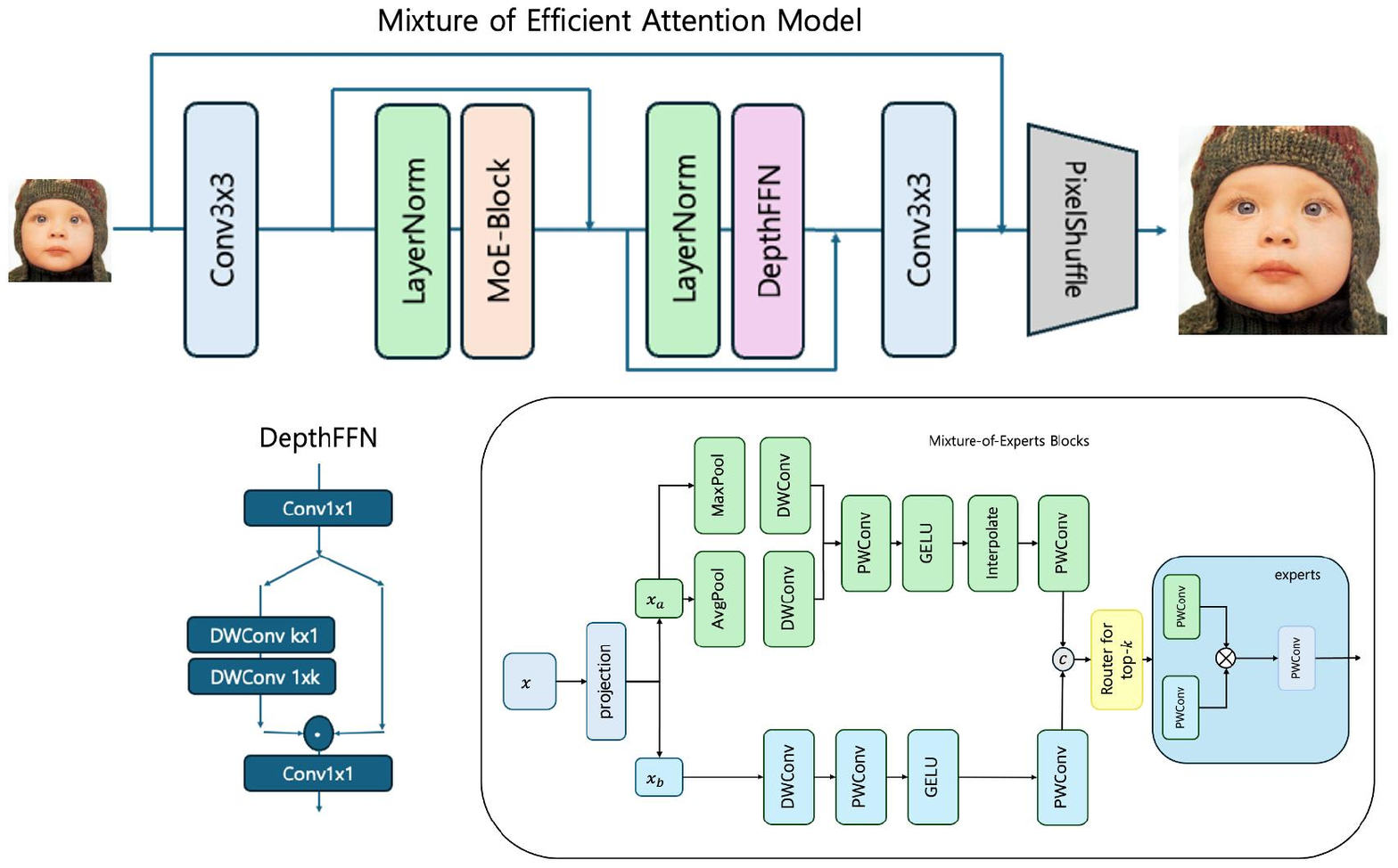}
    \vspace{-12mm}
    \caption{\textit{Team AiMF\_SR}: Main Figure of Proposed Architecture, Mixture of Efficient Attention.}
    \label{fig:main_figure}
\end{figure*}

\noindent
\textbf{Method Details.}
They propose a novel Mixture of Efficient Attention (MoEA) architecture for efficient super-resolution tasks. The architecture includes a shallow feature extractor, multiple Feature Representation Modules (FRMs), and an efficient reconstruction and upsampling module. Initially, a shallow 3×3 convolutional layer reduces computational load, generating compact feature representations. Deep feature extraction employs transformer-inspired blocks with pre-normalization, incorporating Mixture-of-Experts (MoE) Blocks~\cite{zamfir2024see} for efficient attention and Depth Feed Forward Networks (DepthFFN) for capturing depth-wise interactions. Details of the architecture can be seen in ~\cref{fig:main_figure}.

The MoEBlock consists of two parallel feature pathways (~\cref{fig:main_figure}). The input features $x$ are first projected into two distinct feature sets $x_a$ and $x_b$ using a pointwise convolution. The first branch, $x_a$, undergoes both adaptive average and max pooling followed by depth-wise convolutions. The pooling is done in scale of 8~\cite{zheng2024smfanet}. These pooling layers followed by depth-wise convolutions serves as efficient attention-like mechanism. Then, it combines these features through element-wise addition, nonlinear activation (GELU), and interpolation. The second branch, $x_b$, is processed via depth-wise and pointwise convolutions with GELU activation.

\begin{equation}  
\begin{aligned}
x_a &= \text{DWConv}(\text{AvgPool}(x_a)) + \text{DWConv}(\text{MaxPool}(x_a)), \\
x_a' &= \mathcal{U}(\mathcal{G}(\text{PWConv}(x_a))), \\
x_a' &= \text{PWConv}(x_a'), \\
x_b' &= \mathcal{G}(\text{PWConv}(\text{DWConv}(x_b))), \\
x_{ab} &= \mathcal{C}(x_a', x_b').
\end{aligned}
\end{equation}  

\noindent
where $x_a, x_b$ are concatenated and passed through the Router (gating network), $\mathcal{R}$, which adaptively selects the top-$k$ expert paths based on the channel-wise global average-pooled features in the MoE-layer. Each selected expert independently processes $x_a'$ and $x_b'$ through pointwise convolutions, multiplies them element-wise, and applies a final convolution for feature integration:  

\begin{equation}  
\begin{aligned}
\text{logits} &= \mathcal{R}(x_{ab}), \\
x'_a, x'_b &= \text{TopK}(\text{Softmax}(\text{logits}))  \\
\text{Expert}(x_a', x_b') &= \text{PWConv}[\text{PWConv}(x_a') \times \text{PWConv}(x_b')]  
\end{aligned}
\end{equation}  

Multiple FRMs (LayerNorm-MoEBlock-LayerNorm-DepthFFN sequences) are stacked for deep feature extraction. For reconstruction, global contextual features from deep extraction combine with shallow features via residual connections, followed by PixelShuffle-based upsampling to produce high-resolution outputs. The model uses GELU activation, Layer Normalization. Their MoE layer dynamically routes features across $\text{num\_experts}=3$, selecting the top $k=1$ experts at training time, allowing a flexible and adaptive processing pipeline tailored specifically to input feature characteristics.

\noindent
\textbf{Training Strategy.}
The model is trained and tested on BasicSR~\cite{basicsr} setting. First, the model is initially trained on DIV2K\_LSDIR\_x2, then further finetuned with DIV2K\_LSDIR\_x3 dataset for 500,000 iterations respectively, in which these scales are  made with bicubic downsampling. The x4 scale model is finetuned on top of the x3 model over 500,000 iterations with the initial learning rate of 1 $\times 10^-3$ using the Adam optimizer. The learning rate decayed at iterations [250,000, 400,000, 450,000, 475,000]. The training pipeline included data augmentations such as random horizontal flips, vertical flips and rotations. the model is optimized using L1 Loss and Fast Fourier Transform (FFT) Loss~\cite{sun2022shufflemixer} with 1.0 and 0.1 weights, respectively. All reported implementations are  carried out using Python (version 3.9) programming language and PyTorch Framework, utilizing one RTX4090, 24GB VRAM and 16-core CPU. Training is conducted over approximately 2~3 days with a single GPU of batch size of 16. 
% 29th of Main Track
% \input{teams/team33_EagleSR/main}   % NO Update
% 30th of Main Track
\subsection{BVIVSR}

\begin{figure*}[!tb]
    \centering
    \includegraphics[width=1\linewidth]{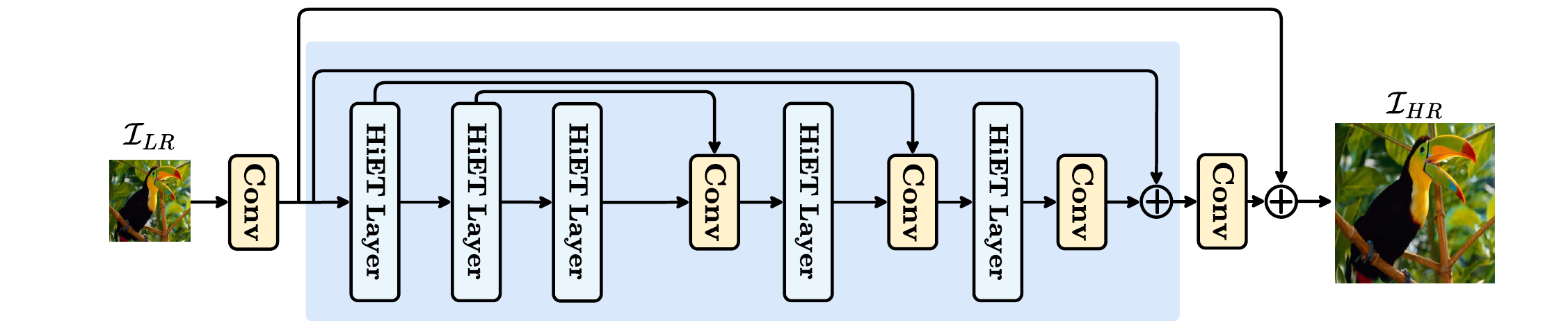}
    \caption{\textit{Team BVIVSR}: The structure of the method.}
    \label{fig:bvivsr}
    \vspace{-10pt}
\end{figure*}

\textbf{Method Description.} Their solution is built on the advances in state-of-the-art single-image super-resolution (SISR) methods \cite{zhu2023attention,qiu2023dual,zhang2023ntire,chen2024ntire,conde2024aim}, particularly the efficient Transformer-based models \cite{zhang2024hit, jiang2025c2d}, the continuous super-resolution approaches, such as HIIF \cite{jiang2024hiif,jiang2025c2d}, and the knowledge distillation strategies \cite{jiang2024mtkd, jiang2024compressing, jiang2024rtsr}. They employ an efficient Transformer-based network architecture, as illustrated in \cref{fig:bvivsr}, where the core component is the Hierarchical Encoding Transformer (HiET) layer. The HiET layer was first proposed in \cite{jiang2025c2d} and it is specifically designed to capture rich structural dependencies across various regions of the image, enabling the model to handle complex visual patterns effectively. To enhance the capacity of the model for multi-scale feature representations, each HiET layer is set with different window sizes, allowing it to attend to both local and global contexts. Furthermore, the overall architecture incorporates a modified U-Net structure, where skip connections are introduced between symmetric HiET layers at different depths. This design facilitates efficient multi-level feature fusion and ensures better preservation and reconstruction of fine-grained details in the super-resolved outputs. In addition, they also apply the multi-teacher knowledge distillation strategy \cite{jiang2024mtkd} to improve the performance of the lightweight C2D-ISR model, where SRFormer \cite{zhou2023srformer}, MambaIR \cite{guo2024mambairv2} and EDSR \cite{lim2017enhanced} are employed as teacher networks.

\noindent
\textbf{Training Details.} They use the DIV2K \cite{timofte2017ntire}, 1000 2K images from BVI-AOM \cite{nawala2024bvi}, Flickr2K \cite{lim2017enhanced} and 5000 images from LSDIR\cite{li2023lsdir} as training dataset. For evaluation, they follow common practice and employ the DIV2K validation set (containing 100 images) \cite{timofte2017ntire}. The maximum learning rate is set to \( 4 \times 10^{-4} \). The learning rate follows a cosine annealing schedule, gradually decreasing after an initial warm-up phase of 50 epochs. They use L1 loss and the Adam \cite{kingma2014adam} optimization during training. Training and testing are implemented based on 4 NVIDIA 4090 GPUs. The model comprises 154.8K parameters with an input size of $64\times64\times3$ and it was trained for 1000 epochs with 16 batch sizes per GPU. 
The training of their solution contains five stages: 

\begin{itemize}
\item Training the teacher networks, including SRFormer \cite{zhou2023srformer}, MambaIR \cite{guo2024mambairv2} and EDSR \cite{lim2017enhanced}, by using the original settings in their papers;

\item The teacher aggregation of multi-teacher knowledge distillation (MTKD) strategy \cite{jiang2024mtkd} was adapted to the above teacher networks to obtain an enhanced teacher network; 

\item Training the lightweight  C2D-ISR model \cite{jiang2025c2d} on continuous scales i.e, from $\times$2 to $\times$ 4, to learn the correlation between multiple scales and better recover high-frequency details;

\item The learned C2D-ISR model was distilled by the MTKD strategy \cite{jiang2024mtkd} with their enhanced teacher network to obtain the enhanced student model;

\item Finetuning the enhanced student model by increasing the patch size from 64 $\times$ 64 to 128 $\times$ 128.
\end{itemize}
% 31st of Main Track
% --- 3rd Sub-Track: FLOPs & 2nd Sub-Track: Params
% \input{teams/team13_HannahSR/main}
% 32nd of Main Track
% --- 1st Sub-Track: FLOPs & 1st Sub-Track: Params
% \input{teams/team20_VPEG_C/main}
% 33rd of Main Track
\subsection{CUIT\_HTT}

\begin{figure*}
    \centering
    \includegraphics[width=\textwidth]{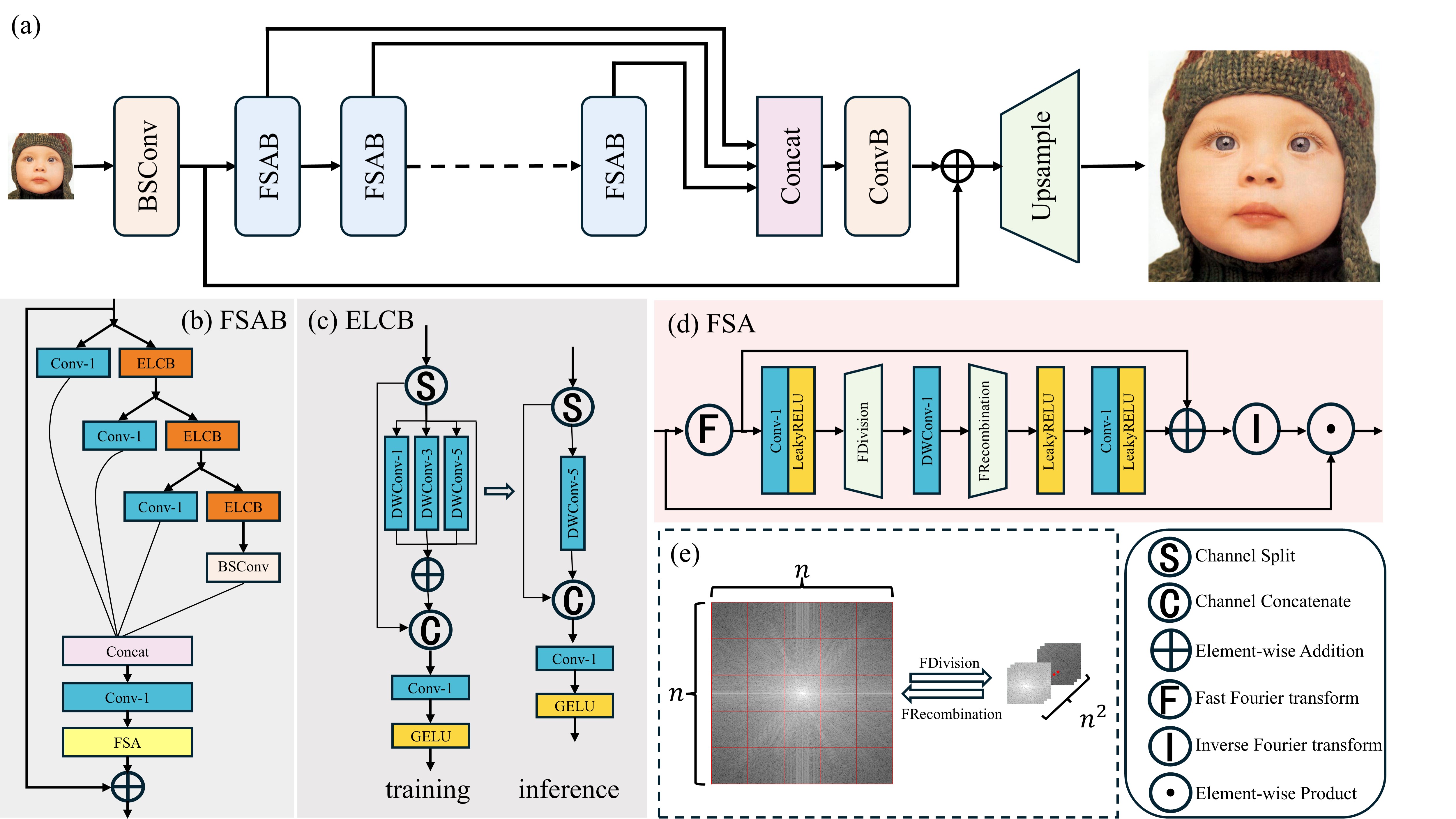}
    \caption{\textit{Team CUIT\_HTT}: Schematic Diagram of the Method. 
    (a) Overall Architecture of the Model;
    (b) Frequency-Segmented Attention Block (FSAB);
    (c) Schematic of the Enhanced Large-kernel Convolution Block (ELCB);
    (d) Mechanism of Frequency-Segmented Attention (FSA);
    (e) Frequency Division and Frequency Recombination.}
    \label{fig:team17_network}
\end{figure*}

\textbf{General Method Description.}
The overall architecture of the proposed method is illustrated in \cref{fig:team17_network}(a), which consists of three main components: the shallow feature extraction module, the deep feature extraction module, and the reconstruction and upsampling module. The shallow feature extraction module employs a BSConv~\cite{BSConv} module to extract low-level features such as edges and textures from the input image $ I^{in}\in\mathbb{R}^{3\times H\times W} $, mapping it to the feature space $f^0\in\mathbb{R}^{C\times H\times W}$ for further processing. The extracted shallow features are then fed into the deep feature extraction module, which is composed of multiple Frequency-Segmented Attention Blocks (FSABs) designed in this work. The outputs of each FSAB are concatenated along the channel dimension and adjusted using a convolutional module group, constituting the deep feature extraction process. As shown in ~\cref{fig:team17_network}(b), the FSAB structure includes a Concat operation for channel concatenation and a ConvB module group, which consists of a $1\ \times1$ convolution, a GELU activation function, and a BSConv stacked sequentially. Finally, the output of the shallow feature extraction module is added element-wise to the output of the deep feature extraction module via a skip connection and passed to the reconstruction and upsampling module. This module upsamples the feature space information $f^{out}\in\mathbb{R}^{C\times H\times W}$and maps it to the high-resolution output image $I^{SR}\in\mathbb{R}^{3\times s c a l e\times H\times s c a l e\times W}$, where $\mathrm{scale}$ is the upscaling factor. In this work, the PixelShuffle method is utilized for upsampling.

The Frequency-Segmented Attention Block (FSAB) primarily consists of an information distillation architecture for local feature processing and the proposed Frequency-Segmented Attention (FSA) mechanism for global feature processing. The overall architecture of FSA is illustrated in Fig.~\ref{fig:team17_network} (d). The input feature map is first transformed into the frequency domain via the Fast Fourier Transform (FFT), enabling global processing in the spatial domain through frequency domain operations. Inspired by windowed attention, the FDivision operation partitions the frequency spectrum into multiple windows, which are concatenated along the channel dimension. A grouped convolution is then applied to process features in different frequency ranges using distinct weights. Subsequently, the FRecombination operation reassembles the segmented frequency windows back into the spectrum. A convolutional layer is applied, and the result is added element-wise to the original spectrum. Finally, the Inverse Fast Fourier Transform (IFFT) is used to convert the processed features back to the spatial domain, and the output is obtained through element-wise multiplication with the original input. As for the information distillation architecture, they adopt the structure of the Residual Feature Distillation Block (RFDB) from RFDN~\cite{RFDN}, as shown in Fig.~\ref{fig:team17_network}. (b). However, they replace the convolutional layers with Enhanced Large-kernel Convolution Blocks (ELCB). This module employs large-kernel depthwise convolution on half of the channels and pointwise convolution on the full channels, achieving a large receptive field without significantly increasing the number of parameters. Additionally, structural reparameterization is utilized during training, where multiple branches with different receptive fields are employed. During inference, these branches are equivalently replaced with a single large-kernel convolution module, thereby enhancing the model's learning capability without increasing inference cost.

\noindent
\textbf{Train details.} They utilize the DIV2K~\cite{agustsson2017ntire} and Flickr2k~\cite{flickr2k} dataset and the first 10K images from the LSDIR~\cite{li2023lsdir} dataset as the training set for their model. 
During training, the dataset undergoes random horizontal flipping and 90° rotation. The mini-batch size and input patch size are set to 64 and 64×64, respectively. The model is optimized using the L1 loss function and the Adam optimizer, with an initial learning rate of $5\times 10^{-3}$. The learning rate follows a cosine annealing decay schedule over a total of 1000K iterations. Subsequently, the model is fine-tuned using the L2 loss to achieve improved performance. Training is conducted using PyTorch 1.12.1 on a Tesla P100 16G GPU.
% 34th of Main Track
\subsection{GXZY\_AI}
\begin{figure*}
  \centering
  \includegraphics[width=.7\textwidth]{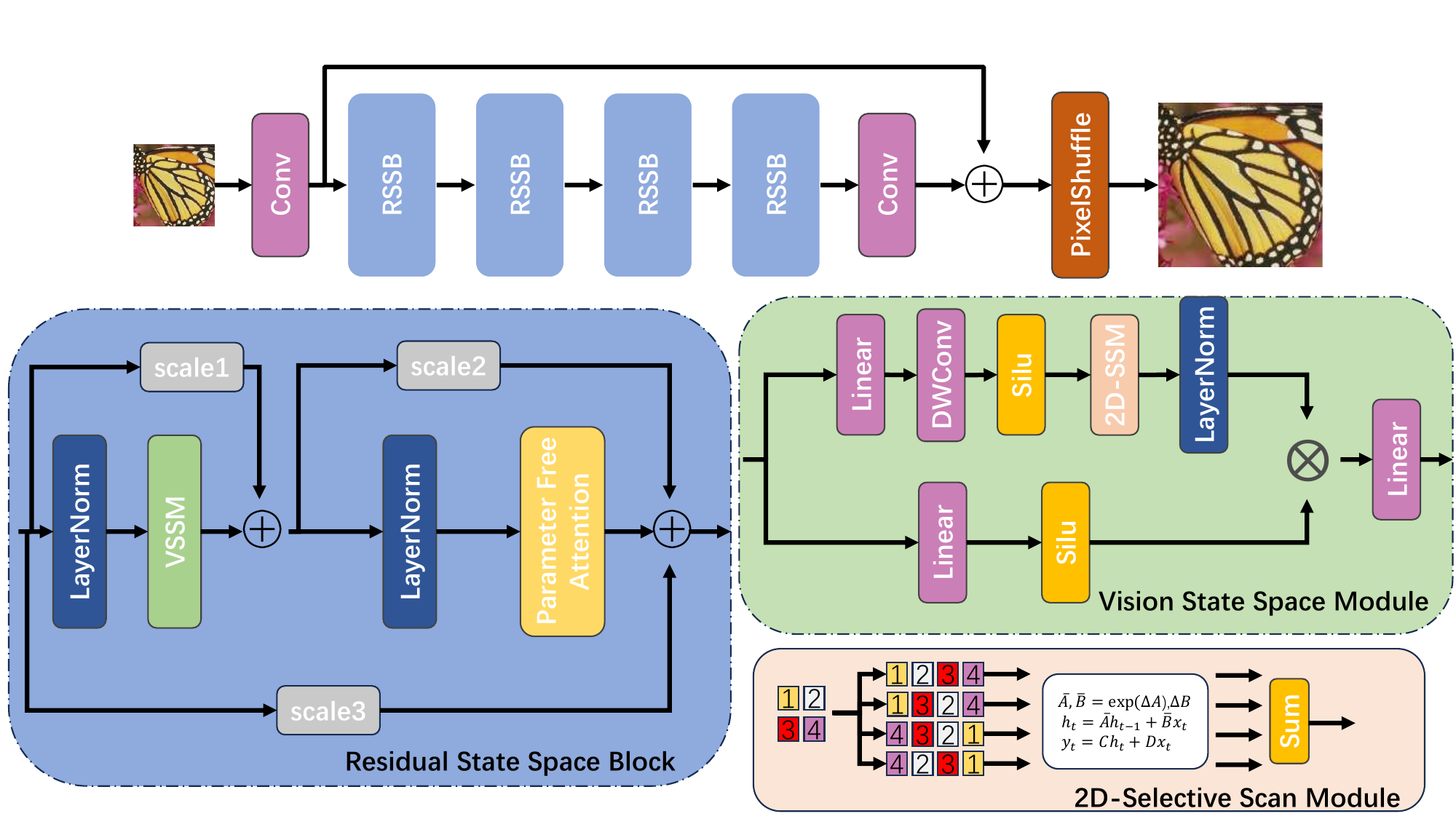} 
  \caption{\textit{Team GXZY\_AI}: The structure of PFVM.} 
  \label{GXZY Figure 1} 
\end{figure*}

\begin{figure*}
  \centering
  \includegraphics[width=.7\textwidth]{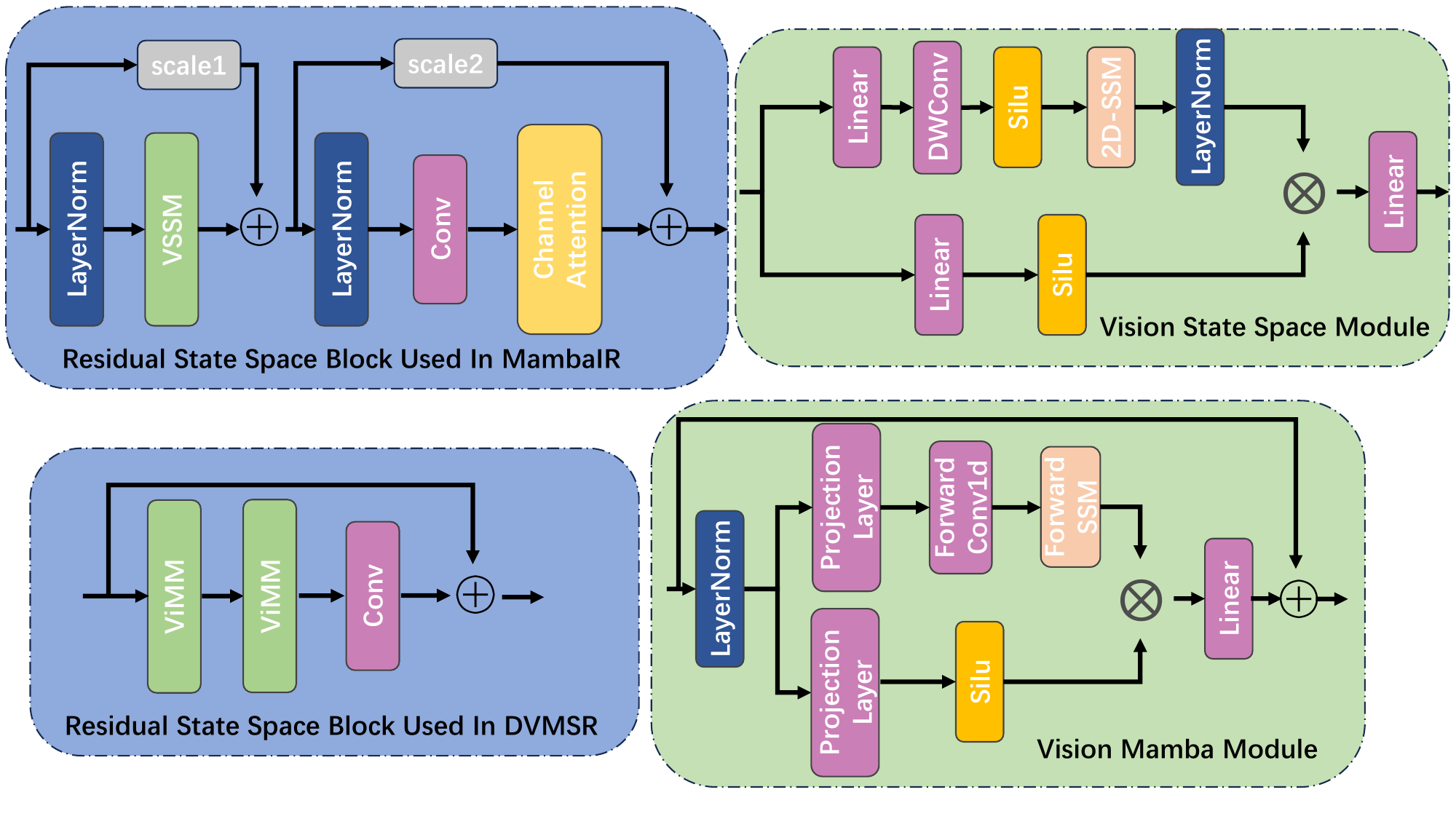} 
  \caption{ \textit{Team GXZY\_AI}: The structural details of MambaIR and DVMSR.} 
  \label{GXZY Figure 2} 
\end{figure*}

\textbf{General Method Description.}
The GXZY AI team proposed a Parameter-free Vision Mamba, as shown in \cref{GXZY Figure 1}. The work is inspired by MambaIR~\cite{guo2025mambair}, SPAN~\cite{wan2024swift} and DVMSR~\cite{Lei_2024_CVPR}, PFVM consists of three parts, shallow feature extraction, deep feature extraction and reconstruction module. Shallow feature extraction is achieved by $3\times3$ convolution, followed by the use of stacked Residue State Space Blocks (RSSBs), which contain the Vision State Space Module (VSSM) to extract deeper features through the capability of Mamba long-range modeling. Then the shallow and deep features are aggregated by a $3\times3$ convolution along with residual concatenation, and finally up-sampling is achieved through a sub-pixel convolutional layer to reconstruct the high resolution image. 

As shown in \cref{GXZY Figure 2}, different from the RSSB used in DVMSR, PFVM does not use stacked ViMM modules, but follows the design paradigm of the RSSB in MambaIR, which differs from MambaIR in that 3-residue branching is used in order to maximize the ability of residual learning. In order to obtain better PSNR with approximate inference time, the convolution layer adopts the bottleneck structure, and the channel attention used in MambaIR is replaced by a parameter-free attention.

\noindent
\textbf{Training Strategy.}
In the training phase, the GXZY AI team uses the LSDIR~\cite{li2023lsdir} dataset for training and the DIV2K~\cite{Agustsson_2017_CVPR_Workshops} validation set for validation. The images in the training set are first cropped with a step size of 240 and a size of 480 to get a series of cropped images. The model was trained on 2 NVIDIA RTX 3090 GPUs. The details of the training steps are as follows: 
\begin{enumerate}
    \item The HR images are randomly cropped to size 192, and the dataset is augmented using random flipping and rotation. The model is trained from scratch with a batch size set to 64, using the Adam optimizer with the learning rate set to 0.0001, $\beta_1=0.9$, $\beta_2=0.99$, and a MultiStepLR scheduler with the learning rate halved for every 200,000 iterations for a total of 1,000,000 iterations. The loss function uses L1 loss.
    \item On the basis of the first step, the model with the optimal PSNR on the DIV2K validation set is loaded as the pre-training model, the size of HR image cropping is adjusted to 256, the learning rate is 0.0002, the learning rate is halved for every 100,000 iterations, and the loss function is still used for 1,000,000 iterations with L1 loss.
\end{enumerate}
% 35th of Main Track
% \input{teams/team16_SCMSR/main}      % NO Update
% 36th of Main Track
\subsection{IPCV}

This team uses HiT-SR: Hierarchical Transformer for Efficient Image Super-Resolution \cite{zhang2024hitsrhierarchicaltransformerefficient} for this challenge.  The Hierarchical Transformer for Efficient Image Super-Resolution (HiT-SR) is a deep learning model designed to upscale low-resolution (LR) images into high-resolution (HR) outputs while maintaining efficiency and high-quality reconstruction. Unlike traditional convolutional neural networks (CNNs), which struggle to capture long-range dependencies, HiT-SR employs a hierarchical self-attention mechanism that efficiently processes multiscale image features. This allows the model to integrate local and global information, improving image detail reconstruction while reducing computational costs.

At the core of the network is a hierarchical feature learning process, where image features are extracted and refined progressively through multiple stages. Instead of applying full-resolution self-attention, which is memory intensive, HiT-SR reduces token complexity using patch merging and downsampling modules, allowing efficient computation without loss of essential information. The model further refines these hierarchical features through multiscale self-attention mechanisms, ensuring that fine-grained details and global structures are effectively captured.

For the final super-resolution reconstruction, HiT-SR aggregates and progressively upsamples the processed features. This multistage refinement approach ensures that high-frequency details are preserved while preventing artifacts common in naive upsampling techniques. The resulting HR image maintains sharp edges, realistic textures, and minimal distortions. They have used available pre-trained model weights \cite{HiT-SR} on the low resolution images of the test data set and predicted high resolution images.
% 37th of Main Track
\subsection{X-L}

\begin{figure*}[!tb]
    \centering
    \includegraphics[width=40em]{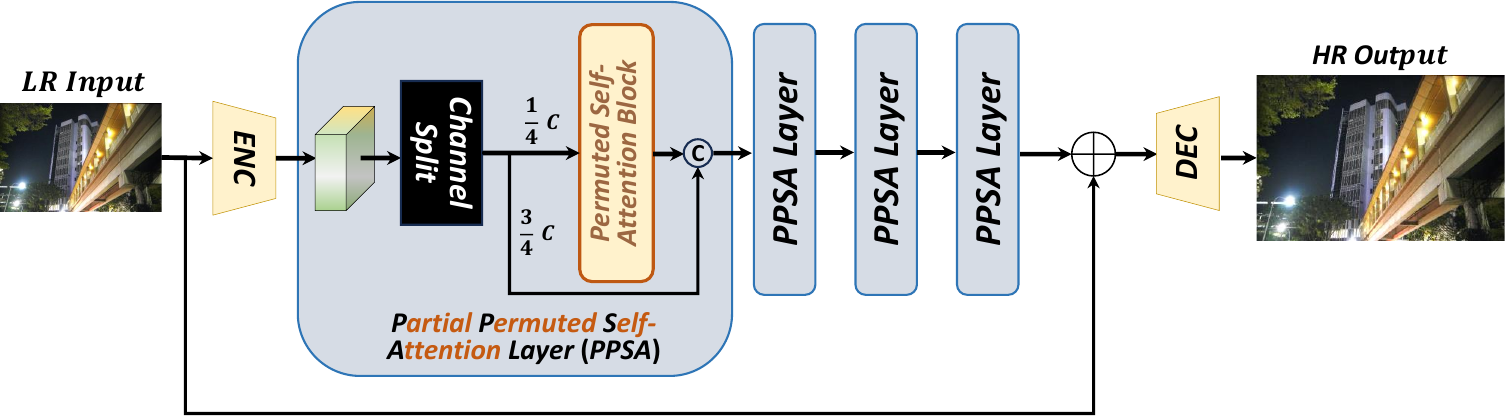}
    \caption{\textit{Team X-L}: Overview of the proposed PPSA-Net.}
    \label{fig:method2}
\end{figure*}

\textbf{General Method Description.} 
Their proposed partial permuted self-attention network (PPSA-Net) is shown in ~\cref{fig:method2}. PPSA-Net is inspired by two works: SRFormer~\cite{zhou2023srformer} and PartialConv~\cite{chen2023run}. 
SRFormer is a lightweight super-resolution (SR) approach, but it inevitably still has significant redundancy in feature dimensions.
To address this, they combine the strengths of PartialConv to further reduce the complexity and the computational cost. 
Specifically, they use a feature encoder to process the low-resolution image and feed it to four partial permuted self-attention (PPSA) layers, before finally feeding it into a feature decoder to obtain the final result.
In more detail, within each PPSA layer, they use channel split to divide the original features into two sub-features: one comprising 1/4 of the channels and the other comprising 3/4 of the channels. 
The 1/4 sub-feature is processed by a permuted self-attention block~\cite{zhou2023srformer}, while the 3/4 sub-feature remains unchanged.
After processing, the two sub-features are concatenated back together. This design allows us to efficiently reduce computational overhead while maintaining the model's ability to capture both local and global information, leading to high-quality SR results.

\noindent
\textbf{Training details.} 
They follow the same training procedure as SRFormer~\cite{zhou2023srformer}. 
However, they conduct their training using a single NVIDIA 4090 GPU. 

% 38th of Main Track
\subsection{Quantum\_Res}

\begin{figure}[!tb]
\centering
\includegraphics[width=\linewidth]{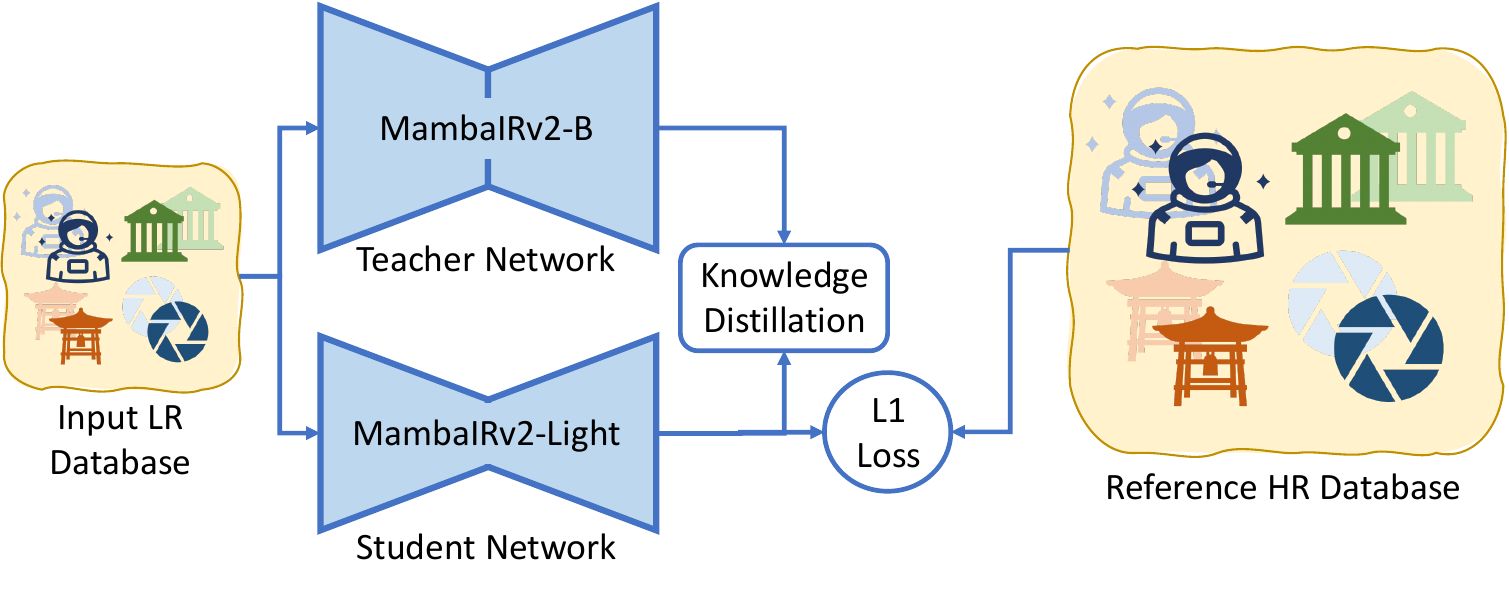}
\vspace{-0.5cm}
\caption{\textit{Team Quantum\_Res}: The overall pipeline of efficient super-resolution approach, which employs a student-teacher training paradigm. The high-capacity Teacher Network (MambaIRv2-B) learning is transferred to the lightweight Student Network (MambaIRv2-Light) using knowledge distillation. The student network is optimized using L1 loss to ensure accurate super-resolution while maintaining efficiency. The input low-resolution (LR) database serves as the training input, guiding the student model to achieve high-fidelity reconstruction with reduced computational complexity.}
 \vspace{-0.5cm}
\label{fig:34model}
\end{figure}

\noindent \textbf{Method Details.}
In this work, they propose a novel student-teacher framework for super-resolution, as shown in \cref{fig:34model} that enables a lightweight student model to achieve better performance comparable to heavier models. Specifically, to adopt this architecture, they used MambaIRv2-Light~\cite{guo2024mambairv2} as the student model, while MambaIRv2-base~\cite{guo2024mambairv2} serves as the teacher. While they use MambaIRv2-light as an efficiency, their key contribution is demonstrating that a guided student-teacher learning strategy can significantly improve SR performance while keeping model complexity low. \cite{wan2023swift}

The student model extracts the initial low-level features from the input low-resolution image using the \( 3 \times 3 \) convolutional layer. The core of the network comprises a series of Attentive State-Space Blocks (ASSBs)~\cite{guo2024mambairv2}to capture long-range dependencies efficiently. For each block, residual connections are used to facilitate stable gradient propagation. Finally, a pixel-shuffle-based upsampling module reconstructs the final high-resolution image. \cite{guo2024mambairv2}

The teacher model, MambaIRv2, follows the same architectural design but with increased depth and wider feature dimensions. This model has significantly more parameters and serves as an upper-bound reference for the student.

\noindent \textbf{Teacher-Guided Inference.} The teacher model remains frozen throughout training and is only used as a qualitative reference to validate architectural choices and improvements. The student model inherits refined architectural principles from the teacher rather than weight transfer or feature alignment. This allows the student to retain its original lightweight nature while benefiting from structural knowledge obtained from a larger-capacity model \cite{wan2023swift}.

\noindent \textbf{Inference Strategy.}  
During inference, an efficient patch-based processing method is applied to handle high-resolution images. Given an input image, it is divided into overlapping patches. Each patch is processed independently by the student network, and final predictions are blended using a weighted averaging scheme to ensure seamless reconstruction. \cite{guo2024mambairv2}

\noindent \textbf{Training Details.}
The student model is initialized using pre-trained weights of MambaIRv2-light. The teacher model is loaded with pre-trained weights from a high-performing MambaIRv2-base variant. 
Fine-tuning was performed on DIV2K and LSDIR, with the number of feature channels set to 48. The training was conducted on patches of size \( 192 \times 192 \) extracted from high-resolution images, using a batch size of 8. The model is finetuned by minimizing the L1 loss function using the Adam optimizer. 
The initial learning rate is set to \( 1 \times 10^{-5} \) and is reduced when training iterations reach specific milestones, following a MultiStepLR decay strategy with a factor of 0.5. The total number of iterations is 150K. The teacher model is only used as a reference for guiding architectural refinement and remains frozen throughout the training.

% --- Teams without ranking
\subsection{SylabSR}

\begin{figure}[t]
  \centering
  \includegraphics[width=1.0\linewidth]{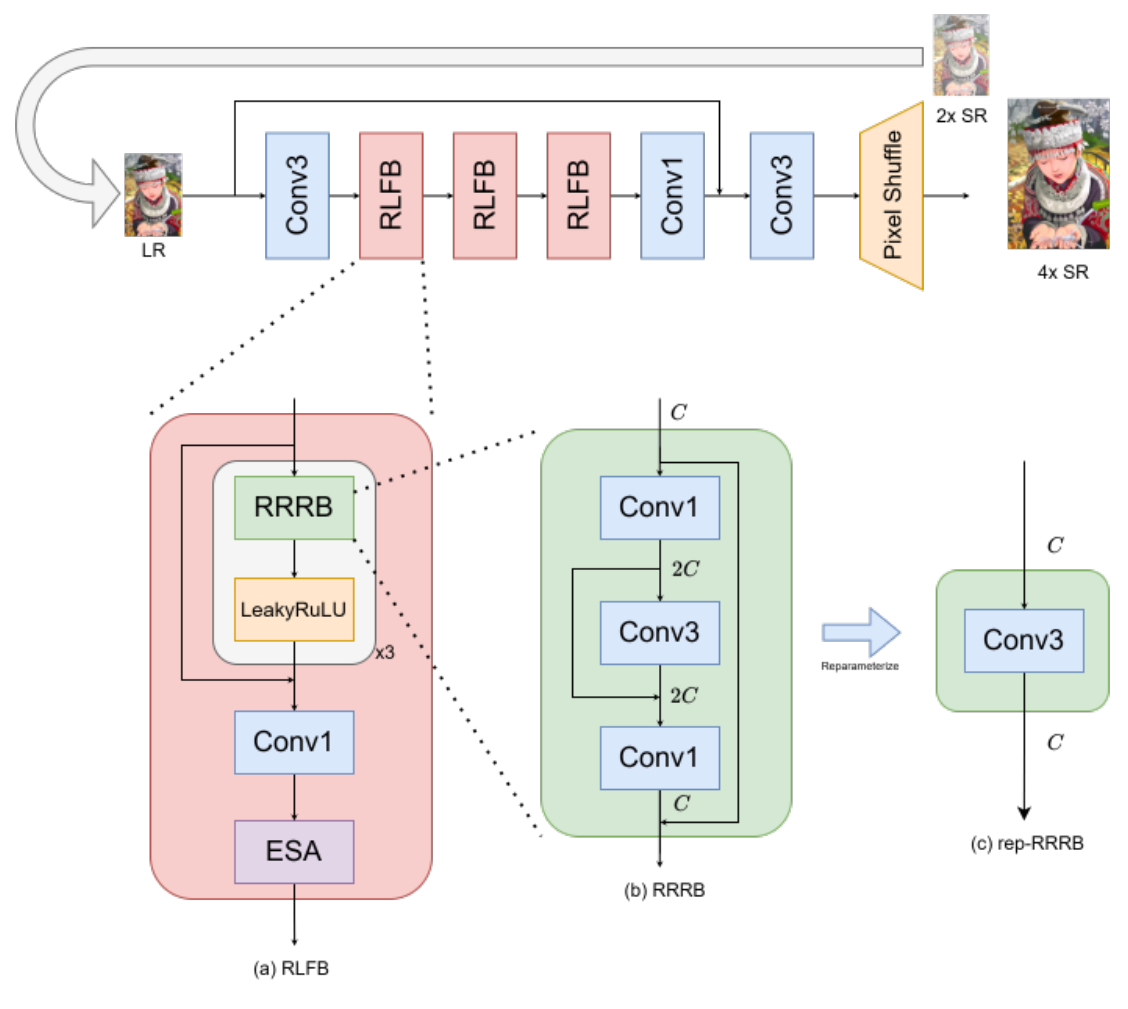}
  \caption{\textit{Team SylabSR}: The structure of (up) AR-RLFN, (a) RLFB, (b) RRRB and (c) its reparameterization.}
  \label{fig:structure35}
\end{figure}

\noindent \textbf{Method.} Inspired by RLFN ~\cite{kong2022residual} and VARSR ~\cite{qu2025varsr}, they propose an AutoRegressive Residual Local Feature Network (AR-RLFN) to implement a two-stage super-resolution framework. Specifically, they build a lightweight version of RLFN targeting 2× super-resolution, meaning that the final 4× SR image is generated from an intermediate 2× SR image produced by the same model. The overall framework of AR-RLFN is shown in \cref{fig:structure35}. Although the model needs to be run twice, the 2× SR task requires significantly fewer parameters and FLOPs compared to the original one, making the approach efficient overall.

The modified structure of RLFN is further inspired by R2Net~\cite{ren2024ninth}. Benefiting from the two-stage strategy, their model is able to operate with fewer parameters. In their framework, they adopt three Residual Local Feature Blocks (RLFBs) with a reduced number of channels compared to the original version. Additionally, they replace ReLU with LeakyReLU to mitigate gradient vanishing. For reparameterization, they employ the Residual-in-Residual Rep Block (RRRB)~\cite{du2022fast} for improved compression, which reduces the number of parameters during inference by approximately 45\%.

\noindent \textbf{Training Strategy.} They train their network on DIV2K ~\cite{timoftediv2k} and LSDIR ~\cite{li2023lsdir} datasets, and augment the training data using random flipping and rotation. The training process is divided into three stages:
\begin{enumerate}
    \item HR patches of size 512$\times$512 are randomly cropped from the ground truth DIV2K images. In this stage, the model performs 2$\times$ super-resolution. The number of channels in the RRRB is set to 12, and the batch size is set to 32. They use the Adam optimizer to minimize the Charbonnier loss, with the learning rate set to 5e$^{-4}$. The training runs for 100k iterations, and the learning rate is halved every 20k iterations.
     \item HR patches of size 256$\times$256 are randomly cropped from the ground truth DIV2K images. The model again performs 2$\times$ super-resolution in this stage. The remaining configurations are the same as in Stage 1.
    
    \item HR patches of size 512$\times$512 are randomly cropped from both the DIV2K and LSDIR datasets. In this stage, they use the Adam optimizer to minimize MSE loss, with the learning rate set to 2e$^{-4}$. The training runs for 50k iterations, and the learning rate is halved every 10k iterations.

\end{enumerate}
\subsection{NJUPCA}

\begin{figure}[!tb]
\begin{center}
\includegraphics[width=1 \linewidth]{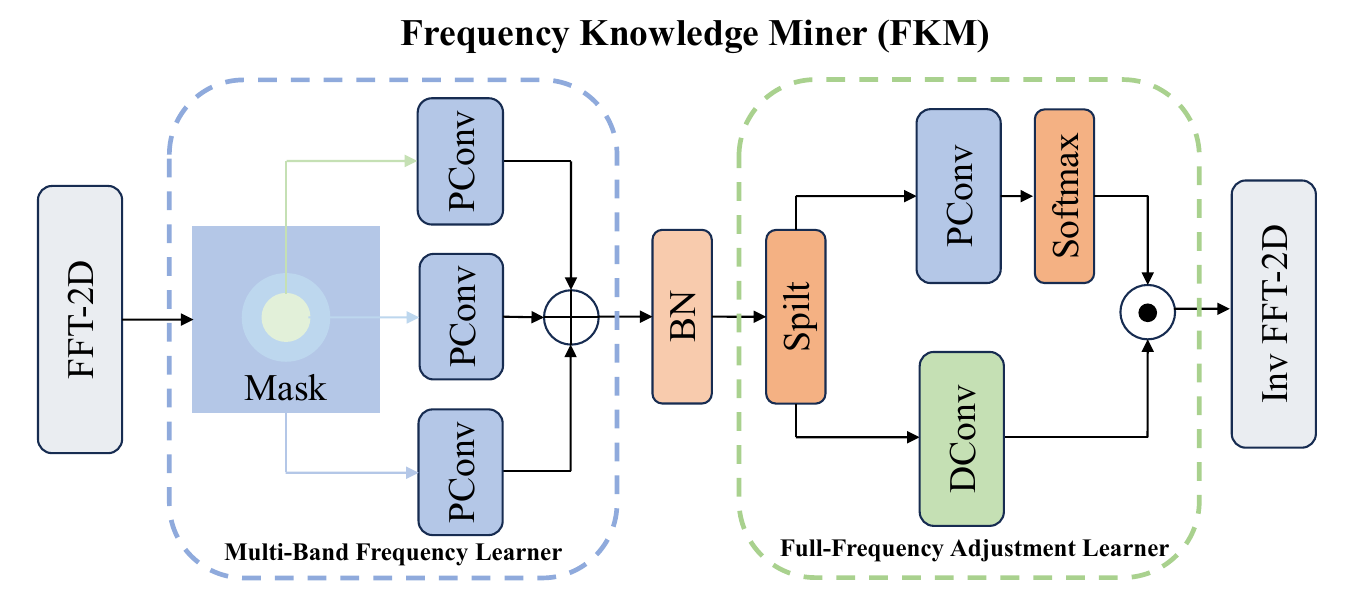}
\vspace{-0.7cm}
\end{center}
   \caption{\textit{Team NJUPCA:} The detailed architecture of the designed FKM.}
\label{fig:FKM}
\vspace{-0.4cm}
\end{figure}

\textbf{General Method Description.}
Inspired by SPAN \cite{wan2024swift}, they propose the Spatial Frequency Network (SFNet), which fully leverages both spatial and frequency domain representations. SFNet integrates Frequency Knowledge Miner (FKM) modules after each Spatial Attention Block (SPAB) to capture frequency domain features, complementing the spatial features extracted by SPAB. This parallel design enables the network to effectively learn and combine spatial and frequency domain representations, enhancing the performance of super-resolution reconstruction.

As illustrated in~\cref{fig:FKM}, the frequency knowledge miner (FKM) is designed to learn frequency representation from input, which comprises two core components:  multi-band frequency learner (MBFL) and  full-frequency adjustment learner (FFAL). MBFL aims to enhancing frequency representation by focusing on distinct frequency bands, while FFAL adjusts frequency-domain features from a full-frequency perspective.

\noindent
\textbf{Training Details.}
They employ two-stage training paradigm:

\begin{itemize}
    \item \textbf{Stage I - Foundation Training:} Randomly initialized weights are trained on DIV2K and full LSDIR datasets using 128$\times$128 HR patches. Configuration: Adam optimizer ($\beta_1=0.9$, $\beta_2=0.999$) with L1 loss, initial learning rate $5\times10^{-4}$ (halved every 200 epochs), batch size 64 over 1,000 epochs (34 hours on 4$\times$NVIDIA A6000).
    
    \item \textbf{Stage II - Refinement:} Initialized with Stage I weights, fine-tuned using DIV2K and LSDIR subset. Configuration: L2 loss with cosine learning schedule ($\eta_{\text{initial}}=1\times10^{-4}$), 500 epochs.
\end{itemize}

\noindent Other details: Training employed standard data augmentation (random rotation and flipping) without additional regularization techniques.

\subsection{DepthIBN}

\begin{figure}[!tb]
    \centering
    \includegraphics[width=1.0\linewidth]{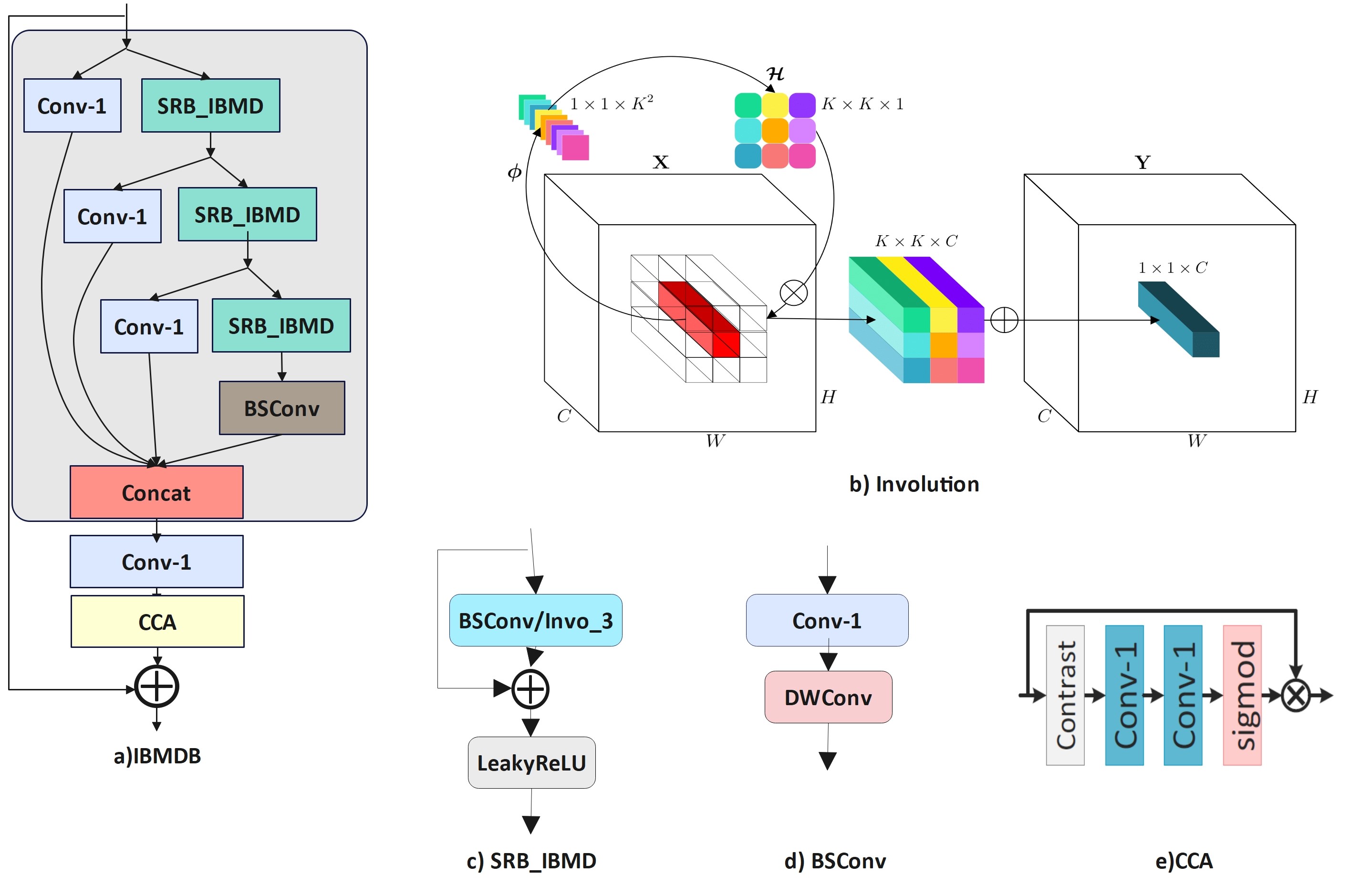}
    \caption{\textit{Team DepthIBN}: Involution and BSConv Multi-Depth Distillation Block (IBMDB).}
    \label{fig:team41_IBMDB}
\end{figure}

Single Image Super-Resolution (SISR) still faces challenges such as a large number of parameters, high memory consumption, and slow training and inference speed, despite significant advancements. These issues limit the practical use of SISR methods in real-world scenarios. Therefore, recent research has focused on developing lightweight models and optimizing network architectures. Among these techniques, Information Distillation is used to extract important features by splitting channels \cite{IDN,IMDN,RFDN,BSRN}. One of the main challenges of CNNs is the high computational cost of convolution operations. To reduce this cost, the Depthwise Separable Convolution (DSConv) \cite{Shufflenet,Mobilenets} method was introduced, but due to the separate processing of channels, some information may be lost. To address this issue, BSConv optimizes feature processing by utilizing kernel correlations, improving performance and reducing computations~\cite{BSConv}. Furthermore, shown in  \cref{fig:team41_IBMDB}, Involution replaces fixed filters with pixel-dependent dynamic filters, making it more sensitive to spatial variations and better at capturing long-range dependencies between pixels~\cite{invo}. Involution not only reduces parameters and resource consumption but also provides better performance compared to convolution-based models due to its superior feature extraction capability.

\noindent \textbf{Method.}
They used the IBMDN model in this challenge, following previous studies in the field of Lightweight Image Super-Resolution \cite{ibmdn}. They propose an Involution and BSConv Multi-Depth Distillation Network (IBMDN), consisting of 6 Involution and BSConv Multi-Depth Distillation Blocks (IBMDB). IBMDB integrates Involution and BSConv to balance computational efficiency and feature extraction. The overall architecture of their proposed model consists of four main sections: shallow feature extraction, deep feature extraction, feature fusion, and reconstruction. A 3×3 convolution is used to extract shallow features. Then, through 6 IBMDB blocks, deep features are extracted and fused using a 1×1 convolution, followed by refinement through a 3×3 convolution. The pixel-shuffle operation is then used as the reconstruction module.

The Involution and BSConv Multi-Depth Distillation Block (IBMDB) consists of three shallow residual blocks (SRB\_IBMD) and one channel contrast attention (CCA) block. Based on previous experiments, the use of 3×3 convolutions, due to computational complexity and a large number of parameters, is not always the best option, especially for lightweight super-resolution models \cite{deepf}. In SISR models, a fixed structure for feature extraction blocks is usually used, while features extracted at different depths of the network may differ. This approach may prevent the model from fully exploiting its capacity. Designing blocks with varying structures tailored to the depth of the network can enhance model performance. In their proposed model, the block structure is adjusted based on network depth to achieve an optimal feature extraction combination at different levels.

BSConv reduces parameters using intra-kernel correlation, better preserves information, and improves model accuracy without increasing complexity. Involution, with fewer learning parameters, extracts visual features through its attention mechanism and increases efficiency. Therefore, in the Information distillation structure, they consider the block structure differently. At the beginning of the network, BSConv is dominant in maintaining pixel correlation and local interactions within the block, and with increasing depth, Involution becomes the dominant operator. If BSConv is denoted by B and Involution by I, the optimal block combination in the deep feature extraction section is as follows: BBB-BBB-BIB-BIB-IBI-IBI. The details of the blocks are shown in the \cref{fig:team41_IBMDB}.

\subsection{Cidaut AI}

They propose a lightweight yet effective network with three blocks: an initial Sobel-based block and two ESA-based edge refinement blocks, regulated by a global residual connection. Upscaling is performed via pixel shuffle for efficient super-resolution.

\begin{figure}[ht]
    \centering
    \includegraphics[width=1.0\linewidth]{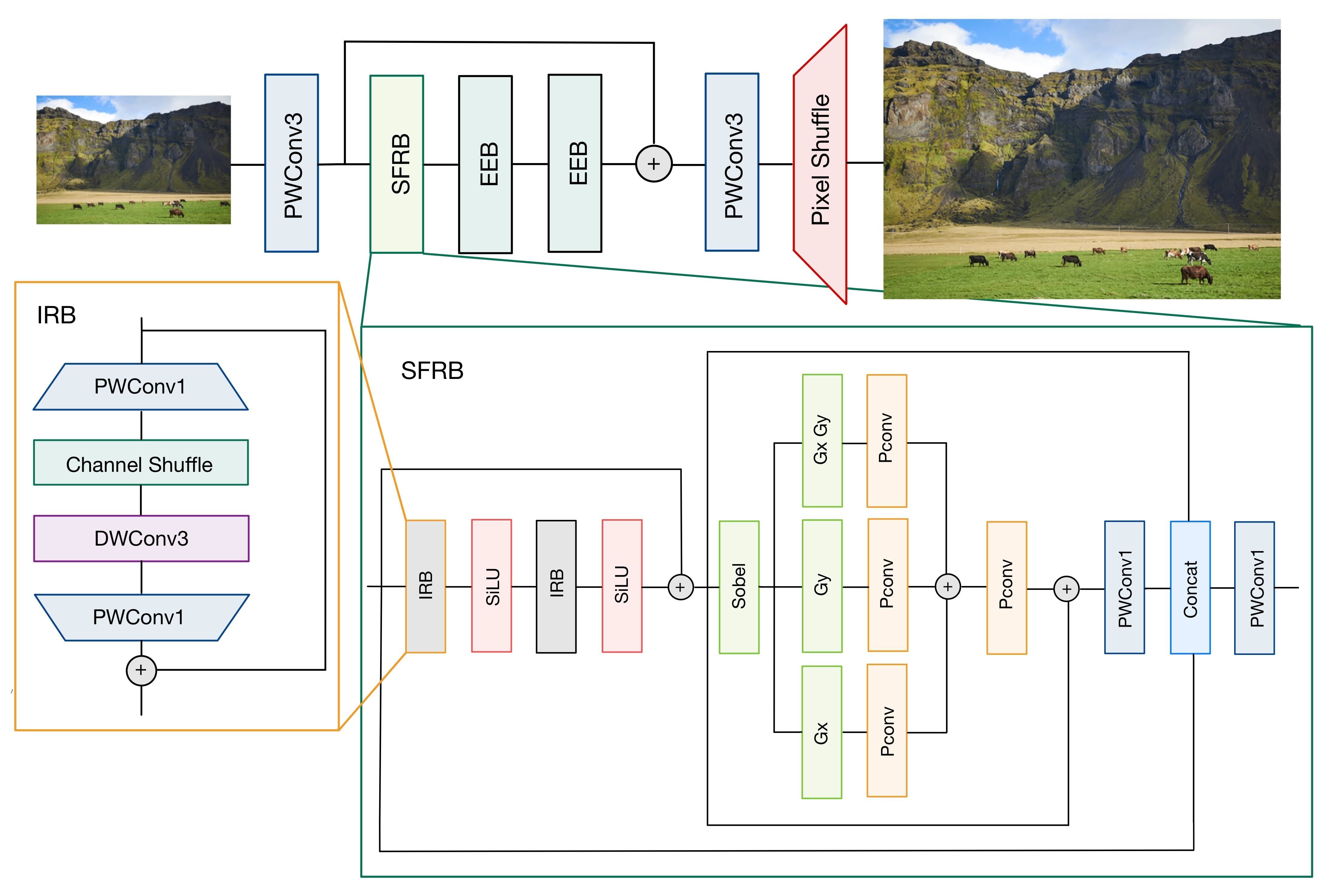}
    \caption{\textit{Team Cidaut AI}: Fused Edge Attention Network (FEAN) structure. They also show the Sobel Fused Residual Block (SFRB) and the Inverted Residual Bottlenecks (IRB) \cite{qin2024mobilenetv4universalmodels}.}
    \label{fig:team42_fig1}
\end{figure}

As shown in \cref{fig:team42_fig1}, the design integrates two MobileNet Inverted Bottlenecks \cite{qin2024mobilenetv4universalmodels} with channel shuffle and SiLU activation for enhanced information mixing. Inspired by EFDN \cite{wang2022edgeenhancedfeaturedistillationnetwork}, Sobel-based attention extracts edge features, refined using partial convolutions \cite{Park_2022} with minimal parameter increase. The final attention map, a weighted sum of refined \textit{Gx}, \textit{Gy}, and \textit{GxGy}, undergoes further refinement via partial convolution. A final 1×1 convolution preserves details while preventing excessive edge processing.

\begin{figure}[!tb]
    \centering
    \includegraphics[width=1.0\linewidth]{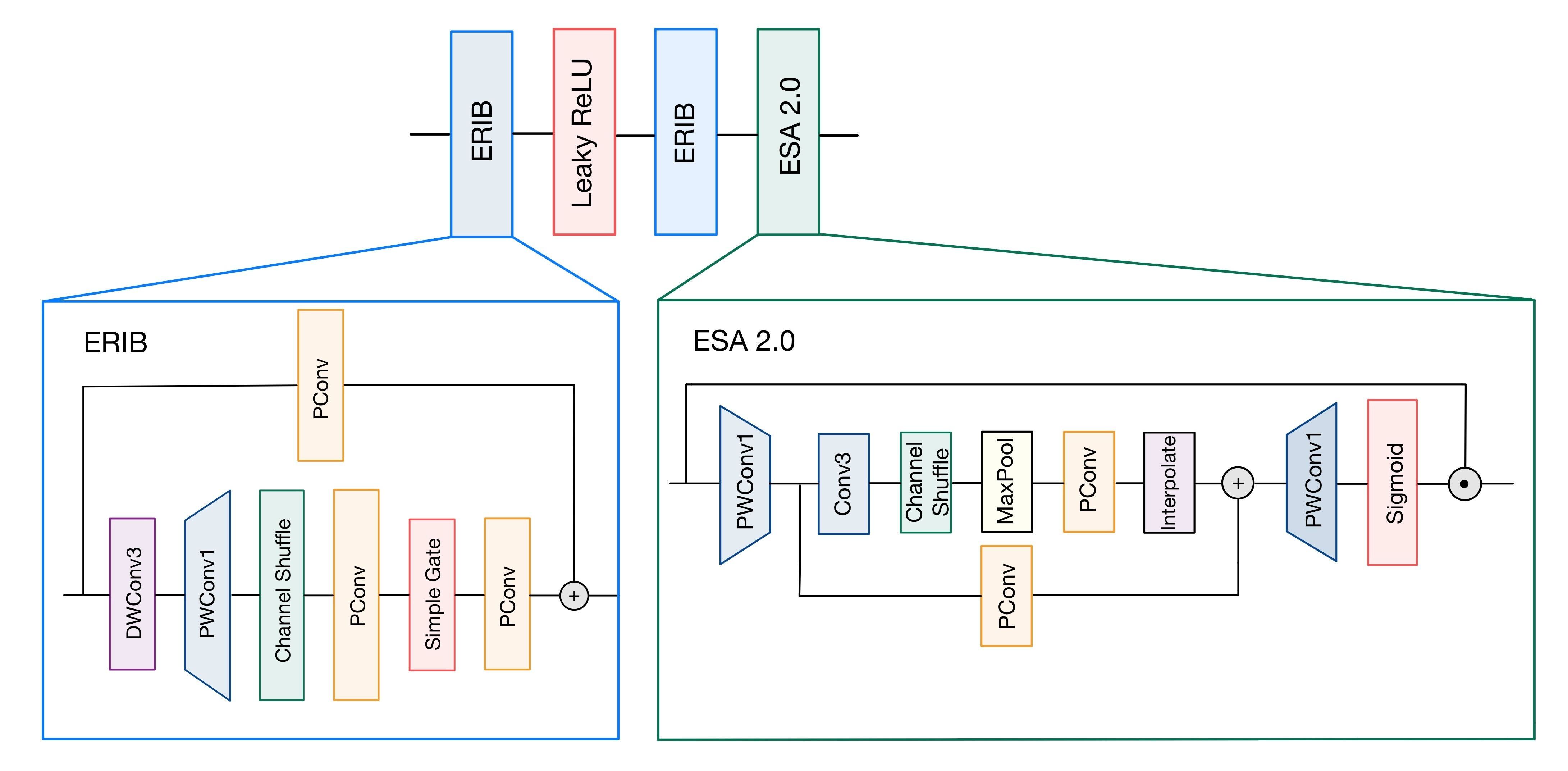}
    \caption{\textit{Team Cidaut AI}: 
 Structure of the Enhanced ESA Block (EEB).}
    \label{fig:team42_fig2}
\end{figure}

The proposed ERIB block, an efficient convolutional unit with self-activation, starts with depthwise convolution and 1×1 feature expansion~\cite{qin2024mobilenetv4universalmodels}. Partial convolutions~\cite{Park_2022} refine features, while channel shuffle enhances mixing. Inspired by Simple Gate \cite{chen2022simplebaselinesimagerestoration}, they introduce nonlinearity by reducing channels without increasing parameters. A weighted residual connection with partial convolution ensures effective information propagation, maintaining competitive performance despite PyTorch inefficiencies.

For the EEB in ~\cref{fig:team42_fig2}, they draw inspiration from the ReNRB block \cite{ren2024ninth}, replacing reparameterized convolutions with ERIB for improved efficiency. Partial convolutions in the ESA bottleneck and residual connections further exploit feature map redundancy.

\noindent
\textbf{Training Strategy}. The training was carried out using the DIV2K, FLICK2R, and LSIDR (30\%) datasets to improve the model's generalization ability. As a baseline, the model was trained for 1000 epochs with a cosine annealing learning rate scheduler, a crop size of 512 x 512, and a batch size of 16. Due to instability in the loss during training, an optimal learning rate analysis was performed whenever the loss diverged. This led to the implementation of a learning rate sweep strategy, which was organized into 5 stages.
\subsection{IVL}

\begin{figure*}[!t]
    \centering  
    \includegraphics[width=1\textwidth]{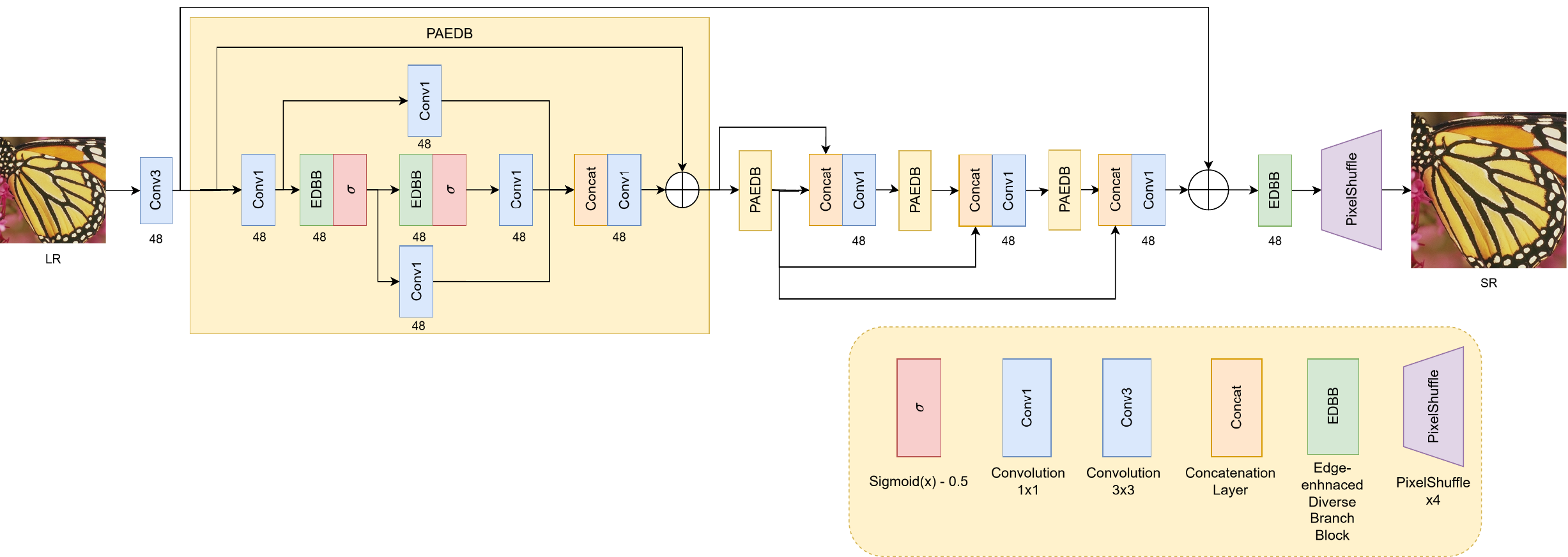}
    \caption{\textit{Team IVL}: Schematic diagram of the method.}
    \label{fig:IVL_method}
\end{figure*}

\noindent \textbf{Method.}
Their approach builds upon the strategy used in SPAN \cite{wan2023swift}, last year's winning method, to extract attention maps and integrates it into the proposed baseline architecture, EFDN \cite{EFDN}, aiming to enhance feature extraction and structural representation in image processing tasks.

%\begin{table*}
%\centering
%\resizebox{\textwidth}{!}{%
%\begin{tabular}{l|cccccc|ccc}
%\hline
%\multirow{3}{*}{Method} & \multicolumn{6}{c|}{PSNR/SSIM} & \multirow{3}{*}{\# Params (M)} & \multirow{3}{*}{Runtime (ms)} & \multirow{3}{*}{FLOPs (G)} \\ \cline{2-7}
% & \multicolumn{2}{c}{DIV2K\_LSDIR} & \multirow{2}{*}{Set5} & \multirow{2}{*}{Set14} & \multirow{2}{*}{Urban100} & \multirow{2}{*}{BSD100} &  &  &  \\ \cline{2-3}
% & Val & Test &  &  &  &  &  &  &  \\ \hline
%EFDN & \textbf{26.96/0.7965} & \textbf{27.01/-} & \textbf{30.18/0.8931} & \textbf{25.65/0.7507} & \textbf{24.48/0.7815} & \textbf{24.99/0.7099} & 0.276 & 29.71 & 16.70 \\ \hline
%Ours & 26.66/0.7890 & 26.76/0.80 & 29.74/0.8868 & 25.50/0.7472 & 24.07/0.7472 & 24.94/0.7061 & \textbf{0.240} & \textbf{28.42} & \textbf{15.64} \\ \hline
%\end{tabular}%
%}

%\caption{Results obtained by the method proposed by IVL team for datasets DIV2K\_LSDIR %\cite{Agustsson_2017_CVPR_Workshops}\cite{lilsdir}, Set5 \cite{bevilacqua2012low}, Set14 \cite{zeyde2010single}, BSD100 \cite{martin2001database}, Urban100 \cite{huang2015single}, compared with the baseline method from the challenge, EFDN. Best results are reported in \textbf{bold}.}
%\label{tab:IVL-results}
%\end{table*}
Specifically, as illustrated in Figure \ref{fig:IVL_method}, this strategy is incorporated within the EDBB blocks of EFDN, which are designed to capture fundamental structural features of an image by applying Sobel and Laplacian filters. These filters emphasize edge and texture information, contributing to improved representation learning. During the inference phase, the EDBB blocks are reparametrized into 3x3 convolutions to maintain computational efficiency while preserving learned feature representations.

The attention maps are derived following the approach implemented in SPAN, leveraging an activation function that is both odd and symmetric to effectively highlight essential regions of the image. These attention maps serve as a direct substitute for the ESA block present in the original EFDN model, aiming to refine feature selection and enhance the model’s overall performance.

As a result of the applied modifications, the final architecture has a lower parameter count and requires fewer floating-point operations compared to the proposed baseline method, EFDN.

\noindent \textbf{Training Details.}
The training process is structured into three progressive phases to optimize performance and stability:
\begin{itemize}
    \item Pre-training: The model undergoes an initial training phase using the DIV2K dataset, incorporating data augmentation techniques such as random rotations, horizontal flipping, and random cropping to generate patches of size $64\times64$. Training is conducted over 30,000 iterations with a batch size of 32, utilizing the Adam optimizer ($\beta_1 = 0.9$, $\beta_2 = 0.999$). The learning rate is initially set to 1e-3 for the first 20,000 iterations and subsequently reduced to 1e-4 for the remaining 10,000 iterations. L1 loss is used throughout this phase.
    \item First training stage: The model is further refined using the DIV2K\_LSDIR dataset, while maintaining the same augmentation strategies as in the pre-training phase. The patch size is increased to $256\times256$, and training is extended to 100,000 iterations with a batch size of 64. The Adam optimizer ($\beta_1 = 0.9$, $\beta_2 = 0.999$) is employed, starting with a learning rate of 5e-4, which undergoes a decay by a factor of 0.5 every 20,000 iterations. L1 loss remains the chosen loss function for this stage.
    \item Second training stage: In the final phase, training continues on the DIV2K\_LSDIR dataset with an expanded patch size of $512\times512$ for an additional 40,000 iterations. The same augmentation methods are retained, and most hyperparameters remain unchanged. However, to ensure stable convergence and fine-tune performance, the learning rate is reduced to 5e-5. During this stage, L1 loss is applied for the first 10,000 iterations, after which L2 loss is utilized to enhance final model performance.

All the training phases were performed of the model a single NVIDIA RTX 4070 Super GPU and required approximately 20 hours.

\end{itemize}

%\paragraph{Experimental results}
%We evaluate our model in terms of Peak Signal-to-Noise Ratio (PSNR) and Structural Similarity Index (SSIM) and compare it against the baseline EFDN model. The obtained results, as indicated in Table \ref{tab:IVL-results}, show that while our proposed approach does not surpass the baseline in terms of these performance metrics, it achieves comparable results while reducing computational complexity.

%Specifically, our modified architecture requires fewer floating-point operations (GFLOPs), has a reduced parameter count, and demonstrates lower runtime (both runtime results were computed locally on our machine).

%Considering the relatively short training duration and the observed efficiency improvements, our approach presents a viable and competitive alternative to the baseline model.

%%%%%%%%% Acknowledgments %%%%%%%%%
% \clearpage % This can be deleted later
\section*{Acknowledgments}
This work was partially supported by the Humboldt Foundation, the Ministry of Education and Science of Bulgaria (support for INSAIT, part of the Bulgarian National Roadmap for Research Infrastructure). We thank the NTIRE 2025 sponsors: ByteDance, Meituan, Kuaishou, and University of Wurzburg (Computer Vision Lab).

%%%%%%%%% Appendix %%%%%%%%%
%-----------------------%
\appendix
%-----------------------%

\section{Teams and Affiliations}
\label{sec:teams}
\subsection*{NTIRE 2025 ESR Teams}
\noindent\textit{\textbf{Title: }} NTIRE 2025 Efficient Super-Resolution Challenge\\
\noindent\textit{\textbf{Members: }} \\
Bin Ren$^{1,2,4}$ (\href{mailto:bin.ren@unitn.it}{bin.ren@unitn.it}),\\
Hang Guo$^3$ (\href{cshguo@gmail.com}{cshguo@gmail.com}),\\
Lei Sun$^4$ (\href{lei.sun@insait.ai}{lei.sun@insait.ai}) \\
Zongwei Wu$^5$ (\href{zongwei.wu@uni-wuerzburg.de}{zongwei.wu@uni-wuerzburg.de}),\\
Radu Timofte$^{5}$ (\href{mailto:radu.timofte@vision.ee.ethz.ch}{radu.timofte@vision.ee.ethz.ch})\\
Yawei Li$^6$ (\href{mailto:li.yawei.ai@gmail.com}{li.yawei.ai@gmail.com}),\\
\noindent\textit{\textbf{Affiliations: }}\\
$^1$ University of Pisa, Italy\\
$^2$ University of Trento, Italy\\
$^3$ Tsinghua University, China\\
$^4$ INSAIT,Sofia University,"St.Kliment Ohridski”, Bulgaria\\
$^5$ Computer Vision Lab, University of W\"urzburg, Germany\\
$^6$ ETH Z\"urich, Switzerland\\

% ---- 1st - 3rd of the Main Track 
\subsection*{EMSR}
\noindent\textit{\textbf{Title: }} Distillation-Supervised Convolutional Low-Rank Adaptation for Efficient Image Super-Resolution \\
\noindent\textit{\textbf{Members: }} \\
Yao Zhang $^1$ (\href{mailto:yao_zhang@sjtu.edu.cn}{yao\_zhang@sjtu.edu.cn}),\\
Xinning Chai $^1$ (\href{mailto:chaixinning@sjtu.edu.cn}{chaixinning@sjtu.edu.cn}),\\
Zhengxue Cheng $^1$ (\href{mailto:zxcheng@sjtu.edu.cn}{zxcheng@sjtu.edu.cn}),\\
Yingsheng Qin $^2$ (\href{mailto:yingsheng.qin@transsion.com}{yingsheng.qin@transsion.com}),\\
Yucai Yang $^2$ (\href{mailto:yucai.yang@transsion.com}{yucai.yang@transsion.com}),\\
Li Song $^1$ (\href{mailto:song_li@sjtu.edu.cn}{song\_li@sjtu.edu.cn}),\\
\noindent\textit{\textbf{Affiliations: }} \\ 
$^1$ Shanghai Jiao Tong University \\
$^2$ Transsion in China \\

\subsection*{XiaomiMM}
\noindent\textit{\textbf{Title: }} SPANF \\
\noindent\textit{\textbf{Members: }} \\
Hongyuan Yu$^1$ (\href{mailto:yuhyuan1995@gmail.com}{yuhyuan1995@gmail.com}),\\
Pufan Xu$^2$ (\href{mailto:xpf22@mails.tsinghua.edu.cn}{xpf22@mails.tsinghua.edu.cn}), \\
Cheng Wan$^3$ (\href{mailto:jouiney666@gmail.com}{jouiney666@gmail.com}), \\
Zhijuan Huang$^1$ (\href{mailto:huangzhijuan@xiaomi.com}{huangzhijuan@xiaomi.com}), \\
Peng Guo$^4$ (\href{mailto:guopeng0100@163.com}{guopeng0100@163.com}), \\
Shuyuan Cui$^5$ (\href{mailto:jouiney666@gmail.com}{jouiney666@gmail.com}), \\
Chenjun Li$^3$ (\href{mailto:cl2733@cornell.edu}{cl2733@cornell.edu}), \\
Xuehai Hu (\href{mailto:hsquare@mail.ustc.edu.cn}{hsquare@mail.ustc.edu.cn}), \\
Pan Pan$^1$ (\href{mailto:panpan@xiaomi.com}{panpan@xiaomi.com}), \\
Xin Zhang$^1$ (\href{mailto:zhangxin14@xiaomi.com}{zhangxin14@xiaomi.com}), \\
Heng Zhang$^1$ (\href{mailto:zhangheng8@xiaomi.com}{zhangheng8@xiaomi.com}), \\

\noindent\textit{\textbf{Affiliations: }} \\ 
$^1$ Multimedia Department, Xiaomi Inc. \\
$^2$ School of Integrated Circuits, Tsinghua University \\
$^3$ Cornell University \\
$^4$ Hanhai Information Technology (Shanghai) Co., Ltd. \\
$^5$ Huatai Insurance Group Co., Ltd.\\

\subsection*{ShannonLab}
\noindent\textit{\textbf{Title: }}Reparameterization Network for Efficient Image Super-Resolution\\
\noindent\textit{\textbf{Members: }} \\
Qing Luo$^1$ (\href{mailto:yourEmail@xxx.xxx}{luoqing.94@qq.com}),\\
Linyan Jiang$^1$, \\
Haibo Lei$^1$, \\
Qifang Gao$^1$, \\
Yaqing Li$^1$, \\

\noindent\textit{\textbf{Affiliations: }} \\ 
$^1$ Tencent \\

% ---- 1st - 3rd of the Sub-Track: Runtime
% 1st Sub-Track: Runtime
% \input{teams/team31_ShannonLab/affiliation}
% 2nd Sub-Track: Runtime
\subsection*{TSSR}
\noindent\textit{\textbf{Title: }} Light Network for Efficient Image Super-Resolution\\
\noindent\textit{\textbf{Members: }} \\
Weihua Luo$^1$ (\href{mailto:yourEmail@xxx.xxx}{185471613@qq.com}),\\
Tsing Li$^1$, \\

\noindent\textit{\textbf{Affiliations: }} \\ 
$^1$ Independent researcher \\

        % Incomplete
% 3rd Sub-Track: Runtime
\subsection*{mbga}
\noindent\textit{\textbf{Title: }} Espanded SPAN for Efficient Super-Resolution\\
\noindent\textit{\textbf{Members: }} \\
Qing Wang$^1$ (\href{mailto:yourEmail@xxx.xxx}{wangqing.keen@bytedance.com}),\\
Yi Liu$^1$, \\
Yang Wang$^1$, \\
Hongyu An$^1$, \\
Liou Zhang$^1$, \\
Shijie Zhao$^1$, \\

\noindent\textit{\textbf{Affiliations: }} \\ 
$^1$ ByteDance \\

% ---- 1st - 3rd of the Sub-Track: FLOPs
% 1st Sub-Track: FLOPs
\subsection*{VPEG\_C}
\noindent\textit{\textbf{Title: }} DAN: Dual Attention Network for lightweight Image Super-Resolution\\
\noindent\textit{\textbf{Members: }} \\
Lianhong Song$^1$ (\href{mailto:yourEmail@xxx.xxx}{songlianhong@njust.edu.cn}),\\
Long Sun$^1$, \\
Jinshan Pan$^1$, \\
Jiangxin Dong$^1$, \\
Jinhui Tang$^1$ \\

\noindent\textit{\textbf{Affiliations: }} \\ 
$^1$ Nanjing University of Science and Technology \\

% 2nd Sub-Track: FLOPs
\subsection*{XUPTBoys}
\noindent\textit{\textbf{Title: }} Frequency-Guided Multi-level Dispersion Network for Efficient Image Super-Resolution\\
\noindent\textit{\textbf{Members: }} \\
Jing Wei$^1$ (\href{mailto:yourEmail@xxx.xxx}
{freedomwj@126.com}),\\
Mengyang Wang$^1$, \\
Ruilong Guo$^1$, \\
Qian Wang$^{1,2}$, \\
\noindent\textit{\textbf{Affiliations: }} \\ 
$^1$ Xi'an University of Posts and Telecommunications \\
$^2$ National Engineering Laboratory for Cyber Event Warning and Control Technologies \\
% 3rd Sub-Track: FLOPs
\subsection*{HannahSR}
\noindent\textit{\textbf{Title: }} Multi-level Refinement and Bias-learnable Attention Dual Branch Network for Efficient Image Super-Resolution\\
\noindent\textit{\textbf{Members: }} \\
Qingliang Liu$^1$ (\href{mailto:yourEmail@xxx.xxx}{liuqingliang1@honor.com}),\\
Yang Cheng$^2$ (\href{mailto:yourEmail@xxx.xxx}{obliviate73@outlook.com})\\
\noindent\textit{\textbf{Affiliations: }} \\ 
$^1$ Beijing Honor Device Co.,Ltd. \\
$^2$ State Key Laboratory of Integrated Chip \& System, Fudan University \\

% ---- 1st - 3rd of the Sub-Track: Params
% 1st Sub-Track: Params
% \input{teams/team20_VPEG_C/affiliation}
% 2nd Sub-Track: Params
% \input{teams/team13_HannahSR/affiliation}
% 3rd Sub-Track: Params
% \input{teams/team26_XUPTBoys/affiliation}

% --- Other Ranked Teams via Main Track order:
% 4th of Main Track
% \input{teams/team58_TSSR/affiliation}        % Incomplete
% 5th of Main Track
\subsection*{Davinci}
\noindent\textit{\textbf{Title: }} PlayerAug\\
\noindent\textit{\textbf{Members: }} \\
Davinci (\href{mailto:yourEmail@xxx.xxx}{1016994139@qq.com}),\\
Enxuan Gu$^1$(\href{mailto:yourEmail@xxx.xxx}{guexstan@163.com}), \\
\noindent\textit{\textbf{Affiliations: }} \\ 
$^1$ Dalian University of Technology \\

% 6th of Main Track
\subsection*{SRCB}
\noindent\textit{\textbf{Title: }} SPAN with pruning.\\
\noindent\textit{\textbf{Members: }} \\
Dafeng Zhang$^1$ (\href{mailto:dfeng.zhang@samsung.com}{dfeng.zhang@samsung.com}),\\
Yang Yong$^1$, \\
\noindent\textit{\textbf{Affiliations: }} \\ 
$^1$ Samsung Research China - Beijing (SRC-B) \\

        % Incomplete
% 7th of Main Track
\subsection*{Rochester}
\noindent\textit{\textbf{Title: }}ESRNet: An enhanced version of SPAN for Efficient Super-Resolution\\
\noindent\textit{\textbf{Members: }} \\
Pinxin Liu$^1$ (\href{mailto:pliu23@ur.rochester.edu}{pliu23@ur.rochester.edu}),\\
Yongsheng Yu$^1$ (\href{mailto:yyu90@ur.rochester.edu}{yyu90@ur.rochester.edu}),\\
Hang Hua$^1$ (\href{mailto:hhua2@cs.rochester.edu}{hhua2@cs.rochester.edu}),\\
Yunlong Tang$^1$ (\href{mailto:yunlong.tang@rochester.edu}{yunlong.tang@rochester.edu}),\\
\noindent\textit{\textbf{Affiliations: }} \\ 
$^1$ University of Rochester \\

% 8th of Main Track
% \input{teams/team36_mbga/affiliation}
% 9th of Main Track
\subsection*{IESR}
\noindent\textit{\textbf{Title: }} Inference Efficient Super-Rosolution Net\\
\noindent\textit{\textbf{Members: }} \\
Shihao Wang$^1$ (\href{mailto:yourEmail@xxx.xxx}{shihao.wsh@antgroup.com}),\\
Yukun Yang$^1$, \\
Zhiyu Zhang$^1$, \\
\noindent\textit{\textbf{Affiliations: }} \\ 
$^1$ Ant Group \\

% 10th of Main Track
\subsection*{ASR}
\noindent\textit{\textbf{Title: }} ASR\\
\noindent\textit{\textbf{Members: }} \\
Yukun Yang$^1$ (\href{mailto:yourEmail@xxx.xxx}{yukun.yyk@antgroup.com}),\\
\noindent\textit{\textbf{Affiliations: }} \\ 
$^1$ None \\
% 11th of Main Track
\subsection*{VPEG\_O}
\noindent\textit{\textbf{Title: }} SAFMNv3: Simple Feature Modulation Network for Real-Time Image
Super-Resolution\\
\noindent\textit{\textbf{Members: }} \\
Long Sun$^1$ (\href{mailto:yourEmail@xxx.xxx}{cs.longsun@njust.edu.cn}),\\
Lianhong Son $^1$, \\
Jinshan Pan$^1$, \\
Jiangxin Dong$^1$, \\
Jinhui Tang$^1$ \\
\noindent\textit{\textbf{Affiliations: }} \\ 
$^1$ Nanjing University of Science and Technology \\

% 12th of Main Track
\subsection*{mmSR}
\noindent\textit{\textbf{Title: }} Efficient Feature Aggregation Network for Image Super-Resolution\\
\noindent\textit{\textbf{Members: }} \\
Jiyu Wu$^1$ (\href{mailto:yourEmail@xxx.xxx}{jiyu\_wu@163.com}),\\
Jiancheng Huang $^1$(\href{mailto:yourEmail@xxx.xxx}{jc.huang@siat.ac.cn}), \\
Yifan Liu $^1$, \\
Yi Huang $^1$, \\
Shifeng Chen $^1$, \\
\noindent\textit{\textbf{Affiliations: }} \\ 
$^1$ Shenzhen Institutes of Advanced Technology, Chinese Academy of Sciences \\

% 13th of Main Track
\subsection*{ChanSR}
\noindent\textit{\textbf{Title: }} EECNet: Edge Enhanced Convolutional Network for Efficient Super-Resolution\\
\noindent\textit{\textbf{Members: }} \\
Rui Chen$^1$ (\href{mailto:yourEmail@xxx.xxx}{chenr269@163.com}),\\
\noindent\textit{\textbf{Affiliations: }} \\ 
$^1$ Shenzhen International Graduate School, Tsinghua University, China \\

% 14th of Main Track
\subsection*{Pixel\_Alchemists}
\noindent\textit{\textbf{Title: }} RCUNet\\
\noindent\textit{\textbf{Members: }} \\
Yi Feng$^1$ (\href{mailto:fenyi_work@163.com}{fenyi\_work@163.com}),\\
Mingxi Li$^1$, \\
Cailu Wan$^1$, \\
Xiangji Wu$^1$, \\

\noindent\textit{\textbf{Affiliations: }} \\ 
$^1$ Independent researcher \\

% 15th of Main Track
% \input{teams/team39_MiSR/affiliation}      % NO Update
% 16th of Main Track
\subsection*{LZ}
\noindent\textit{\textbf{Title:}} Tensor decompose efficient super-resolution network\\
\noindent\textit{\textbf{Members: }} \\
Zibin Liu$^1$ (\href{mailto:yourEmail@xxx.xxx}{1451971605@qq.com}),\\
Jinyang Zhong$^2$ (\href{mailto:yourEmail@xxx.xxx}{1439764064@qq.com}),\\

\noindent\textit{\textbf{Affiliations: }} \\ 
$^1$ Southwest Jiaotong University \\
$^2$ Sichuan University\\

% 17th of Main Track
\subsection*{Z6}
\noindent\textit{\textbf{Title: }} GLoReNet: Global and Local feature Refinement Network for Efficient Super-Resolution \\
\noindent\textit{\textbf{Members: }} \\
Kihwan Yoon$^1$ (\href{mailto:rlghksdbs@gmail.com}{rlghksdbs@gmail.com}),\\
Ganzorig Gankhuyag$^1$, \\
\noindent\textit{\textbf{Affiliations: }} \\ 
$^1$ Korea Electronics Technology Institute (KETI)  \\

% 18th of Main Track
\subsection*{TACO\_SR}
\noindent\textit{\textbf{Title: }} TenInOneSR\\
\noindent\textit{\textbf{Members: }} \\
Shengyun Zhong$^1$ (\href{mailto:yourEmail@xxx.xxx}{shengyunzhong2002@gmail.com}),\\
Mingyang Wu$^2$ (\href{mailto:yourEmail@xxx.xxx}{mingyang@tamu.edu}),\\
Renjie Li$^2$ (\href{mailto:yourEmail@xxx.xxx}{renjie@tamu.edu}),\\
Yushen Zuo$^3$ (\href{mailto:yourEmail@xxx.xxx}{zuoyushen12@gmail.com}),\\
Zhengzhong Tu$^2$ (\href{mailto:yourEmail@xxx.xxx}{tzz@tamu.edu}),\\
\noindent\textit{\textbf{Affiliations: }} \\ 
$^1$ Northeastern University, USA \\
$^2$ Texas A\&M University, USA \\
$^3$ The Hong Kong Polytechnic University, Hong Kong \\

% 19th of Main Track
\subsection*{AIOT\_AI}
\noindent\textit{\textbf{Title: }} Efficient channel attention super-resolution network acting on space\\
\noindent\textit{\textbf{Members: }} \\
Zongang Gao 1$^1$ (\href{mailto:yourEmail@xxx.xxx}{gaozongang@qq.com}),\\
Guannan Chen$^1$, \\
Yuan Tian$^1$, \\
Wenhui Chen$^1$\\
\noindent\textit{\textbf{Affiliations: }} \\ 
$^1$ BOE, AIOT CTO, Beijing, China \\

% 20th of Main Track
\subsection*{JNU620}
\noindent\textit{\textbf{Title: }} Reparameterized Residual Local Feature Network for Efficient
Image Super-Resolution\\
\noindent\textit{\textbf{Members: }} \\
Weijun Yuan$^1$ (\href{mailto:yweijun@stu2022.jnu.edu.cn}{yweijun@stu2022.jnu.edu.cn}),\\
Zhan Li$^1$, \\
Yihang Chen$^1$, \\
Yifan Deng$^1$, \\
Ruting Deng$^1$, \\
\noindent\textit{\textbf{Affiliations: }} \\ 
$^1$ Jinan University \\
% $^2$ Affiliation 2 \\
% $^3$ Affiliation 3 \\

% 21st of Main Track
\subsection*{LVGroup\_HFUT}
\noindent\textit{\textbf{Title: }} Swift Parameter-free Attention Network for Efficient Image Super-Resolution\\
\noindent\textit{\textbf{Members: }} \\
Yilin Zhang$^1$ (\href{mailto:eslzzyl@163.com}{eslzzyl@163.com}),\\
Huan Zheng$^2$, (\href{mailto:huanzheng1998@gmail.com}{huanzheng1998@gmail.com}),\\
Yanyan Wei$^1$ (\href{mailto:weiyy@hfut.edu.cn}{weiyy@hfut.edu.cn}),\\
Wenxuan Zhao$^1$ (\href{mailto:nightvoyagerr@gmail.com}{nightvoyagerr@gmail.com}),\\
Suiyi Zhao$^1$ (\href{mailto:meranderzhao@gmail.com}{meranderzhao@gmail.com}),\\
Fei Wang$^1$ (\href{mailto:jiafei127@gmail.com}{jiafei127@gmail.com}),\\
Kun Li$^1$ (\href{mailto:kunli.hfut@gmail.com}{kunli.hfut@gmail.com}),\\
\noindent\textit{\textbf{Affiliations: }} \\ 
$^1$ Hefei University of Technology \\
$^2$ University of Macau \\

% 22nd of Main Track
% \input{teams/team43_SVM/affiliation}       % NO Update
% 23rd of Main Track
\subsection*{YG}
\noindent\textit{\textbf{Title: }} Spatial-Gate Self-Distillation Network for Efficient Image Super-Resolution\\
\noindent\textit{\textbf{Members: }} \\
Yinggan Tang $^1$ (\href{mailto:yourEmail@xxx.xxx}{ygtang@ysu.edu.cn}),\\
Mengjie Su $^2$, \\
\noindent\textit{\textbf{Affiliations: }} \\ 
$^1$ School of Electrical Engineering, Yanshan University \\

% 24th of Main Track
% \input{teams/team07_NanoSR/affiliation}
% 25th of Main Track
\subsection*{MegastudyEdu\_Vision\_AI}
\noindent\textit{\textbf{Title: }} Multi-scale Aggrgation Attention Network for Efficient Image Super-resolution\\
\noindent\textit{\textbf{Members: }} \\
Jae-hyeon Lee $^1$ (\href{mailto:yourEmail@xxx.xxx}{dlwogus147@gmail.com}),\\
Dong-Hyeop Son $^1$, \\
Ui-Jin Choi $^1$, \\
\noindent\textit{\textbf{Affiliations: }} \\ 
$^1$ MegastudyEdu Vision AI \\

% 26th of Main Track
% \input{teams/team26_XUPTBoys/affiliation}
% 27th of Main Track
\subsection*{MILA}
\noindent\textit{\textbf{Title: }} Multi-Level Variance Feature Modulation Network for Lightweight Image Super-Resolution\\
\noindent\textit{\textbf{Members: }} \\
Tiancheng Shao$^1$ (\href{mailto:yourEmail@xxx.xxx}{shaotiancheng666@outlook.com}),\\
Yuqing Zhang$^2$, \\
Mengcheng Ma$^3$, \\
\noindent\textit{\textbf{Affiliations: }} \\ 
$^1$ Anhui University of Technology \\

% 28th of Main Track
\subsection*{AiMF\_SR}
\noindent\textit{\textbf{Title: }} Mixture of Efficient Attention for Efficient Image Super-Resolution\\
\noindent\textit{\textbf{Members: }} \\
Donggeun Ko$^1$ (\href{mailto:sean.ko@aimfuture.ai}{sean.ko@aimfuture.ai}),\\
Youngsang Kwak$^1$, \\
Jiun Lee$^1$, \\
Jaehwa Kwak$^1$, \\
\noindent\textit{\textbf{Affiliations: }} \\ 
$^1$ AiM Future Inc. \\

% 29th of Main Track
% \input{teams/team33_EagleSR/affiliation}   % NO Update
% 30th of Main Track
\subsection*{BVIVSR}
\noindent\textit{\textbf{Title: }} NTIRE 2025 Efficient SR Challenge Factsheet\\
\noindent\textit{\textbf{Members: }} \\
Yuxuan Jiang$^1$ (\href{yuxuan.jiang@bristol.ac.uk}{yuxuan.jiang@bristol.ac.uk}),\\
Qiang Zhu$^{2,1}$ (\href{zhuqiang@std.uestc.edu.cn}{zhuqiang@std.uestc.edu.cn}),\\
Siyue Teng$^1$ (\href{siyue.teng@bristol.ac.uk}{siyue.teng@bristol.ac.uk}),\\
Fan Zhang$^1$, (\href{fan.zhang@bristol.ac.uk}{fan.zhang@bristol.ac.uk}),\\
Shuyuan Zhu$^2$, (\href{eezsy@uestc.edu.cn}{eezsy@uestc.edu.cn}),\\
Bing Zeng$^2$, (\href{eezeng@uestc.edu.cn}{eezeng@uestc.edu.cn}),\\
David Bull$^1$ (\href{dave.bull@bristol.ac.uk}{dave.bull@bristol.ac.uk}), \\
\noindent\textit{\textbf{Affiliations: }} \\ 
$^1$ University of Bristol \\
$^2$ University of Electronic Science and Technology of China \\
% $^3$ Affiliation 3 \\

% 31st of Main Track
% --- 3rd Sub-Track: FLOPs & 2nd Sub-Track: Params
% \input{teams/team13_HannahSR/affiliation}
% 32nd of Main Track
% --- 1st Sub-Track: FLOPs & 1st Sub-Track: Params
% \input{teams/team20_VPEG_C/affiliation}
% 33rd of Main Track
\subsection*{CUIT\_HTT}
\noindent\textit{\textbf{Title: }} Frequency-Segmented Attention Network for Lightweight Image Super\\
\noindent\textit{\textbf{Members: }} \\
Jing Hu$^1$ (\href{mailto:jing_hu09@163.com}{jing\_hu@163.com}),\\
Hui Deng$^1$, \\
Xuan Zhang$^1$, \\
Lin Zhu$^1$\\
Qinrui Fan$^1$\\
\noindent\textit{\textbf{Affiliations: }} \\ 
$^1$ Chengdu University of Information Technology \\

% 34th of Main Track
\subsection*{GXZY\_AI}
\noindent\textit{\textbf{Title: }} Parameter Free Vision Mamba For Lightweight Image Super-Resolution\\
\noindent\textit{\textbf{Members: }} \\
Weijian Deng$^1$ (\href{mailto:348957269@qq.com}{348957269@qq.com}),\\
Junnan Wu$^1$ (\href{mailto:838050895@qq.com}{838050895@qq.com}), \\
Wenqin Deng$^2$ (\href{mailto:1601524278@qq.com}{1601524278@qq.com}),\\
Yuquan Liu$^1$ (\href{mailto:653060432@qq.com}{653060432@qq.com}),\\
Zhaohong Xu$^1$ (\href{mailto:719357155@qq.com}{719357155@qq.com}),\\
\noindent\textit{\textbf{Affiliations: }} \\ 
$^1$ Guangxi China Tobacco Industry Corporation Limited, China \\
$^2$ Guangxi University, China \\

% 35th of Main Track
% \input{teams/team16_SCMSR/affiliation}      % NO Update
% 36th of Main Track
\subsection*{IPCV}
\noindent\textit{\textbf{Title: }} Efficient HiTSR\\
\noindent\textit{\textbf{Members: }} \\
Jameer Babu Pinjari $^1$ (\href{mailto:jameer.jb@gmail.com}{jameer.jb@gmail.com}),\\
Kuldeep Purohit $^1$, (\href{mailto:kuldeeppurohit3@gmail.com}{kuldeeppurohit3@gmail.com})\\
\noindent\textit{\textbf{Affiliations: }} \\ 
$^1$ Independent researcher \\

% 37th of Main Track
\subsection*{X-L}
\noindent\textit{\textbf{Title: }} Partial Permuted Self-Attention for Lightweight Super-Resolution\\
\noindent\textit{\textbf{Members: }} \\
Zeyu Xiao$^1$ (\href{mailto:zeyuxiao1997@163.com}{zeyuxiao1997@163.com}),\\
Zhuoyuan Li$^2$ (\href{mailto:zhuoyuanli@mail.ustc.edu.cn}{zhuoyuanli@mail.ustc.edu.cn})\\
\noindent\textit{\textbf{Affiliations: }} \\ 
$^1$ National University of Singapore \\
$^2$ University of Science and Technology of China \\

% 38th of Main Track
\subsection*{Quantum\_Res}
\noindent\textit{\textbf{Title: }}Efficient Mamba-Based Image Super-Resolution via Knowledge Distillation\\
\noindent\textit{\textbf{Members: }} \\
Surya Vashisth$^1$ (\href{mailto:surya.vashisth@s.amity.edu}{surya.vashisth@s.amity.edu}),\\
Akshay Dudhane$^2$
(\href{mailto:akshay.dudhane@mbzuai.ac.ae}{akshay.dudhane@mbzuai.ac.ae}), \\
Praful Hambarde$^3$
(\href{mailto:praful@iitmandi.ac.in}{praful@iitmandi.ac.in}), \\
Sachin Chaudhary$^4$
(\href{mailto:sachin.chaudhary@ddn.upes.ac.in}{sachin.chaudhary@ddn.upes.ac.in}), \\
Satya Naryan Tazi$^5$
(\href{mailto:satya.tazi@ecajmer.ac.in}{satya.tazi@ecajmer.ac.in}), \\
Prashant Patil$^6$
(\href{mailto:pwpatil@iitg.ac.in}{pwpatil@iitg.ac.in}), \\
Santosh Kumar Vipparthi$^7$
(\href{mailto:skvipparthi@iitrpr.ac.in}{skvipparthi@iitrpr.ac.in}), \\
Subrahmanyam Murala$^8$
(\href{mailto:muralas@tcd.ie}{muralas@tcd.ie}),\\
\noindent\textit{\textbf{Affiliations: }} \\ 
$^1$ Amity University Punjab, India \\
$^2$ Mohamed Bin Zayed University of Artificial Intelligence, Abu Dhabi \\
$^3$ Indian Institute of Technology Mandi, India \\
$^4$ UPES Dehradun, India \\
$^5$ Government Engineering College Ajmer, India \\
$^6$ Indian Institute of Technology Guwahati, India \\
$^7$ Indian Institute of Technology Ropar, India \\
$^8$ Trinity College Dublin, Ireland \\

% --- Teams without ranking
\subsection*{SylabSR}
\noindent\textit{\textbf{Title: }} AutoRegressive Residual Local Feature Network\\
\noindent\textit{\textbf{Members: }} \\
Wei-Chen Shen$^1$ (\href{mailto:r11921a38@ntu.edu.tw}{r11921a38@ntu.edu.tw}),\\
I-Hsiang Chen$^{1,2}$ , \\
\noindent\textit{\textbf{Affiliations: }} \\ 
$^1$ National Taiwan University \\
$^2$ University of Washington \\

\subsection*{NJUPCA}
\noindent\textit{\textbf{Title: }} Spatial-Frequency Fusion Model for Efficient Super-Resolution\\
\noindent\textit{\textbf{Members: }} \\
Yunzhe Xu$^1$ (\href{mailto:221900144@smail.nju.edu.cn}{221900144@smail.nju.edu.cn}),\\
Chen Zhao$^1$, \\
Zhizhou Chen$^1$, \\
\noindent\textit{\textbf{Affiliations: }} \\ 
$^1$ Nanjing University \\

\subsection*{DepthIBN}
\noindent\textit{\textbf{Title: }}  Involution and BSConv Multi-Depth Distillation Network for Lightweight Image Super-Resolution\\
\noindent\textit{\textbf{Members: }} \\
Akram Khatami-Rizi $^1$ (\href{mailto:yourEmail@xxx.xxx}{akramkhatami67@gmail.com}),\\
Ahmad Mahmoudi-Aznaveh $^1$, (\href{mailto:yourEmail@xxx.xxx}{a\_mahmoudi@sbu.ac.ir}), \\
\noindent\textit{\textbf{Affiliations: }} \\ 
$^1$ Cyberspace Research Institute of Shahid Beheshti University of Iran \\

\subsection*{Cidaut\_AI}
\noindent\textit{\textbf{Title: }} Fused Edge Attention Network\\
\noindent\textit{\textbf{Members: }} \\
Alejandro Merino$^1$ (\href{mailto:alemer@cidaut.es}{alemer@cidaut.es}),\\
Bruno Longarela$^1$ (\href{mailto:brulon@cidaut.es}{brulon@cidaut.es}), \\
Javier Abad$^1$ (\href{mailto:javaba@cidaut.es}{javaba@cidaut.es}),  \\
Marcos V. Conde$^2$ (\href{mailto:marcos.conde@uni-wuerzburg.de}{marcos.conde@uni-wuerzburg.de}),  \\
\noindent\textit{\textbf{Affiliations: }} \\ 
$^1$ Cidaut AI, Spain \\
$^2$ University of W\"urzburg, Germany \\

\subsection*{IVL}
\noindent\textit{\textbf{Title: }} PAEDN\\
\noindent\textit{\textbf{Members: }} \\
Simone Bianco$^1$ (\href{mailto:simone.bianco@unimib.com}{simone.bianco@unimib.com}),\\
Luca Cogo$^1$ (\href{mailto:luca.cogo@unimib.com}{luca.cogo@unimib.com}),\\
Gianmarco Corti$^1$ (\href{mailto:g.corti1967@campus.unimib.com}{g.corti1967@campus.unimib.com}),\\

\noindent\textit{\textbf{Affiliations: }} \\ 
$^1$ Department of Informatics Systems and Communication, University of Milano-Bicocca, Viale Sarca 336, Building U14, Milan, Italy \\

%%%%%%%%% References %%%%%%%%%
% \clearpage % This can be deleted later
{
    \small
    \bibliographystyle{ieeenat_fullname}
    \bibliography{ref}

\begin{thebibliography}{149}
\providecommand{\natexlab}[1]{#1}
\providecommand{\url}[1]{\texttt{#1}}
\expandafter\ifx\csname urlstyle\endcsname\relax
  \providecommand{\doi}[1]{doi: #1}\else
  \providecommand{\doi}{doi: \begingroup \urlstyle{rm}\Url}\fi

\bibitem[Abrahamyan et~al.(2022)Abrahamyan, Truong, Philips, and Deligiannis]{abrahamyan2022gradient}
Lusine Abrahamyan, Anh~Minh Truong, Wilfried Philips, and Nikos Deligiannis.
\newblock Gradient variance loss for structure-enhanced image super-resolution.
\newblock In \emph{ICASSP 2022-2022 IEEE International Conference on Acoustics, Speech and Signal Processing (ICASSP)}, pages 3219--3223. IEEE, 2022.

\bibitem[Agustsson and Timofte(2017{\natexlab{a}})]{8014884}
Eirikur Agustsson and Radu Timofte.
\newblock Ntire 2017 challenge on single image super-resolution: Dataset and study.
\newblock In \emph{2017 IEEE Conference on Computer Vision and Pattern Recognition Workshops (CVPRW)}, pages 1122--1131, 2017{\natexlab{a}}.

\bibitem[Agustsson and Timofte(2017{\natexlab{b}})]{Agustsson_2017_CVPR_Workshops}
Eirikur Agustsson and Radu Timofte.
\newblock Ntire 2017 challenge on single image super-resolution: Dataset and study.
\newblock In \emph{The IEEE Conference on Computer Vision and Pattern Recognition (CVPR) Workshops}, 2017{\natexlab{b}}.

\bibitem[Agustsson and Timofte(2017{\natexlab{c}})]{agustsson2017ntire}
Eirikur Agustsson and Radu Timofte.
\newblock {NTIRE} 2017 challenge on single image super-resolution: Dataset and study.
\newblock In \emph{Proceedings of the IEEE Conference on Computer Vision and Pattern Recognition Workshops}, pages 126--135, 2017{\natexlab{c}}.

\bibitem[Ahmad Mahmoudi-Aznaveh(2024)]{deepf}
Akram Khatami-Rizi Ahmad Mahmoudi-Aznaveh.
\newblock The role of involution in lightweight super resolution.
\newblock \emph{2024 13th Iranian/3rd International Machine Vision and Image Processing Conference (MVIP)}, 2024.

\bibitem[Ahmad Mahmoudi-Aznaveh(2025)]{ibmdn}
Akram Khatami-Rizi Ahmad Mahmoudi-Aznaveh.
\newblock Involution and bsconv multi-depth distillation network for lightweight image super-resolution.
\newblock \emph{arXiv preprint arXiv:2503.14779}, 2025.

\bibitem[Aleem et~al.(2024)Aleem, Dietlmeier, Arazo, and Little]{aleem2024convlora}
Sidra Aleem, Julia Dietlmeier, Eric Arazo, and Suzanne Little.
\newblock Convlora and adabn based domain adaptation via self-training.
\newblock In \emph{2024 IEEE International Symposium on Biomedical Imaging (ISBI)}, pages 1--5. IEEE, 2024.

\bibitem[Cao et~al.(2023)Cao, Wang, Xian, Li, Ni, Pi, Zhang, Zhang, Timofte, and Van~Gool]{cao2023ciaosr}
Jiezhang Cao, Qin Wang, Yongqin Xian, Yawei Li, Bingbing Ni, Zhiming Pi, Kai Zhang, Yulun Zhang, Radu Timofte, and Luc Van~Gool.
\newblock Ciaosr: Continuous implicit attention-in-attention network for arbitrary-scale image super-resolution.
\newblock In \emph{Proceedings of the IEEE/CVF Conference on Computer Vision and Pattern Recognition}, pages 1796--1807, 2023.

\bibitem[Chen et~al.(2023)Chen, Kao, He, Zhuo, Wen, Lee, and Chan]{chen2023run}
Jierun Chen, Shiu-hong Kao, Hao He, Weipeng Zhuo, Song Wen, Chul-Ho Lee, and S-H~Gary Chan.
\newblock Run, don't walk: Chasing higher flops for faster neural networks.
\newblock In \emph{IEEE Conf. Comput. Vis. Pattern Recog.}, 2023.

\bibitem[Chen et~al.(2022)Chen, Chu, Zhang, and Sun]{chen2022simplebaselinesimagerestoration}
Liangyu Chen, Xiaojie Chu, Xiangyu Zhang, and Jian Sun.
\newblock Simple baselines for image restoration, 2022.

\bibitem[Chen et~al.(2024)Chen, Wu, Zamfir, Zhang, Zhang, Timofte, Yang, Yu, Wan, Hong, et~al.]{chen2024ntire}
Zheng Chen, Zongwei Wu, Eduard Zamfir, Kai Zhang, Yulun Zhang, Radu Timofte, Xiaokang Yang, Hongyuan Yu, Cheng Wan, Yuxin Hong, et~al.
\newblock Ntire 2024 challenge on image super-resolution (x4): Methods and results.
\newblock In \emph{Proceedings of the IEEE/CVF Conference on Computer Vision and Pattern Recognition}, pages 6108--6132, 2024.

\bibitem[Chen et~al.(2025{\natexlab{a}})Chen, Liu, Gong, Wang, Sun, Wu, Timofte, Zhang, et~al.]{ntire2025srx4}
Zheng Chen, Kai Liu, Jue Gong, Jingkai Wang, Lei Sun, Zongwei Wu, Radu Timofte, Yulun Zhang, et~al.
\newblock {NTIRE} 2025 challenge on image super-resolution (×4): Methods and results.
\newblock In \emph{Proceedings of the IEEE/CVF Conference on Computer Vision and Pattern Recognition (CVPR) Workshops}, 2025{\natexlab{a}}.

\bibitem[Chen et~al.(2025{\natexlab{b}})Chen, Wang, Liu, Gong, Sun, Wu, Timofte, Zhang, et~al.]{ntire2025face}
Zheng Chen, Jingkai Wang, Kai Liu, Jue Gong, Lei Sun, Zongwei Wu, Radu Timofte, Yulun Zhang, et~al.
\newblock {NTIRE} 2025 challenge on real-world face restoration: Methods and results.
\newblock In \emph{Proceedings of the IEEE/CVF Conference on Computer Vision and Pattern Recognition (CVPR) Workshops}, 2025{\natexlab{b}}.

\bibitem[Cho et~al.(2021{\natexlab{a}})Cho, Ji, Hong, Jung, and Ko]{MIMO}
Sung-Jin Cho, Seo-Won Ji, Jun-Pyo Hong, Seung-Won Jung, and Sung-Jea Ko.
\newblock Rethinking coarse-to-fine approach in single image deblurring.
\newblock In \emph{ICCV}, 2021{\natexlab{a}}.

\bibitem[Cho et~al.(2021{\natexlab{b}})Cho, Ji, Hong, Jung, and Ko]{cho2021rethinking}
Sung-Jin Cho, Seo-Won Ji, Jun-Pyo Hong, Seung-Won Jung, and Sung-Jea Ko.
\newblock Rethinking coarse-to-fine approach in single image deblurring.
\newblock In \emph{ICCV}, pages 4641--4650, 2021{\natexlab{b}}.

\bibitem[Conde et~al.(2025{\natexlab{a}})Conde, Timofte, et~al.]{ntire2025raw}
Marcos Conde, Radu Timofte, et~al.
\newblock {NTIRE} 2025 challenge on raw image restoration and super-resolution.
\newblock In \emph{Proceedings of the IEEE/CVF Conference on Computer Vision and Pattern Recognition (CVPR) Workshops}, 2025{\natexlab{a}}.

\bibitem[Conde et~al.(2025{\natexlab{b}})Conde, Timofte, et~al.]{ntire2025rawrgb}
Marcos Conde, Radu Timofte, et~al.
\newblock Raw image reconstruction from {RGB} on smartphones. {NTIRE} 2025 challenge report.
\newblock In \emph{Proceedings of the IEEE/CVF Conference on Computer Vision and Pattern Recognition (CVPR) Workshops}, 2025{\natexlab{b}}.

\bibitem[Conde et~al.(2024)Conde, Lei, Li, Bampis, Katsavounidis, and Timofte]{conde2024aim}
Marcos~V Conde, Zhijun Lei, Wen Li, Christos Bampis, Ioannis Katsavounidis, and Radu Timofte.
\newblock Aim 2024 challenge on efficient video super-resolution for av1 compressed content.
\newblock \emph{arXiv preprint arXiv:2409.17256}, 2024.

\bibitem[Deng et~al.(2023)Deng, Yuan, Deng, and Lu]{deng2023reparameterized}
Weijian Deng, Hongjie Yuan, Lunhui Deng, and Zengtong Lu.
\newblock Reparameterized residual feature network for lightweight image super-resolution.
\newblock In \emph{Proceedings of the IEEE/CVF conference on computer vision and pattern recognition}, pages 1712--1721, 2023.

\bibitem[Ding et~al.(2019)Ding, Guo, Ding, and Han]{ding2019acnet}
Xiaohan Ding, Yuchen Guo, Guiguang Ding, and Jungong Han.
\newblock Acnet: Strengthening the kernel skeletons for powerful cnn via asymmetric convolution blocks.
\newblock In \emph{Proceedings of the IEEE/CVF international conference on computer vision}, pages 1911--1920, 2019.

\bibitem[Ding et~al.(2021{\natexlab{a}})Ding, Zhang, Han, and Ding]{ding2021diverse}
Xiaohan Ding, Xiangyu Zhang, Jungong Han, and Guiguang Ding.
\newblock Diverse branch block: Building a convolution as an inception-like unit.
\newblock In \emph{Proceedings of the IEEE/CVF Conference on Computer Vision and Pattern Recognition}, pages 10886--10895, 2021{\natexlab{a}}.

\bibitem[Ding et~al.(2021{\natexlab{b}})Ding, Zhang, Ma, Han, Ding, and Sun]{ding2021repvgg}
Xiaohan Ding, Xiangyu Zhang, Ningning Ma, Jungong Han, Guiguang Ding, and Jian Sun.
\newblock Repvgg: Making vgg-style convnets great again.
\newblock In \emph{Proceedings of the IEEE/CVF Conference on Computer Vision and Pattern Recognition}, pages 13733--13742, 2021{\natexlab{b}}.

\bibitem[Ding et~al.(2021{\natexlab{c}})Ding, Zhang, Ma, Han, Ding, and Sun]{repvgg}
Xiaohan Ding, Xiangyu Zhang, Ningning Ma, Jungong Han, Guiguang Ding, and Jian Sun.
\newblock Repvgg: Making vgg-style convnets great again.
\newblock In \emph{CVPR}, 2021{\natexlab{c}}.

\bibitem[Du et~al.(2022{\natexlab{a}})Du, Guan, Zhou, Li, and Wang]{du2022parameter}
Jie Du, Kai Guan, Yanhong Zhou, Yuanman Li, and Tianfu Wang.
\newblock Parameter-free similarity-aware attention module for medical image classification and segmentation.
\newblock \emph{IEEE Transactions on Emerging Topics in Computational Intelligence}, 2022{\natexlab{a}}.

\bibitem[Du et~al.(2022{\natexlab{b}})Du, Liu, Liu, Tang, Wu, and Fu]{FMEN}
Zongcai Du, Ding Liu, Jie Liu, Jie Tang, Gangshan Wu, and Lean Fu.
\newblock Fast and memory-efficient network towards efficient image super-resolution.
\newblock In \emph{Proceedings of the IEEE/CVF Conference on Computer Vision and Pattern Recognition}, pages 853--862, 2022{\natexlab{b}}.

\bibitem[Du et~al.(2022{\natexlab{c}})Du, Liu, Liu, Tang, Wu, and Fu]{du2022fast}
Zongcai Du, Ding Liu, Jie Liu, Jie Tang, Gangshan Wu, and Lean Fu.
\newblock Fast and memory-efficient network towards efficient image super-resolution.
\newblock In \emph{Proceedings of the IEEE/CVF Conference on Computer Vision and Pattern Recognition}, pages 853--862, 2022{\natexlab{c}}.

\bibitem[Elfwing et~al.(2017)Elfwing, Uchibe, and Doya]{silu}
Stefan Elfwing, Eiji Uchibe, and Kenji Doya.
\newblock Sigmoid-weighted linear units for neural network function approximation in reinforcement learning, 2017.

\bibitem[Ershov et~al.(2025)Ershov, Korchagin, Khalin, Panshin, Terekhin, Zaychenkova, Lobarev, Plokhotnyuk, Abramov, Zhdanov, Dorogova, Mamedov, Banic, Perevozchikov, Timofte, et~al.]{ntire2025night}
Egor Ershov, Sergey Korchagin, Alexei Khalin, Artyom Panshin, Arseniy Terekhin, Ekaterina Zaychenkova, Georgiy Lobarev, Vsevolod Plokhotnyuk, Denis Abramov, Elisey Zhdanov, Sofia Dorogova, Yasin Mamedov, Nikola Banic, Georgii Perevozchikov, Radu Timofte, et~al.
\newblock {NTIRE} 2025 challenge on night photography rendering.
\newblock In \emph{Proceedings of the IEEE/CVF Conference on Computer Vision and Pattern Recognition (CVPR) Workshops}, 2025.

\bibitem[Fu et~al.(2025)Fu, Qiu, Fu, Timofte, Sebe, Yang, Van~Gool, et~al.]{ntire2025cross}
Yuqian Fu, Xingyu Qiu, Bin Ren~Yanwei Fu, Radu Timofte, Nicu Sebe, Ming-Hsuan Yang, Luc Van~Gool, et~al.
\newblock {NTIRE} 2025 challenge on cross-domain few-shot object detection: Methods and results.
\newblock In \emph{Proceedings of the IEEE/CVF Conference on Computer Vision and Pattern Recognition (CVPR) Workshops}, 2025.

\bibitem[Gu and Dao(2023)]{mamba}
Albert Gu and Tri Dao.
\newblock Mamba: Linear-time sequence modeling with selective state spaces.
\newblock \emph{arXiv preprint arXiv:2312.00752}, 2023.

\bibitem[Gu et~al.(2024)Gu, Ge, and Guo]{gu2024code}
Enxuan Gu, Hongwei Ge, and Yong Guo.
\newblock Code: An explicit content decoupling framework for image restoration.
\newblock In \emph{Proceedings of the IEEE/CVF Conference on Computer Vision and Pattern Recognition}, pages 2920--2930, 2024.

\bibitem[Guo et~al.(2024{\natexlab{a}})Guo, Guo, Zha, Zhang, Li, Dai, Xia, and Li]{guo2024mambairv2}
Hang Guo, Yong Guo, Yaohua Zha, Yulun Zhang, Wenbo Li, Tao Dai, Shu-Tao Xia, and Yawei Li.
\newblock Mambairv2: Attentive state space restoration.
\newblock \emph{arXiv preprint arXiv:2411.15269}, 2024{\natexlab{a}}.

\bibitem[Guo et~al.(2024{\natexlab{b}})Guo, Li, Dai, Ouyang, Ren, and Xia]{guo2025mambair}
Hang Guo, Jinmin Li, Tao Dai, Zhihao Ouyang, Xudong Ren, and Shu-Tao Xia.
\newblock Mambair: A simple baseline for image restoration with state-space model.
\newblock In \emph{European Conference on Computer Vision}, pages 222--241. Springer, 2024{\natexlab{b}}.

\bibitem[Haase and Amthor(2020)]{BSConv}
Daniel Haase and Manuel Amthor.
\newblock Rethinking depthwise separable convolutions: How intra-kernel correlations lead to improved mobilenets.
\newblock In \emph{Proceedings of the IEEE/CVF conference on computer vision and pattern recognition}, pages 14600--14609, 2020.

\bibitem[Han et~al.(2020)Han, Wang, Tian, Guo, Xu, and Xu]{ghostnet}
Kai Han, Yunhe Wang, Qi Tian, Jianyuan Guo, Chunjing Xu, and Chang Xu.
\newblock Ghostnet: More features from cheap operations.
\newblock In \emph{IEEE Conf. Comput. Vis. Pattern Recog.}, pages 1580--1589, 2020.

\bibitem[Han et~al.(2025)Han, Fan, Kong, Liao, Guo, Li, Timofte, et~al.]{ntire2025text}
Shuhao Han, Haotian Fan, Fangyuan Kong, Wenjie Liao, Chunle Guo, Chongyi Li, Radu Timofte, et~al.
\newblock {NTIRE} 2025 challenge on text to image generation model quality assessment.
\newblock In \emph{Proceedings of the IEEE/CVF Conference on Computer Vision and Pattern Recognition (CVPR) Workshops}, 2025.

\bibitem[He et~al.(2020)He, Dai, Lu, Jiang, and Xia]{he2020fakd}
Zibin He, Tao Dai, Jian Lu, Yong Jiang, and Shu-Tao Xia.
\newblock Fakd: Feature-affinity based knowledge distillation for efficient image super-resolution.
\newblock In \emph{2020 IEEE international conference on image processing (ICIP)}, pages 518--522. IEEE, 2020.

\bibitem[Hendrycks and Gimpel(2016)]{hendrycks2016gaussian}
Dan Hendrycks and Kevin Gimpel.
\newblock Gaussian error linear units (gelus).
\newblock \emph{arXiv preprint arXiv:1606.08415}, 2016.

\bibitem[Howard et~al.(2019)Howard, Sandler, Chu, Chen, Chen, Tan, Wang, Zhu, Pang, Vasudevan, et~al.]{howard2019searching}
Andrew Howard, Mark Sandler, Grace Chu, Liang-Chieh Chen, Bo Chen, Mingxing Tan, Weijun Wang, Yukun Zhu, Ruoming Pang, Vijay Vasudevan, et~al.
\newblock Searching for {MobileNetV3}.
\newblock In \emph{Proceedings of the IEEE International Conference on Computer Vision}, pages 1314--1324, 2019.

\bibitem[Howard et~al.(2017)Howard, Zhu, Chen, Kalenichenko, Wang, Weyand, Andreetto, and Adam]{Mobilenets}
Andrew~G Howard, Menglong Zhu, Bo Chen, Dmitry Kalenichenko, Weijun Wang, Tobias Weyand, Marco Andreetto, and Hartwig Adam.
\newblock Mobilenets: Efficient convolutional neural networks for mobile vision applications.
\newblock \emph{arXiv preprint arXiv:1704.04861}, 2017.

\bibitem[Hu et~al.(2022)Hu, Feng, Hua, Lai, Huang, Gong, and Hua]{Online_Convolutional_Re-parameterization}
Mu Hu, Junyi Feng, Jiashen Hua, Baisheng Lai, Jianqiang Huang, Xiaojin Gong, and Xian{-}Sheng Hua.
\newblock Online convolutional re-parameterization.
\newblock \emph{CoRR}, abs/2204.00826, 2022.

\bibitem[Huang et~al.(2022)Huang, Zhang, Heng, Shi, and Zhou]{rife}
Zhewei Huang, Tianyuan Zhang, Wen Heng, Boxin Shi, and Shuchang Zhou.
\newblock Real-time intermediate flow estimation for video frame interpolation.
\newblock In \emph{Proceedings of the European Conference on Computer Vision (ECCV)}, 2022.

\bibitem[Hui et~al.(2018)Hui, Wang, and Gao]{IDN}
Zheng Hui, Xiumei Wang, and Xinbo Gao.
\newblock Fast and accurate single image super-resolution via information distillation network.
\newblock In \emph{Proceedings of the IEEE Conference on Computer Vision and Pattern Recognition}, pages 723--731, 2018.

\bibitem[Hui et~al.(2019{\natexlab{a}})Hui, Gao, Yang, and Wang]{CCA}
Zheng Hui, Xinbo Gao, Yunchu Yang, and Xiumei Wang.
\newblock Lightweight image super-resolution with information multi-distillation network.
\newblock In \emph{Proceedings of the 27th acm international conference on multimedia}, pages 2024--2032, 2019{\natexlab{a}}.

\bibitem[Hui et~al.(2019{\natexlab{b}})Hui, Gao, Yang, and Wang]{IMDN}
Zheng Hui, Xinbo Gao, Yunchu Yang, and Xiumei Wang.
\newblock Lightweight image super-resolution with information multi-distillation network.
\newblock In \emph{Proceedings of the 27th acm international conference on multimedia}, pages 2024--2032, 2019{\natexlab{b}}.

\bibitem[Izmailov et~al.(2018)Izmailov, Podoprikhin, Garipov, Vetrov, and Wilson]{izmailov2018averaging}
Pavel Izmailov, Dmitrii Podoprikhin, Timur Garipov, Dmitry Vetrov, and Andrew~Gordon Wilson.
\newblock Averaging weights leads to wider optima and better generalization.
\newblock \emph{arXiv preprint arXiv:1803.05407}, 2018.

\bibitem[Jain et~al.(2025)Jain, Wu, Zou, Florentin, Turbell, Siddhartha, Timofte, et~al.]{ntire2025vqe}
Varun Jain, Zongwei Wu, Quan Zou, Louis Florentin, Henrik Turbell, Sandeep Siddhartha, Radu Timofte, et~al.
\newblock {NTIRE} 2025 challenge on video quality enhancement for video conferencing: Datasets, methods and results.
\newblock In \emph{Proceedings of the IEEE/CVF Conference on Computer Vision and Pattern Recognition (CVPR) Workshops}, 2025.

\bibitem[Jiang et~al.(2024{\natexlab{a}})Jiang, Feng, Zhang, and Bull]{jiang2024mtkd}
Yuxuan Jiang, Chen Feng, Fan Zhang, and David Bull.
\newblock Mtkd: Multi-teacher knowledge distillation for image super-resolution.
\newblock In \emph{European Conference on Computer Vision}, pages 364--382. Springer, 2024{\natexlab{a}}.

\bibitem[Jiang et~al.(2024{\natexlab{b}})Jiang, Kwan, Peng, Gao, Zhang, Zhu, Sole, and Bull]{jiang2024hiif}
Yuxuan Jiang, Ho~Man Kwan, Tianhao Peng, Ge Gao, Fan Zhang, Xiaoqing Zhu, Joel Sole, and David Bull.
\newblock {HIIF}: Hierarchical encoding based implicit image function for continuous super-resolution.
\newblock \emph{arXiv preprint arXiv:2412.03748}, 2024{\natexlab{b}}.

\bibitem[Jiang et~al.(2024{\natexlab{c}})Jiang, Nawa{\l}a, Feng, Zhang, Zhu, Sole, and Bull]{jiang2024rtsr}
Yuxuan Jiang, Jakub Nawa{\l}a, Chen Feng, Fan Zhang, Xiaoqing Zhu, Joel Sole, and David Bull.
\newblock Rtsr: A real-time super-resolution model for av1 compressed content.
\newblock \emph{arXiv preprint arXiv:2411.13362}, 2024{\natexlab{c}}.

\bibitem[Jiang et~al.(2024{\natexlab{d}})Jiang, Nawa{\l}a, Zhang, and Bull]{jiang2024compressing}
Yuxuan Jiang, Jakub Nawa{\l}a, Fan Zhang, and David Bull.
\newblock Compressing deep image super-resolution models.
\newblock In \emph{2024 Picture Coding Symposium (PCS)}, pages 1--5. IEEE, 2024{\natexlab{d}}.

\bibitem[Jiang et~al.(2025)Jiang, Zeng, Teng, Zhang, Zhu, Sole, and Bull]{jiang2025c2d}
Yuxuan Jiang, Chengxi Zeng, Siyue Teng, Fan Zhang, Xiaoqing Zhu, Joel Sole, and David Bull.
\newblock {C2D-ISR}: Optimizing attention-based image super-resolution from continuous to discrete scales.
\newblock \emph{arXiv preprint arXiv:2503.13740}, 2025.

\bibitem[Kingma and Ba(2014{\natexlab{a}})]{Adam}
Diederik~P Kingma and Jimmy Ba.
\newblock Adam: A method for stochastic optimization.
\newblock \emph{arXiv preprint arXiv:1412.6980}, 2014{\natexlab{a}}.

\bibitem[Kingma and Ba(2014{\natexlab{b}})]{kingma2014adam}
Diederik~P Kingma and Jimmy Ba.
\newblock Adam: A method for stochastic optimization.
\newblock \emph{arXiv preprint arXiv:1412.6980}, 2014{\natexlab{b}}.

\bibitem[Kong et~al.(2022{\natexlab{a}})Kong, Li, Liu, Liu, He, Bai, Chen, and Fu]{RLFN}
F. Kong, Mingxi Li, Songwei Liu, Ding Liu, Jingwen He, Yang Bai, Fangmin Chen, and Lean Fu.
\newblock Residual local feature network for efficient super-resolution.
\newblock \emph{2022 IEEE/CVF Conference on Computer Vision and Pattern Recognition Workshops (CVPRW)}, pages 765--775, 2022{\natexlab{a}}.

\bibitem[Kong et~al.(2022{\natexlab{b}})Kong, Li, Liu, Liu, He, Bai, Chen, and Fu]{kong2022residual}
Fangyuan Kong, Mingxi Li, Songwei Liu, Ding Liu, Jingwen He, Yang Bai, Fangmin Chen, and Lean Fu.
\newblock Residual local feature network for efficient super-resolution.
\newblock In \emph{Proceedings of the IEEE/CVF Conference on Computer Vision and Pattern Recognition}, pages 766--776, 2022{\natexlab{b}}.

\bibitem[Lau et~al.(2023)Lau, Po, and Rehman]{lau2024large}
Kin~Wai Lau, Lai-Man Po, and Yasar Abbas~Ur Rehman.
\newblock Large separable kernel attention: Rethinking the large kernel attention design in cnn.
\newblock \emph{Expert Systems with Applications}, 236:\penalty0 121352, 2023.

\bibitem[Lee et~al.(2025)Lee, Park, Canelo, Park, Kim, Chun, Jin, Li, Guo, Timofte, et~al.]{ntire2025ebhdr}
Sangmin Lee, Eunpil Park, Angel Canelo, Hyunhee Park, Youngjo Kim, Hyungju Chun, Xin Jin, Chongyi Li, Chun-Le Guo, Radu Timofte, et~al.
\newblock {NTIRE} 2025 challenge on efficient burst hdr and restoration: Datasets, methods, and results.
\newblock In \emph{Proceedings of the IEEE/CVF Conference on Computer Vision and Pattern Recognition (CVPR) Workshops}, 2025.

\bibitem[Lei et~al.(2024)Lei, Zhang, and Cao]{Lei_2024_CVPR}
Xiaoyan Lei, Wenlong Zhang, and Weifeng Cao.
\newblock Dvmsr: Distillated vision mamba for efficient super-resolution.
\newblock In \emph{Proceedings of the IEEE/CVF Conference on Computer Vision and Pattern Recognition (CVPR) Workshops}, pages 6536--6546, 2024.

\bibitem[Li et~al.(2021)Li, Hu, Wang, Li, She, Zhu, Zhang, and Chen]{invo}
Duo Li, Jie Hu, Changhu Wang, Xiangtai Li, Qi She, Lei Zhu, Tong Zhang, and Qifeng Chen.
\newblock Involution: Inverting the inherence of convolution for visual recognition.
\newblock \emph{Proceedings of the IEEE/CVF Conference on Computer Vision and Pattern Recognition}, 2021.

\bibitem[Li et~al.(2025{\natexlab{a}})Li, Jin, Jin, Wu, Li, Wang, Yang, Li, Chen, Wen, Tan, Timofte, et~al.]{ntire2025day}
Xin Li, Yeying Jin, Xin Jin, Zongwei Wu, Bingchen Li, Yufei Wang, Wenhan Yang, Yu Li, Zhibo Chen, Bihan Wen, Robby Tan, Radu Timofte, et~al.
\newblock {NTIRE} 2025 challenge on day and night raindrop removal for dual-focused images: Methods and results.
\newblock In \emph{Proceedings of the IEEE/CVF Conference on Computer Vision and Pattern Recognition (CVPR) Workshops}, 2025{\natexlab{a}}.

\bibitem[Li et~al.(2025{\natexlab{b}})Li, Wang, Li, Yuan, Shao, Yao, Sun, Zhou, Timofte, and Chen]{ntire2025shortugc_data}
Xin Li, Xijun Wang, Bingchen Li, Kun Yuan, Yizhen Shao, Suhang Yao, Ming Sun, Chao Zhou, Radu Timofte, and Zhibo Chen.
\newblock {NTIRE} 2025 challenge on short-form ugc video quality assessment and enhancement: Kwaisr dataset and study.
\newblock In \emph{Proceedings of the IEEE/CVF Conference on Computer Vision and Pattern Recognition (CVPR) Workshops}, 2025{\natexlab{b}}.

\bibitem[Li et~al.(2025{\natexlab{c}})Li, Yuan, Li, Guan, Shao, Yu, Wang, Lu, Luo, Yao, Sun, Zhou, Chen, Timofte, et~al.]{ntire2025shortugc}
Xin Li, Kun Yuan, Bingchen Li, Fengbin Guan, Yizhen Shao, Zihao Yu, Xijun Wang, Yiting Lu, Wei Luo, Suhang Yao, Ming Sun, Chao Zhou, Zhibo Chen, Radu Timofte, et~al.
\newblock {NTIRE} 2025 challenge on short-form ugc video quality assessment and enhancement: Methods and results.
\newblock In \emph{Proceedings of the IEEE/CVF Conference on Computer Vision and Pattern Recognition (CVPR) Workshops}, 2025{\natexlab{c}}.

\bibitem[Li et~al.(2023{\natexlab{a}})Li, Zhang, Liang, Cao, Liu, Gong, Zhang, Tang, Liu, Demandolx, et~al.]{li2023lsdir}
Yawei Li, Kai Zhang, Jingyun Liang, Jiezhang Cao, Ce Liu, Rui Gong, Yulun Zhang, Hao Tang, Yun Liu, Denis Demandolx, et~al.
\newblock Lsdir: A large scale dataset for image restoration.
\newblock In \emph{Proceedings of the IEEE/CVF Conference on Computer Vision and Pattern Recognition Workshops}, 2023{\natexlab{a}}.

\bibitem[Li et~al.(2023{\natexlab{b}})Li, Zhang, Van~Gool, Timofte, et~al.]{li2023ntire_esr}
Yawei Li, Yulun Zhang, Luc Van~Gool, Radu Timofte, et~al.
\newblock {NTIRE} 2023 challenge on efficient super-resolution: Methods and results.
\newblock In \emph{Proceedings of the IEEE/CVF Conference on Computer Vision and Pattern Recognition Workshops}, 2023{\natexlab{b}}.

\bibitem[Li et~al.(2022{\natexlab{a}})Li, Liu, Chen, Cai, Gu, Qiao, and Dong]{9857155}
Zheyuan Li, Yingqi Liu, Xiangyu Chen, Haoming Cai, Jinjin Gu, Yu Qiao, and Chao Dong.
\newblock Blueprint separable residual network for efficient image super-resolution.
\newblock In \emph{2022 IEEE/CVF Conference on Computer Vision and Pattern Recognition Workshops (CVPRW)}, pages 832--842, 2022{\natexlab{a}}.

\bibitem[Li et~al.(2022{\natexlab{b}})Li, Liu, Chen, Cai, Gu, Qiao, and Dong]{BSRN}
Zheyuan Li, Yingqi Liu, Xiangyu Chen, Haoming Cai, Jinjin Gu, Yu Qiao, and Chao Dong.
\newblock Blueprint separable residual network for efficient image super-resolution.
\newblock In \emph{Proceedings of the IEEE/CVF Conference on Computer Vision and Pattern Recognition}, pages 833--843, 2022{\natexlab{b}}.

\bibitem[Liang et~al.(2025)Liang, Timofte, Yi, Zhang, Liu, Sun, Wu, Zhang, Zeng, Zhang, et~al.]{ntire2025raim}
Jie Liang, Radu Timofte, Qiaosi Yi, Zhengqiang Zhang, Shuaizheng Liu, Lingchen Sun, Rongyuan Wu, Xindong Zhang, Hui Zeng, Lei Zhang, et~al.
\newblock {NTIRE} 2025 the 2nd restore any image model {(RAIM)} in the wild challenge.
\newblock In \emph{Proceedings of the IEEE/CVF Conference on Computer Vision and Pattern Recognition (CVPR) Workshops}, 2025.

\bibitem[Lim et~al.(2017{\natexlab{a}})Lim, Son, Kim, Nah, and Lee]{8014885}
Bee Lim, Sanghyun Son, Heewon Kim, Seungjun Nah, and Kyoung~Mu Lee.
\newblock Enhanced deep residual networks for single image super-resolution.
\newblock In \emph{2017 IEEE Conference on Computer Vision and Pattern Recognition Workshops (CVPRW)}, pages 1132--1140, 2017{\natexlab{a}}.

\bibitem[Lim et~al.(2017{\natexlab{b}})Lim, Son, Kim, Nah, and Lee]{lim2017enhanced}
Bee Lim, Sanghyun Son, Heewon Kim, Seungjun Nah, and Kyoung~Mu Lee.
\newblock Enhanced deep residual networks for single image super-resolution.
\newblock In \emph{Proceedings of the IEEE Conference on Computer Vision and Pattern Recognition Workshops}, pages 1132--1140, 2017{\natexlab{b}}.

\bibitem[Liu et~al.(2020{\natexlab{a}})Liu, Tang, and Wu]{RFDN}
Jie Liu, Jie Tang, and Gangshan Wu.
\newblock Residual feature distillation network for lightweight image super-resolution.
\newblock In \emph{Proceedings of the European Conference on Computer Vision Workshops}, pages 41--55. Springer, 2020{\natexlab{a}}.

\bibitem[Liu et~al.(2020{\natexlab{b}})Liu, Tang, and Wu]{liu2020residual}
Jie Liu, Jie Tang, and Gangshan Wu.
\newblock Residual feature distillation network for lightweight image super-resolution.
\newblock In \emph{Computer Vision--ECCV 2020 Workshops: Glasgow, UK, August 23--28, 2020, Proceedings, Part III 16}, pages 41--55. Springer, 2020{\natexlab{b}}.

\bibitem[Liu et~al.(2020{\natexlab{c}})Liu, Zhang, Tang, Tang, and Wu]{ESA}
Jie Liu, Wenjie Zhang, Yuting Tang, Jie Tang, and Gangshan Wu.
\newblock Residual feature aggregation network for image super-resolution.
\newblock In \emph{Proceedings of the IEEE/CVF conference on computer vision and pattern recognition}, pages 2359--2368, 2020{\natexlab{c}}.

\bibitem[Liu et~al.(2025{\natexlab{a}})Liu, Min, Hu, Zhang, Guo, et~al.]{ntire2025xgc}
Xiaohong Liu, Xiongkuo Min, Qiang Hu, Xiaoyun Zhang, Jie Guo, et~al.
\newblock {NTIRE} 2025 {XGC} quality assessment challenge: Methods and results.
\newblock In \emph{Proceedings of the IEEE/CVF Conference on Computer Vision and Pattern Recognition (CVPR) Workshops}, 2025{\natexlab{a}}.

\bibitem[Liu et~al.(2025{\natexlab{b}})Liu, Wu, Vasluianu, Yan, Ren, Zhang, Gu, Zhang, Zhu, Timofte, et~al.]{ntire2025lowlight}
Xiaoning Liu, Zongwei Wu, Florin-Alexandru Vasluianu, Hailong Yan, Bin Ren, Yulun Zhang, Shuhang Gu, Le Zhang, Ce Zhu, Radu Timofte, et~al.
\newblock {NTIRE} 2025 challenge on low light image enhancement: Methods and results.
\newblock In \emph{Proceedings of the IEEE/CVF Conference on Computer Vision and Pattern Recognition (CVPR) Workshops}, 2025{\natexlab{b}}.

\bibitem[Liu et~al.(2019)Liu, Sun, Zhou, Huang, and Darrell]{liu2018rethinking}
Zhuang Liu, Mingjie Sun, Tinghui Zhou, Gao Huang, and Trevor Darrell.
\newblock Rethinking the value of network pruning.
\newblock In \emph{ICLR}, 2019.

\bibitem[Liu et~al.(2021)Liu, Lin, Cao, Hu, Wei, Zhang, Lin, and Guo]{l2loss}
Zhaoyang Liu, Yutong Lin, Yue Cao, Han Hu, Yixuan Wei, Zheng Zhang, Stephen Lin, and Baining Guo.
\newblock Proceedings of the ieee/cvf international conference on computer vision.
\newblock In \emph{Proceedings of the IEEE/CVF International Conference on Computer Vision}, pages 10012--10022, 2021.

\bibitem[Loshchilov and Hutter(2017)]{SGDR}
Ilya Loshchilov and Frank Hutter.
\newblock Sgdr: Stochastic gradient descent with warm restarts.
\newblock In \emph{ICLR}, 2017.

\bibitem[Ma et~al.(2024)Ma, Li, Ren, Sebe, Konukoglu, Gevers, Van~Gool, and Paudel]{ma2024shapesplat}
Qi Ma, Yue Li, Bin Ren, Nicu Sebe, Ender Konukoglu, Theo Gevers, Luc Van~Gool, and Danda~Pani Paudel.
\newblock Shapesplat: A large-scale dataset of gaussian splats and their self-supervised pretraining.
\newblock In \emph{International Conference on 3D Vision 2025}, 2024.

\bibitem[Mao et~al.(2023{\natexlab{a}})Mao, Zhang, Wang, Bai, Bai, Fang, Liu, Li, and Yan]{10208430}
Yanyu Mao, Nihao Zhang, Qian Wang, Bendu Bai, Wanying Bai, Haonan Fang, Peng Liu, Mingyue Li, and Shengbo Yan.
\newblock Multi-level dispersion residual network for efficient image super-resolution.
\newblock In \emph{2023 IEEE/CVF Conference on Computer Vision and Pattern Recognition Workshops (CVPRW)}, pages 1660--1669, 2023{\natexlab{a}}.

\bibitem[Mao et~al.(2023{\natexlab{b}})Mao, Zhang, Wang, Bai, Bai, Fang, Liu, Li, and Yan]{MDRN}
Yanyu Mao, Nihao Zhang, Qian Wang, Bendu Bai, Wanying Bai, Haonan Fang, Peng Liu, Mingyue Li, and Shengbo Yan.
\newblock Multi-level dispersion residual network for efficient image super-resolution.
\newblock In \emph{Proceedings of the IEEE/CVF Conference on Computer Vision and Pattern Recognition Workshops}, pages 1660--1669, 2023{\natexlab{b}}.

\bibitem[Nawa{\l}a et~al.(2024)Nawa{\l}a, Jiang, Zhang, Zhu, Sole, and Bull]{nawala2024bvi}
Jakub Nawa{\l}a, Yuxuan Jiang, Fan Zhang, Xiaoqing Zhu, Joel Sole, and David Bull.
\newblock Bvi-aom: A new training dataset for deep video compression optimization.
\newblock In \emph{2024 IEEE International Conference on Visual Communications and Image Processing (VCIP)}, pages 1--5. IEEE, 2024.

\bibitem[Nie et~al.(2021)Nie, Han, Liu, Xiao, Deng, Xu, and Wang]{ghostSR}
Ying Nie, Kai Han, Zhenhua Liu, An Xiao, Yiping Deng, Chunjing Xu, and Yunhe Wang.
\newblock Ghostsr: Learning ghost features for efficient image super-resolution.
\newblock \emph{CoRR}, abs/2101.08525, 2021.

\bibitem[Park et~al.(2022)Park, Yeo, and Shin]{Park_2022}
Seung Park, Yoon-Jae Yeo, and Yong-Goo Shin.
\newblock Pconv: simple yet effective convolutional layer for generative adversarial network.
\newblock \emph{Neural Computing and Applications}, 34\penalty0 (9):\penalty0 7113–7124, 2022.

\bibitem[Paszke et~al.(2019)Paszke, Gross, Massa, Lerer, and Chintala]{2019PyTorch}
Adam Paszke, Sam Gross, Francisco Massa, Adam Lerer, and Soumith Chintala.
\newblock Pytorch: An imperative style, high-performance deep learning library.
\newblock \emph{arXiv preprint arXiv:1912.01703}, 2019.

\bibitem[Qin et~al.(2024)Qin, Leichner, Delakis, Fornoni, Luo, Yang, Wang, Banbury, Ye, Akin, Aggarwal, Zhu, Moro, and Howard]{qin2024mobilenetv4universalmodels}
Danfeng Qin, Chas Leichner, Manolis Delakis, Marco Fornoni, Shixin Luo, Fan Yang, Weijun Wang, Colby Banbury, Chengxi Ye, Berkin Akin, Vaibhav Aggarwal, Tenghui Zhu, Daniele Moro, and Andrew Howard.
\newblock Mobilenetv4 -- universal models for the mobile ecosystem, 2024.

\bibitem[Qiu et~al.(2023)Qiu, Zhu, Zhu, and Zeng]{qiu2023dual}
Yajun Qiu, Qiang Zhu, Shuyuan Zhu, and Bing Zeng.
\newblock Dual circle contrastive learning-based blind image super-resolution.
\newblock \emph{IEEE Transactions on Circuits and Systems for Video Technology}, 34\penalty0 (3):\penalty0 1757--1771, 2023.

\bibitem[Qu et~al.(2025)Qu, Yuan, Hao, Zhao, Xie, Sun, and Zhou]{qu2025varsr}
Yunpeng Qu, Kun Yuan, Jinhua Hao, Kai Zhao, Qizhi Xie, Ming Sun, and Chao Zhou.
\newblock Visual autoregressive modeling for image super-resolution.
\newblock \emph{arXiv preprint arXiv:2501.18993}, 2025.

\bibitem[Ren et~al.(2023)Ren, Liu, Song, Bi, Cucchiara, Sebe, and Wang]{ren2023masked}
Bin Ren, Yahui Liu, Yue Song, Wei Bi, Rita Cucchiara, Nicu Sebe, and Wei Wang.
\newblock Masked jigsaw puzzle: A versatile position embedding for vision transformers.
\newblock In \emph{Proceedings of the IEEE/CVF Conference on Computer Vision and Pattern Recognition}, pages 20382--20391, 2023.

\bibitem[Ren et~al.(2024{\natexlab{a}})Ren, Li, Liang, Ranjan, Liu, Cucchiara, Gool, Yang, and Sebe]{ren2024sharing}
Bin Ren, Yawei Li, Jingyun Liang, Rakesh Ranjan, Mengyuan Liu, Rita Cucchiara, Luc~V Gool, Ming-Hsuan Yang, and Nicu Sebe.
\newblock Sharing key semantics in transformer makes efficient image restoration.
\newblock \emph{Advances in Neural Information Processing Systems}, 37:\penalty0 7427--7463, 2024{\natexlab{a}}.

\bibitem[Ren et~al.(2024{\natexlab{b}})Ren, Li, Mehta, Timofte, Yu, Wan, Hong, Han, Wu, Zou, et~al.]{ren2024ninth}
Bin Ren, Yawei Li, Nancy Mehta, Radu Timofte, Hongyuan Yu, Cheng Wan, Yuxin Hong, Bingnan Han, Zhuoyuan Wu, Yajun Zou, et~al.
\newblock The ninth ntire 2024 efficient super-resolution challenge report.
\newblock In \emph{Proceedings of the IEEE/CVF Conference on Computer Vision and Pattern Recognition}, pages 6595--6631, 2024{\natexlab{b}}.

\bibitem[Ren et~al.(2025)Ren, Guo, Sun, Wu, Timofte, Li, et~al.]{ntire2025esr}
Bin Ren, Hang Guo, Lei Sun, Zongwei Wu, Radu Timofte, Yawei Li, et~al.
\newblock The tenth {NTIRE} 2025 efficient super-resolution challenge report.
\newblock In \emph{Proceedings of the IEEE/CVF Conference on Computer Vision and Pattern Recognition (CVPR) Workshops}, 2025.

\bibitem[Safonov et~al.(2025)Safonov, Bryntsev, Moskalenko, Kulikov, Vatolin, Timofte, et~al.]{ntire2025ugc}
Nickolay Safonov, Alexey Bryntsev, Andrey Moskalenko, Dmitry Kulikov, Dmitriy Vatolin, Radu Timofte, et~al.
\newblock {NTIRE} 2025 challenge on {UGC} video enhancement: Methods and results.
\newblock In \emph{Proceedings of the IEEE/CVF Conference on Computer Vision and Pattern Recognition (CVPR) Workshops}, 2025.

\bibitem[Shi et~al.(2016)Shi, Caballero, Husz{\'a}r, Totz, Aitken, Bishop, Rueckert, and Wang]{shi2016real}
Wenzhe Shi, Jose Caballero, Ferenc Husz{\'a}r, Johannes Totz, Andrew~P Aitken, Rob Bishop, Daniel Rueckert, and Zehan Wang.
\newblock Real-time single image and video super-resolution using an efficient sub-pixel convolutional neural network.
\newblock In \emph{Proceedings of the IEEE Conference on Computer Vision and Pattern Recognition}, pages 1874--1883, 2016.

\bibitem[Sun et~al.(2022)Sun, Pan, and Tang]{sun2022shufflemixer}
Long Sun, Jinshan Pan, and Jinhui Tang.
\newblock Shufflemixer: An efficient convnet for image super-resolution.
\newblock \emph{Advances in Neural Information Processing Systems}, 35:\penalty0 17314--17326, 2022.

\bibitem[Sun et~al.(2023)Sun, Dong, Tang, and Pan]{sun2023safmn}
Long Sun, Jiangxin Dong, Jinhui Tang, and Jinshan Pan.
\newblock Spatially-adaptive feature modulation for efficient image super-resolution.
\newblock In \emph{ICCV}, 2023.

\bibitem[Sun et~al.(2025{\natexlab{a}})Sun, Alfarano, Duan, Su, Wang, Shi, Timofte, Paudel, Van~Gool, et~al.]{ntire2025event}
Lei Sun, Andrea Alfarano, Peiqi Duan, Shaolin Su, Kaiwei Wang, Boxin Shi, Radu Timofte, Danda~Pani Paudel, Luc Van~Gool, et~al.
\newblock {NTIRE} 2025 challenge on event-based image deblurring: Methods and results.
\newblock In \emph{Proceedings of the IEEE/CVF Conference on Computer Vision and Pattern Recognition (CVPR) Workshops}, 2025{\natexlab{a}}.

\bibitem[Sun et~al.(2025{\natexlab{b}})Sun, Guo, Ren, Van~Gool, Timofte, Li, et~al.]{ntire2025denoising}
Lei Sun, Hang Guo, Bin Ren, Luc Van~Gool, Radu Timofte, Yawei Li, et~al.
\newblock The tenth ntire 2025 image denoising challenge report.
\newblock In \emph{Proceedings of the IEEE/CVF Conference on Computer Vision and Pattern Recognition (CVPR) Workshops}, 2025{\natexlab{b}}.

\bibitem[Tang et~al.(2025)Tang, Guo, Liu, Wang, Hua, Zhong, Xiao, Huang, Song, Liang, Song, He, Bi, Feng, Li, Zhang, and Xu]{tang2025ai4anime}
Yunlong Tang, Junjia Guo, Pinxin Liu, Zhiyuan Wang, Hang Hua, Jia-Xing Zhong, Yunzhong Xiao, Chao Huang, Luchuan Song, Susan Liang, Yizhi Song, Liu He, Jing Bi, Mingqian Feng, Xinyang Li, Zeliang Zhang, and Chenliang Xu.
\newblock Generative ai for cel-animation: A survey.
\newblock \emph{arXiv preprint arXiv:2501.06250}, 2025.

\bibitem[Timofte et~al.(2017{\natexlab{a}})Timofte, Agustsson, Van~Gool, Yang, and Zhang]{NTIRE_2017}
Radu Timofte, Eirikur Agustsson, Luc Van~Gool, Ming-Hsuan Yang, and Lei Zhang.
\newblock Ntire 2017 challenge on single image super-resolution: Methods and results.
\newblock In \emph{CVPR Workshops}, 2017{\natexlab{a}}.

\bibitem[Timofte et~al.(2017{\natexlab{b}})Timofte, Agustsson, Van~Gool, Yang, and Zhang]{flickr2k}
Radu Timofte, Eirikur Agustsson, Luc Van~Gool, Ming-Hsuan Yang, and Lei Zhang.
\newblock Ntire 2017 challenge on single image super-resolution: Methods and results.
\newblock In \emph{Proceedings of the IEEE conference on computer vision and pattern recognition workshops}, pages 114--125, 2017{\natexlab{b}}.

\bibitem[Timofte et~al.(2017{\natexlab{c}})Timofte, Agustsson, Van~Gool, Yang, and Zhang]{timofte2017ntire}
Radu Timofte, Eirikur Agustsson, Luc Van~Gool, Ming-Hsuan Yang, and Lei Zhang.
\newblock Ntire 2017 challenge on single image super-resolution: Methods and results.
\newblock In \emph{CVPR workshops}, pages 114--125, 2017{\natexlab{c}}.

\bibitem[Timofte et~al.(2017{\natexlab{d}})Timofte, Agustsson, Van~Gool, Yang, Zhang, et~al.]{div2k}
Radu Timofte, Eirikur Agustsson, Luc Van~Gool, Ming-Hsuan Yang, Lei Zhang, et~al.
\newblock {NTIRE} 2017 challenge on single image super-resolution: Methods and results.
\newblock In \emph{CVPR Workshops}, 2017{\natexlab{d}}.

\bibitem[Timofte et~al.(2018)Timofte, Agustsson, Gu, Wu, Ignatov, and Van~Gool]{timoftediv2k}
Radu Timofte, Eirikur Agustsson, Shuhang Gu, J Wu, A Ignatov, and L Van~Gool.
\newblock Div2k dataset: Diverse 2k resolution high quality images as used for the challenges@ ntire (cvpr 2017 and cvpr 2018) and@ pirm (eccv 2018), 2018.

\bibitem[Vasluianu et~al.(2025{\natexlab{a}})Vasluianu, Seizinger, Zhou, Chen, Wu, Timofte, et~al.]{ntire2025shadow}
Florin-Alexandru Vasluianu, Tim Seizinger, Zhuyun Zhou, Cailian Chen, Zongwei Wu, Radu Timofte, et~al.
\newblock {NTIRE} 2025 image shadow removal challenge report.
\newblock In \emph{Proceedings of the IEEE/CVF Conference on Computer Vision and Pattern Recognition (CVPR) Workshops}, 2025{\natexlab{a}}.

\bibitem[Vasluianu et~al.(2025{\natexlab{b}})Vasluianu, Seizinger, Zhou, Wu, Timofte, et~al.]{ntire2025ambient}
Florin-Alexandru Vasluianu, Tim Seizinger, Zhuyun Zhou, Zongwei Wu, Radu Timofte, et~al.
\newblock {NTIRE} 2025 ambient lighting normalization challenge.
\newblock In \emph{Proceedings of the IEEE/CVF Conference on Computer Vision and Pattern Recognition (CVPR) Workshops}, 2025{\natexlab{b}}.

\bibitem[Vasu et~al.(2022)Vasu, Gabriel, Zhu, Tuzel, and Ranjan]{mobileone}
Pavan Kumar~Anasosalu Vasu, James Gabriel, Jeff Zhu, Oncel Tuzel, and Anurag Ranjan.
\newblock An improved one millisecond mobile backbone.
\newblock \emph{arXiv preprint arXiv:2206.04040}, 2022.

\bibitem[Wan et~al.(2023)Wan, Yu, Li, Chen, Zou, Liu, Yin, and Zuo]{wan2023swift}
Cheng Wan, Hongyuan Yu, Zhiqi Li, Yihang Chen, Yajun Zou, Yuqing Liu, Xuanwu Yin, and Kunlong Zuo.
\newblock Swift parameter-free attention network for efficient super-resolution.
\newblock \emph{arXiv preprint arXiv:2311.12770}, 2023.

\bibitem[Wan et~al.(2024{\natexlab{a}})Wan, Yu, Li, Chen, Zou, Liu, Yin, and Zuo]{10678610}
Cheng Wan, Hongyuan Yu, Zhiqi Li, Yihang Chen, Yajun Zou, Yuqing Liu, Xuanwu Yin, and Kunlong Zuo.
\newblock Swift parameter-free attention network for efficient super-resolution.
\newblock In \emph{2024 IEEE/CVF Conference on Computer Vision and Pattern Recognition Workshops (CVPRW)}, pages 6246--6256, 2024{\natexlab{a}}.

\bibitem[Wan et~al.(2024{\natexlab{b}})Wan, Yu, Li, Chen, Zou, Liu, Yin, and Zuo]{SPAN2024}
Cheng Wan, Hongyuan Yu, Zhiqi Li, Yihang Chen, Yajun Zou, Yuqing Liu, Xuanwu Yin, and Kunlong Zuo.
\newblock Swift parameter-free attention network for efficient super-resolution.
\newblock In \emph{IEEE Conf. Comput. Vis. Pattern Recog. Worksh.}, 2024{\natexlab{b}}.
\newblock NTIRE 2024 ESR Challenge.

\bibitem[Wan et~al.(2024{\natexlab{c}})Wan, Yu, Li, Chen, Zou, Liu, Yin, and Zuo]{span}
Cheng Wan, Hongyuan Yu, Zhiqi Li, Yihang Chen, Yajun Zou, Yuqing Liu, Xuanwu Yin, and Kunlong Zuo.
\newblock Swift parameter-free attention network for efficient super-resolution.
\newblock In \emph{Proceedings of the IEEE/CVF Conference on Computer Vision and Pattern Recognition}, pages 6246--6256, 2024{\natexlab{c}}.

\bibitem[Wan et~al.(2024{\natexlab{d}})Wan, Yu, Li, Chen, Zou, Liu, Yin, and Zuo]{wan2024swift}
Cheng Wan, Hongyuan Yu, Zhiqi Li, Yihang Chen, Yajun Zou, Yuqing Liu, Xuanwu Yin, and Kunlong Zuo.
\newblock Swift parameter-free attention network for efficient super-resolution.
\newblock In \emph{Proceedings of the IEEE/CVF Conference on Computer Vision and Pattern Recognition (CVPR) Workshops}, 2024{\natexlab{d}}.

\bibitem[Wang et~al.(2023)Wang, Chen, Ni, Liu, and Liu]{wang2023omni}
Hang Wang, Xuanhong Chen, Bingbing Ni, Yutian Liu, and Jinfan Liu.
\newblock Omni aggregation networks for lightweight image super-resolution.
\newblock In \emph{Proceedings of the IEEE/CVF Conference on Computer Vision and Pattern Recognition}, pages 22378--22387, 2023.

\bibitem[Wang et~al.(2024{\natexlab{a}})Wang, Wei, Tang, Cheng, Wang, and Li]{10678094}
Hongyuan Wang, Ziyan Wei, Qingting Tang, Shuli Cheng, Liejun Wang, and Yongming Li.
\newblock Attention guidance distillation network for efficient image super-resolution.
\newblock In \emph{2024 IEEE/CVF Conference on Computer Vision and Pattern Recognition Workshops (CVPRW)}, pages 6287--6296, 2024{\natexlab{a}}.

\bibitem[Wang et~al.(2022)Wang, Xie, Yu, Chan, Loy, and Dong]{basicsr}
Xintao Wang, Liangbin Xie, Ke Yu, Kelvin~C.K. Chan, Chen~Change Loy, and Chao Dong.
\newblock {BasicSR}: Open source image and video restoration toolbox.
\newblock \url{https://github.com/XPixelGroup/BasicSR}, 2022.

\bibitem[Wang(2022{\natexlab{a}})]{EFDN}
Yan Wang.
\newblock Edge-enhanced feature distillation network for efficient super-resolution.
\newblock In \emph{Proceedings of the IEEE/CVF Conference on Computer Vision and Pattern Recognition (CVPR) Workshops}, pages 777--785, 2022{\natexlab{a}}.

\bibitem[Wang(2022{\natexlab{b}})]{wang2022edgeenhancedfeaturedistillationnetwork}
Yan Wang.
\newblock Edge-enhanced feature distillation network for efficient super-resolution, 2022{\natexlab{b}}.

\bibitem[Wang and Cai(2023)]{wang2023single}
Yucong Wang and Minjie Cai.
\newblock A single residual network with esa modules and distillation.
\newblock In \emph{Proceedings of the IEEE/CVF Conference on Computer Vision and Pattern Recognition}, pages 1970--1980, 2023.

\bibitem[Wang et~al.(2024{\natexlab{b}})Wang, Li, Wang, and Liu]{wang2024multi}
Yan Wang, Yusen Li, Gang Wang, and Xiaoguang Liu.
\newblock Multi-scale attention network for single image super-resolution.
\newblock In \emph{Proceedings of the IEEE/CVF Conference on Computer Vision and Pattern Recognition (CVPR) Workshops}, 2024{\natexlab{b}}.

\bibitem[Wang et~al.(2024{\natexlab{c}})Wang, Li, Wang, and Liu]{wang2024plainusr}
Yan Wang, Yusen Li, Gang Wang, and Xiaoguang Liu.
\newblock Plainusr: Chasing faster convnet for efficient super-resolution.
\newblock \emph{arXiv preprint arXiv:2409.13435}, 2024{\natexlab{c}}.

\bibitem[Wang et~al.(2025)Wang, Liang, Zhang, Tian, Wang, Li, Yang, Timofte, Guo, et~al.]{ntire2025lightfield}
Yingqian Wang, Zhengyu Liang, Fengyuan Zhang, Lvli Tian, Longguang Wang, Juncheng Li, Jungang Yang, Radu Timofte, Yulan Guo, et~al.
\newblock {NTIRE} 2025 challenge on light field image super-resolution: Methods and results.
\newblock In \emph{Proceedings of the IEEE/CVF Conference on Computer Vision and Pattern Recognition (CVPR) Workshops}, 2025.

\bibitem[Wu et~al.(2024)Wu, Jiang, Jiang, and Liu]{Wu_cfsr}
Gang Wu, Junjun Jiang, Junpeng Jiang, and Xianming Liu.
\newblock Transforming image super-resolution: A convformer-based efficient approach.
\newblock \emph{IEEE Transactions on Image Processing}, 2024.

\bibitem[Xie et~al.(2025)Xie, Zhang, Li, Fu, Gong, Li, and Zhang]{xie2025mat}
Chengxing Xie, Xiaoming Zhang, Linze Li, Yuqian Fu, Biao Gong, Tianrui Li, and Kai Zhang.
\newblock Mat: Multi-range attention transformer for efficient image super-resolution.
\newblock \emph{IEEE Transactions on Circuits and Systems for Video Technology}, 2025.

\bibitem[Xie et~al.(2024)Xie, Zhou, Li, Lin, and Yan]{xie2024adan}
Xingyu Xie, Pan Zhou, Huan Li, Zhouchen Lin, and Shuicheng Yan.
\newblock Adan: Adaptive nesterov momentum algorithm for faster optimizing deep models.
\newblock \emph{IEEE Transactions on Pattern Analysis and Machine Intelligence}, 2024.

\bibitem[Yang et~al.(2025)Yang, Cai, Ouyang, Vasluianu, Timofte, Ding, Sun, Fu, Li, Ho, Meng, et~al.]{ntire2025reflection}
Kangning Yang, Jie Cai, Ling Ouyang, Florin-Alexandru Vasluianu, Radu Timofte, Jiaming Ding, Huiming Sun, Lan Fu, Jinlong Li, Chiu~Man Ho, Zibo Meng, et~al.
\newblock {NTIRE} 2025 challenge on single image reflection removal in the wild: Datasets, methods and results.
\newblock In \emph{Proceedings of the IEEE/CVF Conference on Computer Vision and Pattern Recognition (CVPR) Workshops}, 2025.

\bibitem[Yang et~al.(2021)Yang, Zhang, Li, and Xie]{yang2021simam}
Lingxiao Yang, Ru-Yuan Zhang, Lida Li, and Xiaohua Xie.
\newblock Simam: A simple, parameter-free attention module for convolutional neural networks.
\newblock In \emph{International conference on machine learning}, pages 11863--11874. PMLR, 2021.

\bibitem[Yoon et~al.(2024)Yoon, Gankhuyag, Park, Son, and Min]{yoon2024casr}
Kihwan Yoon, Ganzorig Gankhuyag, Jinman Park, Haengseon Son, and Kyoungwon Min.
\newblock Casr: Efficient cascade network structure with channel aligned method for 4k real-time single image super-resolution.
\newblock In \emph{Proceedings of the IEEE/CVF Conference on Computer Vision and Pattern Recognition}, pages 7911--7920, 2024.

\bibitem[Yu et~al.(2023)Yu, Li, Li, Jiang, Wu, Fan, and Liu]{yu2023dipnet}
Lei Yu, Xinpeng Li, Youwei Li, Ting Jiang, Qi Wu, Haoqiang Fan, and Shuaicheng Liu.
\newblock Dipnet: Efficiency distillation and iterative pruning for image super-resolution.
\newblock In \emph{Proceedings of the IEEE/CVF Conference on Computer Vision and Pattern Recognition}, pages 1692--1701, 2023.

\bibitem[Yu et~al.(2017)Yu, Liu, Wang, and Tao]{yu2017compressing}
Xiyu Yu, Tongliang Liu, Xinchao Wang, and Dacheng Tao.
\newblock On compressing deep models by low rank and sparse decomposition.
\newblock In \emph{Proceedings of the IEEE Conference on Computer Vision and Pattern Recognition}, pages 7370--7379, 2017.

\bibitem[Zama~Ramirez et~al.(2025)Zama~Ramirez, Tosi, Di~Stefano, Timofte, Costanzino, Poggi, Salti, Mattoccia, et~al.]{ntire2025hrdepth}
Pierluigi Zama~Ramirez, Fabio Tosi, Luigi Di~Stefano, Radu Timofte, Alex Costanzino, Matteo Poggi, Samuele Salti, Stefano Mattoccia, et~al.
\newblock {NTIRE} 2025 challenge on hr depth from images of specular and transparent surfaces.
\newblock In \emph{Proceedings of the IEEE/CVF Conference on Computer Vision and Pattern Recognition (CVPR) Workshops}, 2025.

\bibitem[Zamfir et~al.(2024)Zamfir, Wu, Mehta, Zhang, and Timofte]{zamfir2024see}
Eduard Zamfir, Zongwei Wu, Nancy Mehta, Yulun Zhang, and Radu Timofte.
\newblock See more details: Efficient image super-resolution by experts mining.
\newblock In \emph{Forty-first International Conference on Machine Learning}, 2024.

\bibitem[Zamir et~al.(2022)Zamir, Arora, Khan, Hayat, Khan, and Yang]{Restormer}
Syed~Waqas Zamir, Aditya Arora, Salman Khan, Munawar Hayat, Fahad~Shahbaz Khan, and Ming-Hsuan Yang.
\newblock Restormer: Efficient transformer for high-resolution image restoration.
\newblock In \emph{CVPR}, 2022.

\bibitem[Zhang et~al.(2022)Zhang, Huang, Liu, Wang, and Jin]{swinfir}
Dafeng Zhang, Feiyu Huang, Shizhuo Liu, Xiaobing Wang, and Zhezhu Jin.
\newblock Swinfir: Revisiting the swinir with fast fourier convolution and improved training for image super-resolution, 2022.

\bibitem[Zhang(2024)]{HiT-SR}
Xiang Zhang.
\newblock Hit-sr: Hierarchical transformer for efficient image super-resolution.
\newblock \url{https://github.com/XiangZ-0/HiT-SR}, 2024.
\newblock GitHub repository.

\bibitem[Zhang et~al.(2018)Zhang, Zhou, Lin, and Sun]{Shufflenet}
Xiangyu Zhang, Xinyu Zhou, Mengxiao Lin, and Jian Sun.
\newblock Shufflenet: An extremely efficient convolutional neural network for mobile devices.
\newblock \emph{Proceedings of the IEEE conference on computer vision and pattern recognition}, 2018.

\bibitem[Zhang et~al.(2021{\natexlab{a}})Zhang, Zeng, and Zhang]{ECBSR}
Xindong Zhang, Hui Zeng, and Lei Zhang.
\newblock Edge-oriented convolution block for real-time super resolution on mobile devices.
\newblock In \emph{{MM} '21: {ACM} Multimedia Conference, Virtual Event, China, October 20 - 24, 2021}, pages 4034--4043. {ACM}, 2021{\natexlab{a}}.

\bibitem[Zhang et~al.(2021{\natexlab{b}})Zhang, Zeng, and Zhang]{Zhang2021EdgeorientedCB}
Xindong Zhang, Huiyu Zeng, and Lei Zhang.
\newblock Edge-oriented convolution block for real-time super resolution on mobile devices.
\newblock \emph{Proceedings of the 29th ACM International Conference on Multimedia}, 2021{\natexlab{b}}.

\bibitem[Zhang et~al.(2021{\natexlab{c}})Zhang, Zeng, and Zhang]{zhang2021edge}
Xindong Zhang, Hui Zeng, and Lei Zhang.
\newblock Edge-oriented convolution block for real-time super resolution on mobile devices.
\newblock In \emph{Proceedings of the 29th ACM International Conference on Multimedia}, pages 4034--4043, 2021{\natexlab{c}}.

\bibitem[Zhang et~al.(2024{\natexlab{a}})Zhang, Zhang, and Yu]{zhang2024hit}
Xiang Zhang, Yulun Zhang, and Fisher Yu.
\newblock Hit-sr: Hierarchical transformer for efficient image super-resolution.
\newblock In \emph{European Conference on Computer Vision}, pages 483--500. Springer, 2024{\natexlab{a}}.

\bibitem[Zhang et~al.(2024{\natexlab{b}})Zhang, Zhang, and Yu]{zhang2024hitsrhierarchicaltransformerefficient}
Xiang Zhang, Yulun Zhang, and Fisher Yu.
\newblock Hit-sr: Hierarchical transformer for efficient image super-resolution.
\newblock \emph{arXiv preprint}, arXiv:2407.05878, 2024{\natexlab{b}}.

\bibitem[Zhang et~al.(2023)Zhang, Zhang, Chen, Li, Timofte, et~al.]{zhang2023ntire}
Yulun Zhang, Kai Zhang, Zheng Chen, Yawei Li, Radu Timofte, et~al.
\newblock {NTIRE} 2023 challenge on image super-resolution (x4): Methods and results.
\newblock In \emph{Proceedings of the IEEE/CVF Conference on Computer Vision and Pattern Recognition Workshops}, 2023.

\bibitem[Zhao et~al.(2020)Zhao, Kong, He, Qiao, and Dong]{zhao2020efficient}
Hengyuan Zhao, Xiangtao Kong, Jingwen He, Yu Qiao, and Chao Dong.
\newblock Efficient image super-resolution using pixel attention.
\newblock In \emph{European Conference on Computer Vision}, pages 56--72. Springer, 2020.

\bibitem[Zhao et~al.(2024)Zhao, Liu, Ren, Dai, and Sebe]{zhao2024denoising}
Mengyi Zhao, Mengyuan Liu, Bin Ren, Shuling Dai, and Nicu Sebe.
\newblock Denoising diffusion probabilistic models for action-conditioned 3d motion generation.
\newblock In \emph{ICASSP 2024-2024 IEEE International Conference on Acoustics, Speech and Signal Processing (ICASSP)}, pages 4225--4229. IEEE, 2024.

\bibitem[Zheng et~al.(2024{\natexlab{a}})Zheng, Sun, Dong, and Pan]{smfanet}
Mingjun Zheng, Long Sun, Jiangxin Dong, and Jinshan Pan.
\newblock Smfanet: A lightweight self-modulation feature aggregation network for efficient image super-resolution.
\newblock In \emph{ECCV}, 2024{\natexlab{a}}.

\bibitem[Zheng et~al.(2024{\natexlab{b}})Zheng, Sun, Dong, and Pan]{zheng2024smfanet}
Mingjun Zheng, Long Sun, Jiangxin Dong, and Jinshan Pan.
\newblock Smfanet: A lightweight self-modulation feature aggregation network for efficient image super-resolution.
\newblock In \emph{European Conference on Computer Vision}, pages 359--375. Springer, 2024{\natexlab{b}}.

\bibitem[Zheng et~al.(2025)Zheng, Luo, Zhou, and Wang]{zheng2025distilling}
Xu Zheng, Yunhao Luo, Pengyuan Zhou, and Lin Wang.
\newblock Distilling efficient vision transformers from cnns for semantic segmentation.
\newblock \emph{Pattern Recognition}, 158:\penalty0 111029, 2025.

\bibitem[Zhou et~al.(2023)Zhou, Li, Guo, Bai, Cheng, and Hou]{zhou2023srformer}
Yupeng Zhou, Zhen Li, Chun-Le Guo, Song Bai, Ming-Ming Cheng, and Qibin Hou.
\newblock Srformer: Permuted self-attention for single image super-resolution.
\newblock In \emph{Proceedings of the IEEE/CVF International Conference on Computer Vision}, pages 12780--12791, 2023.

\bibitem[Zhu et~al.(2024)Zhu, Liao, Zhang, Wang, Liu, and Wang]{vim}
Lianghui Zhu, Bencheng Liao, Qian Zhang, Xinlong Wang, Wenyu Liu, and Xinggang Wang.
\newblock Vision mamba: Efficient visual representation learning with bidirectional state space model.
\newblock In \emph{Forty-first International Conference on Machine Learning}, 2024.

\bibitem[Zhu et~al.(2023)Zhu, Li, and Li]{zhu2023attention}
Qiang Zhu, Pengfei Li, and Qianhui Li.
\newblock Attention retractable frequency fusion transformer for image super resolution.
\newblock In \emph{Proceedings of the IEEE/CVF Conference on Computer Vision and Pattern Recognition}, pages 1756--1763, 2023.

\end{thebibliography}
}

\end{document}